\documentclass{article} 
\usepackage{iclr2026_conference,times}
\iclrpreprint
\usepackage{hyperref}
\usepackage[utf8]{inputenc} 
\usepackage[T1]{fontenc}    
\usepackage{url}            
\usepackage{booktabs}       
\usepackage{amsfonts}       
\usepackage{nicefrac}       
\usepackage{microtype}      
\usepackage[table]{xcolor}
\usepackage{multirow}       
\usepackage{adjustbox}      
\usepackage{enumitem}
\usepackage{array}
\usepackage{graphicx}
\usepackage{wrapfig}
\usepackage{epsfig}
\usepackage{amsthm}
\usepackage{amsmath}
\usepackage{amssymb}
\usepackage{subfigure}
\usepackage{marvosym}
\usepackage[ruled,vlined]{algorithm2e}
\usepackage{color}
\usepackage{algpseudocode}
\usepackage{babel}
\usepackage{caption}
\usepackage{hhline} 
\usepackage{xspace} 
\usepackage{colortbl}
\usepackage{dblfloatfix}
\usepackage{balance}
\usepackage{diagbox}
\usepackage{bbm}
\usepackage{forest}
\usepackage{subcaption}
\usepackage{comment}
\definecolor{lowrisk}{RGB}{0,128,0}      
\definecolor{medrisk}{RGB}{255,140,0}    
\definecolor{highrisk}{RGB}{220,20,60}   
\makeatletter
\DeclareRobustCommand\onedot{\futurelet\@let@token\@onedot}
\def\@onedot{\ifx\@let@token.\else.\null\fi\xspace}
\def\eg{e.g\onedot} 
\def\ie{i.e\onedot} 
 
 \def\vs{vs\onedot}

\newcommand{\bestResults}{%
Best result \textbf{bolded}. The second best  \underline{underlined}.}
\newcommand{\briefExplain}{%
LVEF refers to left ventricular ejection fraction; Dx denotes diagnosis; SubDx denotes subdiagnosis; SupDx denotes superdiagnosis; Form refers to waveform morphology; Rhyth denotes rhythm; Arrhy denotes arrhythmia.}

\newcommand{\@toptitlebar}{
  \hrule height 4\p@
  \vskip 0.15in
  \vskip -\parskip%
}
\newcommand{\@bottomtitlebar}{
  \vskip 0.29in
  \vskip -\parskip
  \hrule height 1\p@
  \vskip 0.29in%
}
\usepackage{tikz}
\newcommand{\cmark}{\tikz[scale=0.18,baseline=-0.6ex]{\draw[line width=0.8mm, color=green!70!black]  (-0.5,0.8) --(0,0) -- (1,1.5);}}

\newcommand{\samerel}{$\spadesuit$} 
\newcommand{\diffrel}{$\diamondsuit$}
\newcommand{\acro}{\textsl{CLEF}}

\title{
\protect\includegraphics[width=0.045\textwidth]{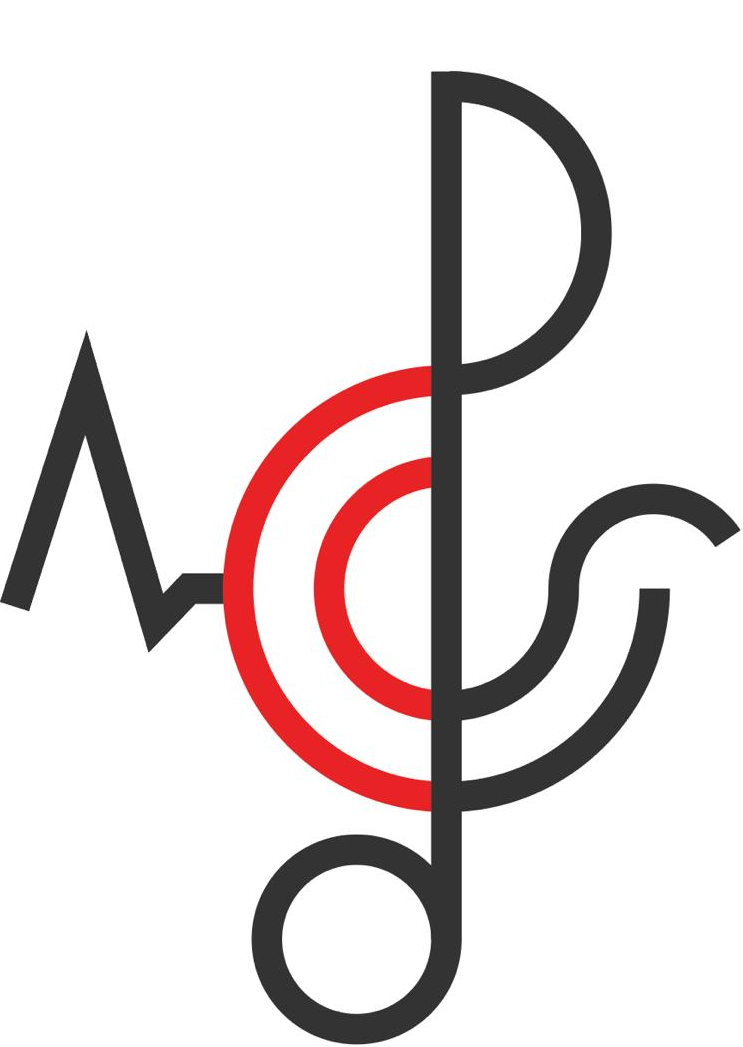}
    \acro{}:~Clinically-Guided~Contrastive~Learning for Electrocardiogram Foundation Models
}
\author{Yuxuan Shu\thanks{Work done during the author’s internship at Nokia Bell Labs (email: \texttt{{\small yuxuan.shu.22@ucl.ac.uk}})} \\ 
      University College London \\
      \And
        Peter H. Charlton \\
       Nokia Bell Labs  \\
      \And
      Fahim Kawsar \\
      Nokia Bell Labs \& University of Glasgow\\
      \AND
      Jussi Hernesniemi \\
      Heart Hospital, Tampere University \\
      \And
      Mohammad Malekzadeh \\ 
      Nokia Bell Labs \\
      }
\begin{document}
\maketitle
\begin{abstract}
The electrocardiogram~(ECG) is a key diagnostic tool in cardiovascular health. Single-lead ECG recording is integrated into both clinical-grade and consumer wearables. While self-supervised pretraining of foundation models on unlabeled ECGs improves diagnostic performance, existing approaches do not incorporate domain knowledge from clinical metadata. We introduce a novel contrastive learning approach that utilizes an established clinical risk score to adaptively weight negative pairs: {\em clinically-guided contrastive learning}. It aligns the similarities of ECG embeddings with clinically meaningful differences between subjects, with an explicit mechanism to handle missing metadata. On 12-lead ECGs from 161K patients in MIMIC-IV dataset, we pretrain single-lead {\em ECG foundation models} at three scales, collectively called \acro{}, using only routinely-collected metadata without requiring per-sample ECG annotations. We evaluate \acro{} on $18$ clinical classification and regression tasks across $7$ held-out datasets, and benchmark against $5$ foundation model baselines and $3$ self-supervised algorithms. When pretrained on 12-lead ECG data and tested on lead-I data, \acro{} outperforms self-supervised foundation model baselines: the medium-sized \acro{} achieves average AUROC improvements of at least $2.6\%$ in classification and average reductions in MAEs of at least $3.2\%$ in regression. Comparing with existing self-supervised learning algorithms, \acro{} improves the average AUROC by at least $1.8\%$. Moreover, when pretrained only on lead-I data for classification tasks, \acro{} performs comparably to the state-of-the-art ECGFounder, which was trained in a supervised manner. Overall, \acro{} allows more accurate and scalable single-lead ECG analysis, advancing remote health monitoring. Code and pretrained \acro{} models are available at: \href{https://github.com/Nokia-Bell-Labs/ecg-foundation-model}{github.com/Nokia-Bell-Labs/ecg-foundation-model}.

\end{abstract}

\section{Introduction}
\label{sec:introduction}

The electrocardiogram (ECG)~\citep{geselowitz1989theory} captures the heart's electrical activity as a sequence of voltage fluctuations~\citep{lilly2012pathophysiology}. While ECG interpretation is a clinical task, ECG-based AI started matching clinician performance, and enabled large-scale applications such as detecting heart conditions from wearables. \citet{yao2021artificial} found out that clinicians using AI were 30\% more accurate in diagnosing left ventricular dysfunction from ECGs than those without AI support. The rise of wearables and mobile health devices with single-lead ECGs enables continuous monitoring beyond clinical settings, supporting early cardiac event detection, long-term tracking, and proactive interventions~\citep{miller2025wearable, xu2025lsm}. Nevertheless, accurate analysis of single-lead ECGs remains a challenge due to the reduced spatial information and the increased noise and motion artifacts~\citep{khamisQRS2016,halvaeiIdentification2021,khunte2023detection}. Prior work has shown that single-lead ECG devices can be used to diagnose atrial fibrillation \cite{svennbergMass2015}, and that spatial correlations among ECG leads can help infer missing leads~\citep{pnelwan2004reconstruction}. Single-lead ECG is used for practical diagnoses, especially in wearable contexts~\citep{qin2023mvkt,li2024electrocardiogram}.

Collecting sufficiently large, human-annotated, single-lead ECGs is impractical, given the variability in downstream tasks across devices and health conditions. ECGs from healthcare organizations are usually provided in the standard 12-lead format but contain limited annotations. A more practical solution is to pretrain on 12-lead ECGs to learn generalizable representations, then fine-tune on smaller, task-specific data curated for the target health condition.
Often, pretraining is performed using {\em contrastive learning} with data augmentation~\citep{chen2020simple, soltaniehAnalysis2022a}, encouraging the representations of similar samples to be as close to one another while pushing apart those of dissimilar ones. This not only uses information from within samples (\eg data augmentations), but also uses contextual factors, which can provide important cues for similarity that are not directly observable in the input. Contrastive learning for images is guided by the spatial proximity of image viewpoints or content~\citep{thoma2020geometrically,thoma2020soft}, or by sample identities~\citep{haslum2024metadata}. Prior work also incorporates subject and signal attributes in contrastive learning~\citep{kim2025adaptive}, but did not incorporate clinical knowledge, nor handle missing metadata.

Metadata recorded alongside ECGs, such as patient demographics, provides a potential source of contextual data. Some use metadata as model inputs~\citep{erturk2025sensor}, while others have incorporated metadata prediction as part of the pretraining objective~\citep{wang2024anyecg}, or as downstream prediction targets (e.g., predicting gender or age from ECGs~\citep{li2024electrocardiogram}). Since gender or age is not a primary objective, this merely demonstrates the model’s ability to distinguish between patient groups. In this paper, we take a more integrative strategy to employ metadata as contextual information in contrastive pretraining, since different people might be associated with different risks of a particular health outcome. We leverage metadata in the form of {\em clinically validated risk scores} to pretrain a foundation model (FM) for single-lead ECGs. Risk scores estimate a person's risk of experiencing an adverse health outcome in a specified time period, such as risk of developing cardiovascular disease in the next ten years~\citep{conroy2003estimation,score22021score2,score2-op2021score2-op}. They take various metadata variables as inputs (\eg, patient demographics, past medical history, and comorbidities), and output a quantitative assessment of the risk of adverse outcomes. Risk scores can guide clinical decisions~\cite{hughes2023deep}, such as whether to prescribe a drug~\citep{lipRefining2010}, or perform invasive procedures~\citep{foxPrediction2006}. Our \textbf{contributions} are:

\begin{figure}
    \centering
    \includegraphics[width=\linewidth]{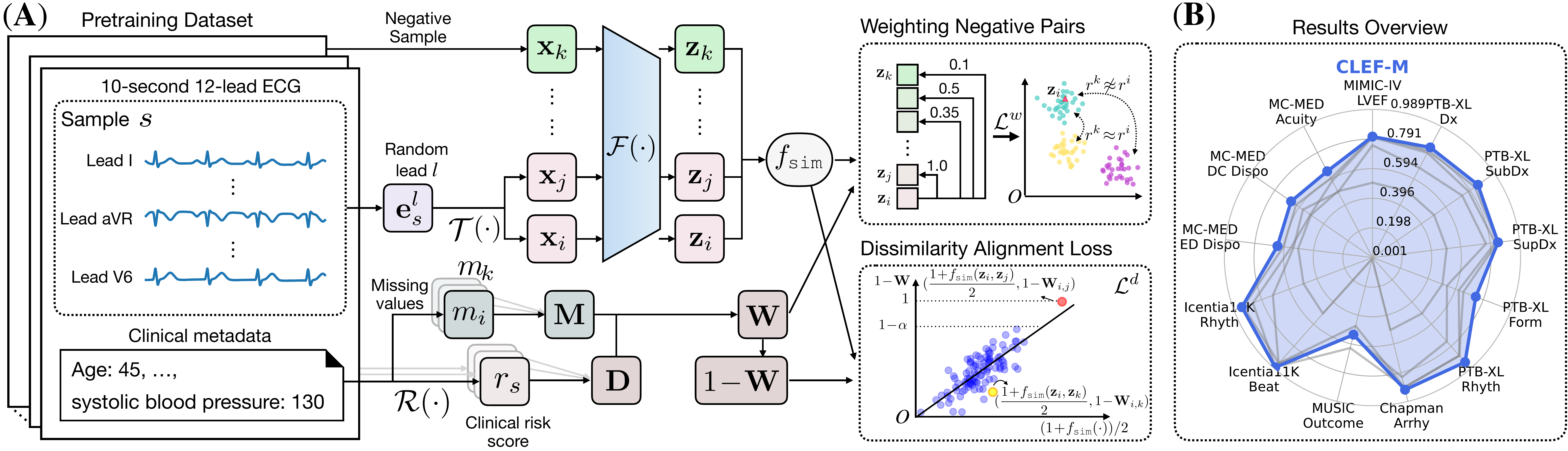}
    \caption{\acro{}'s framework and performance overview (see~\S\ref{sec:method} for notations). \textbf{(A)} Our clinically-guided contrastive pretraining. Key components include a negative weighting loss $\mathcal{L}^w$ and a dissimilarity alignment loss $\mathcal{L}^d$ that work in tandem to guide contrastive learning with clinical knowledge. \textbf{(B)} Spider plot on AUROC performance of \acro-M (our medium-sized model)  across $13$ downstream classification tasks. Baseline performances are in gray lines (see \S\ref{sec:results}).}
    \label{fig:figure1}
    \vspace{-0.4cm}
\end{figure}

\begin{enumerate}[leftmargin=*, topsep=-2pt, itemsep=-1.6pt]
    \item \textbf{Clinically-guided contrastive learning for ECG representation learning.} We leverage metadata to calculate clinically informative risk scores that provide flexible similarity relationships between unlabeled ECG samples. Unlike classic contrastive approaches that rely on binary notions of similarity/dissimilarity, our method incorporates clinical risk scores as soft guidance, enabling the model to learn richer and more nuanced representations.
    \item \textbf{Extensive empirical analysis and benchmarking for ECG FMs.} 
    We: (i) evaluate downstream performance on lead I/II ECG classification and regression, (ii) compare against 12-lead ECG approaches, (iii) assess robustness across different model architectures and pretraining methods, (iv) perform ablation studies on loss components and handling missing metadata, and (v) use linear probing analysis to assess representation quality.
    \item \textbf{Effectiveness across diverse downstream tasks.} We evaluate on a wide range of applications. \acro{} consistently consistently outperforms strong baselines, achieving an average improvement of $3.1\%$ in classification AUROC and a reduction of $2.9\%$ in regression MAE.
\end{enumerate}

\section{Clinically-Guided Contrastive Pretraining}
\label{sec:method}

Let $\mathbbm{D}=\{(\mathbf{e}_s=(\mathbf{e}_s^1, \dots, \mathbf{e}_s^{12}), \texttt{a}_s)\}_{s=1}^{N}$ denote the {\em pretraining dataset}, where $\mathbf{e}_s^l\in \mathbb{R}^t$, of length $t$, represents the ECG signal from lead $l$ of the $s$-th sample, $\texttt{a}_s=\{\texttt{a}^1_s,\dots,\texttt{a}^A_s\}$ denotes demographic or clinical {\em metadata} (i.e., attributes like age or blood pressure) associated with the subject and not labels specific to the current ECG recording, and $N$ is the number of 12-lead ECG samples. Importantly, certain attributes might be missing in some samples' metadata. Let $\mathcal{R}$ denote a standard, clinically-validated function that combines a subject's metadata into a numerical estimate of the likelihood of a future health outcome (\eg sudden cardiac arrest). We define $r_s = \mathcal{R}(\texttt{a}_s)$ as a {\em risk score} associated to the sample $\mathbf{e}_s$. Our objective is to train a single-lead ECG FM $\mathcal{F}(\cdot)$ on $\mathbb{D}$ such that the geometry of the {\em embedding} space, i.e., outputs $\mathbf{z}=\mathcal{F}(\mathbf{e}_s^l)\in \mathbb{R}^h$, with a dimension of $h$, reflects clinically meaningful similarities according to the risk score function $\mathcal{R}(\cdot)$. 

To measure the success of our objective, we attach a linear layer $\mathcal{G}(\cdot)$ on top of the embeddings of condition-specific single-lead ECGs produced by $\mathcal{F}(\cdot)$. Let $\mathbbm{C}=\{(\mathbf{e}^l_s, {y}_s)\}_{s=1}^{M}$ denote a {\em downstream evaluation dataset}, where $\mathbf{e}_s^l\in \mathbb{R}^t$, represents the ECG signal from lead $l$, and $y_s$ indicates the ground-truth label associated to that ECG signal. Note that the lead is fixed across all samples in the test set (\eg lead I). We evaluate $\mathcal{F}(\cdot)$ under two downstream scenarios: (1) {\em fine-tuning}, where all parameters of $\mathcal{F}(\cdot)$ and $\mathcal{G}(\cdot)$ are updated during training on the training split of $\mathbbm{C}$, and (2) {\em linear probing}, where the pretrained $\mathcal{F}(\cdot)$ is kept frozen and only the linear head $\mathcal{G}(\cdot)$ is updated to map the embeddings $\mathbf{z}=\mathcal{F}(\mathbf{e}_s^l)$ to task-specific labels. To maintain a fair comparison, we adopt a consistent set of hyperparameters across all experiments; further details are provided in~\appendixautorefname~\ref{appsubsec:experiment_setup}.

For binary labels $y\in \{0,1\}$, we report the AUROC metric, which measures the model's ability to discriminate between the positive and negative classes by computing the area under the Receiver Operating Characteristic (ROC) curve, representing the trade-off between true positive rate and false positive rate across all possible classification thresholds. For categorical labels $y\in \{1,\cdots,K\}$ where $K$ is the number of classes, we report the macro AUROC score using the one-vs-rest extension, where each class is compared against all others and the resulting scores are macro-averaged (\ie unweighted mean). For numerical labels $y\!\in\!\mathbb{R}_0$, we report the mean absolute error (MAE) between the predicted and true values, providing an interpretable metric of prediction accuracy in the same units as the target variable. Performance is summarized as a percentage improvement over baseline.

\subsection{Contrastive Learning}

{\em Contrastive Loss.} 
\label{subsec:preliminary}
We revisit contrastive learning~\citep{chen2020simple}. Given a sample $\mathbf{e}^l_s\overset{\text{i.i.d.}}{\sim} \mathbf{e}_s\overset{\text{i.i.d.}}{\sim} \mathbbm{D}$, an independent stochastic transformation $\mathcal{T}(\cdot)$ is applied to $\mathbf{e}^l_s$ to obtain an augmented {\em view}, denoted as $\mathbf{x}_i=\mathcal{T}(\mathbf{e}^l_s)$. Similarly, we generate another view $\mathbf{x}_j=\mathcal{T}(\mathbf{e}^l_s)$ forming a {\em positive pair} $(\mathbf{x}_i, \mathbf{x}_j)$ in contrastive learning. A batch of $B$ samples yields $2B$ augmented views, such that any $k\in\{1,2,\dots,2B\} \setminus \{i,j\}$ forms {\em negative pairs} with both $\mathbf{x}_i$ and $\mathbf{x}_j$, \ie, $(\mathbf{x}_i, \mathbf{x}_k)$ and $(\mathbf{x}_j, \mathbf{x}_k)$. Using a model $\mathcal{F}(\cdot)$, the augmented views are encoded into embeddings denoted as $\mathbf{z}_i, \mathbf{z}_j, \, \text{and} \, \mathbf{z}_k\!\in\!\mathbb{R}^h$. For each pair of $(\mathbf{x}_i, \mathbf{x}_j)$, the classic contrastive loss, termed NT-Xent~\citep{chen2020simple}, is given by
\begin{equation}\label{eqn_ccl}
    \mathcal{L}_{i,j} = - \log \frac{\exp \left( f_{\texttt{sim}} \left( \mathbf{z}_i, \mathbf{z}_j \right) / \tau \right) }{\sum^{2B}_{k=1}  \mathbbm{1}_{[k \neq i]} \exp \left(f_{\texttt{sim}} \left(\mathbf{z}_i, \mathbf{z}_k \right) / \tau \right)} \, \, ,
\end{equation}
where $f_{\texttt{sim}}(\mathbf{z}_i, \mathbf{z}_j) = \frac{\mathbf{z}_i^\top \mathbf{z}_j}{\vert\vert \mathbf{z}_i \vert\vert \cdot \vert\vert \mathbf{z}_j \vert\vert}\!\in\![-1,1]$ is cosine similarity, and $\tau$ is a temperature hyperparameter.

{\em Weighted Contrastive Loss.}
\label{subsec:soft_contrastive}
While Eq.~\eqref{eqn_ccl} treats all negative pairs equally, recent works highlight that negative pairs could contribute unequally to representation learning~\cite{robinson2021contrastive,li2023rethinking,zhuang2024not,yang2024does}. Conceptually, in the embedding space, negative samples are pushed away from a positive sample according to their similarity. Specifically, the less similar a negative sample is to the positive, the more it is pushed away. By introducing a weighting term $\mathbf{W}_{ik}$ for each pair of negative embeddings $(\mathbf{z}_i, \mathbf{z}_k)$, we can adjust the relative importance of negatives according to their similarity to the positive, having
\begin{equation}\label{eqn_wcl}
   \mathcal{L}^{w}_{i,j} = - \log \frac{\exp \left( f_{\texttt{sim}} \left( \mathbf{z}_i, \mathbf{z}_j \right) / \tau \right) }{\sum^{2B}_{k=1}  \mathbbm{1}_{[k \neq i]} \mathbf{W}_{ik} \exp \left(f_{\texttt{sim}} \left(\mathbf{z}_i, \mathbf{z}_k \right) / \tau \right)} \, \,.
\end{equation}
Eq.~\eqref{eqn_wcl} generalizes Eq.~\eqref{eqn_ccl} by replacing the uniform treatment of negatives with an adaptive weighting.

\subsection{Clinical Metadata}
\label{subsec:risk_score}
{\em Cardiovascular Risk Score.} A risk score, quantified on clinical and demographic attributes, is the likelihood of experiencing an adverse outcome. They translate complex physiological factors into interpretable values that guide clinical actions and interventions. The Systematic Coronary Risk Evaluation 2 score (SCORE2) commonly used by healthcare professionals, estimates the $10$-year risk of cardiovascular disease events (death, myocardial infarction, or non-fatal stroke). SCORE2 is obtained based on $7$ variables: age $(\texttt{a}^1_s)$, gender $(\texttt{a}^2_s)$, smoking status $(\texttt{a}^3_s)$, systolic blood pressure (SBP, $\texttt{a}^4_s$), diabetes status $(\texttt{a}^5_s)$, total cholesterol $(\texttt{a}^6_s)$, and high-density lipoprotein cholesterol $(\texttt{a}^7_s)$. Let $\mathbf{u}_s = [u_s^1, \cdots,u_s^6] = g(\mathbf{a}_s)$ denote the vector after standardization of the ordered metadata feature $\mathbf{a}_s = [\texttt{a}^1_s,\texttt{a}^3_s,\dots,\texttt{a}^7_s]$. For each sample $\mathbf{e}_s$, we assign a risk score $r_s\!\in\![0,1]$, defined by the risk score function $\mathcal{R}(\cdot)$ as that SCORE2 value of $\mathbf{e}_s$:
\begin{equation}\label{eqn_risk_score}
    r_s \;\overset{\text{def}}{\equiv}\; \mathcal{R}(\texttt{a}_s) 
= 1 - S_0(t)^{\exp(\mathbf{b}_1 \mathbf{u}_s^\top + u_s^1 \mathbf{b}_2 \mathbf{u}_s^\top)} \, .
\end{equation}
where $S_0(t)$ is the baseline survival
\footnote{The probability of not experiencing a cardiovascular event over $10$ years for a reference individual.} function at year $t$, and $\mathbf{b}_1, \mathbf{b}_2$ are the variable's coefficient estimated from the European cohort data covering over 600K individuals with more than 30K cardiovascular events (values varying by age and gender). Given that augmentations do not change metadata, for $\mathbf{x}_i, \mathbf{x}_j$ of sample $s$, we have $r^{i}=r^{j}=r_s$ ($r^{i},r^{j}$ are risk scores for $\mathbf{x}_i,\mathbf{x}_j$). We provide further details on calculating SCORE2 in~\appendixautorefname~\ref{appsubsec:obtain_riskscore}. The \acro~framework supports any validated risk score. While we use SCORE2, we note that its development on European populations might limit its generalisability, and the optimal score may differ by application (see Appendix \ref{app:choice_risk_score}).

{\em Handling Missing Metadata.}
MIMIC-IV-ECG~\citep{gow2023mimic-iv-ecg} contains $161{,}352$ unique subjects, each with multiple 12-lead ECG recordings. For each subject, the dataset reports only three variables: age, gender, and systolic blood pressure. This is a common challenge in real-world settings, where certain metadata variables are often unavailable either at a specific institution or for particular patients. Such variables are typically collected independently rather than at the time of the ECG recording. To address this issue, we assign different {\em multipliers} to negative pairs $(\mathbf{x}_i,\mathbf{x}_k)$ based on the number of metadata variables available when calculating their risk scores ($r^i$, $r^k$). Specifically, we calculate
\begin{equation}\label{eqn_missing_meta}
    \mathbf{M}_{ik} = \exp \left( - \frac{A-m_i}{A}\!\times\!\frac{A-m_k}{A} \right) \, \, ,
\end{equation}
where $A$ is the number of variables (e.g., $A=7$ in SCORE2), and $m_i$/$m_k$ are the number of missing variables for $\mathbf{x}_i$/$\mathbf{x}_k$. Thus, $\mathbf{M}_{ik}\!\in\!(0, 1]$ indicates the level of missing metadata for a pair $(\mathbf{x}_i,\mathbf{x}_k)$, which functions as a relative reliability adjustment to ensure that only well-supported risk differences meaningfully shape the embedding geometry, we further elaborate this in~\sectionautorefname~\ref{subsec:guiding_representation}.

\subsection{Guiding Representation Learning with Clinical Risk Scores}
\label{subsec:guiding_representation}
{\em Risk Score Dissimilarity.}
We aim to enhance weighted contrastive loss in Eq.~\eqref{eqn_wcl} by using risk scores in Eq.~\eqref{eqn_risk_score} for pretraining our ECG FM $\mathcal{F}(\cdot)$. Our objective is to guide $\mathcal{F}(\cdot)$ toward learning embeddings that capture clinically relevant patterns, building a latent space for ECG signals where the distance between embeddings reflects dissimilarities in risk scores. For a negative pair $(\mathbf{x}_i,\mathbf{x}_k)$ with corresponding $(r^i,r^k)$, let $\delta_{ik}=(r^i - r^k)^2$. We define a negative pair's {\em dissimilarity}:
\begin{equation} \label{eqn_rs_diss}
    \mathbf{D}_{i,k} = (1 - \alpha) \frac{\delta_{ik} - \delta^{min}}{\delta^{max} - \delta^{min}} + \alpha, \quad \text{where} \quad \delta^{max} = \max_{i, k} (\delta_{ik}) \quad \text{and} \quad \delta^{min} = \min_{i, k} (\delta_{ik}).
\end{equation}
Parameter $\alpha\!\in\![0,1]$ controls the minimum distance between the positive pair and all negative pairs. Dissimilarity to all negative pairs stays within  
$\left[\alpha, 1 \right]$, having $\mathbf{D}_{i,j}=0$ only for the positive pair. For $(\mathbf{x}_i,\mathbf{x}_k)$, even when their metadata indicates $r^i=r^k$, we have their dissimilarity $\mathbf{D}_{i,k}$ at least $\alpha$.

{\em Weighting Negative Pairs.}
By combining  Eq.~\eqref{eqn_missing_meta}~and~\eqref{eqn_rs_diss} via the Hadamard product, we define the {\em weight matrix} $\mathbf{W} = \mathbf{D} \odot \mathbf{M}$, where each entry is given by $\mathbf{W}_{ik} = \mathbf{D}_{i,k}\cdot\mathbf{M}_{i,k}\!\in\![0,1]$. This value represents the weight assigned to pushing the embedding of $\mathbf{x}_k$ away from that of $\mathbf{x}_i$. Intuitively, the weight increases when the risk scores of $\mathbf{x}_i$ and $\mathbf{x}_k$ differ more; conversely, it decreases when the risk scores are similar. Importantly, when many metadata variables are missing, and SCORE2 would rely heavily on default imputations (see~\appendixautorefname~\ref{appsubsec:obtain_riskscore}), the resulting risk differences are thus less trustworthy. In such cases, the corresponding $\textbf{M}_{ik}$ between the negative pairs moves the weighting closer to a uniform SimCLR-like behavior, preventing uncertain or noisy risk differences from exerting undue influence on the contrastive objective. For each batch of data, we calculate our {\em clinically-guided contrastive loss} by
\begin{equation}\label{eqn_weighted_cl}
    \mathcal{L}^{w} = - \frac{1}{2B} \sum^{2B}_{i=1} \mathcal{L}^{w}_{i,j} = - \frac{1}{2B} \sum^{2B}_{i=1} \log \frac{\exp \left( f_{\texttt{sim}} \left( \mathbf{z}_i, \mathbf{z}_j \right) / \tau \right) }{\sum^{2B}_{k=1}  \mathbbm{1}_{[k \neq i]}\mathbf{W}_{ik} \exp \left(f_{\texttt{sim}} \left(\mathbf{z}_i, \mathbf{z}_k \right) / \tau \right)} \, \, \, .
\end{equation}

{\em Dissimilarity Alignment Loss.}
We compute the mean squared error loss between the cosine similarity of each embedding and its corresponding weight, averaged across all pairs in the batch.
\begin{equation}\label{eqn_dissim_align}
    \mathcal{L}^{d} = \frac{1}{B^2} \sum_{i,j} \left( \frac{1+f_{\texttt{sim}}(\mathbf{z}_i, \mathbf{z}_j)}{2} - (1-\mathbf{W}_{ij}) \right)^2  \, .
\end{equation}
Note that cosine similarity $f_{\texttt{sim}}\!\in\![-1,1]$ is rescaled to $[0,1]$. This encourages the model to map clinically similar ECG signals to nearby embeddings, while pushing dissimilar signals apart. This supports clinical use, where models must highlight differences between ECG signals to distinguish clinical categories (e.g., diagnoses or prognoses). Our final objective is:
$
    \mathcal{L} = \mathcal{L}^{w} + \mathcal{L}^{d}, 
$
where, in this paper, the contribution of the two losses are considered equal (details in \appendixautorefname~\ref{appsubsec:toy_cifar100}).

\section{Experimental Evaluation}
\label{sec:experiment_evaluation}
We evaluate against
(i) existing FMs and models pretrained on ECG data,
(ii) widely-used self-supervised learning algorithms, and
(iii) a state-of-the-art (SOTA) supervised foundation model.

{\bf Foundation Model  $\mathcal{F}$.}
We use ResNeXt1D~\citep{hong2020holmes}\footnote{The PyTorch implementation of the model is adapted from \href{https://github.com/hsd1503/resnet1d}{github.com/hsd1503/resnet1d}.}, built on ResNeXt~\cite{xie2017aggregated}, using one-dimensional convolutional filters, and widely used as a benchmark for ECG processing~\cite{li2024electrocardiogram}. We evaluate three model configurations: small (448K parameters), medium (30.7M parameters), and large (296M parameters), which differ in network depth, width, and complexity. A more detailed model structure description can be found in \appendixautorefname~\ref{appsubsec:backbone_resnet} and \tableautorefname~\ref{apptab:resnet1d_configs}.

{\bf Pretraining Dataset  $\mathbb{D}$.} We train $\mathcal{F}(\cdot)$ using the MIMIC-IV-ECG dataset~\citep{gow2023mimic-iv-ecg}, which contains $161,352$ unique patients, each with multiple ECG recordings. To ensure that signals from the same patient are not considered as negative pairs during model pretraining, we use the first ECG recording of each patient. Each ECG record $\mathbf{e}$ is sampled at $500$Hz over $10$ seconds, resulting in a sequence length of $t=5{,}000$. Following the common practice, we apply a Butterworth bandpass filter between $0.67$ and $40$ Hz, followed by z-score normalization for each sample.

{\bf Stochastic Data Augmentation $\mathcal{T}$.} To simulate real-world signal perturbations, we apply some noise derived from free-living\footnote{The noise data is available at: \href{https://physionet.org/content/ecg-ppg-simulator-arrhythmia/1.3.1/}{physionet.org/content/ecg-ppg-simulator-arrhythmia}.} ECG recordings, including (i) muscle noise, (ii) movement artifacts, (iii) baseline wander, (iv) white noise, or (v) no perturbations. Function $\mathcal{T}$ is a random selection from these five choices, each with equal probability of $p=0.2$. For details, see~\appendixautorefname~\ref{appsubsec:data_augmentation}.

{\bf Random Lead Selection.} Our \acro{} can be fine-tuned to any lead, which is crucial since devices and health-monitoring scenarios vary. Also, many wearables only approximate a lead; \eg, smartwatches approximate lead I, and the target lead may not be known at pretraining time. With this motivation, we do not restrict training to a specific lead as pretraining exclusively on a specific lead (explored in~\sectionautorefname~\ref{sec:results},~\figureautorefname~\ref{fig:spider_pretrain_1lead}) requires one separate model per lead and is less flexible for cross-device use. Instead, for each sample in a batch, we randomly select one of the 12 leads and apply stochastic augmentation, thus every batch contains diverse samples from multiple leads. In this way, the model is collectively trained across all 12 leads, allowing $\mathcal{F}$ to generalize and later be fine-tuned to any lead.

{\bf Downstream Datasets $\mathbb{C}$.} We evaluate on various downstream tasks from cardiovascular conditions, including $3$ well-established benchmarks used in~\citep{li2024electrocardiogram,na2024guiding}, and $4$ newly curated datasets from wearables or emergency visits. Tasks include multi-label diagnostic (e.g. heart block, myocardial infarction, hypertrophy), form (i.e. wave morphology), and rhythm statements on PTB-XL dataset~\citep{wagner2020ptb-xl}, left ventricular ejection fraction (LVEF) regression and classification at 50\% threshold on MIMIC‑IV-ECG dataset~\citep{gow2023mimic-iv-ecg}, and multi-label disease classification on the Chapman dataset~\citep{Zheng2020a}. We also include MC-MED~\citep{kansal2025multimodal} and Aurora BP~\citep{mieloszyk2022comparison} for blood pressure estimation, the MUSIC dataset~\citep{martin2025music} for long-term cardiovascular outcomes such as sudden cardiac death (SCD), and Icentia11K~\citep{tan2019icentia11k}, a large-scale continuous wearable ECG dataset supporting beat and rhythm classification. More details are provided in~\appendixautorefname~\ref{appsec:dataset_details} and~\ref{appsubsec:experiment_setup}, respectively.

{\bf Foundation model baselines.}
(1) ST-MEM~\citep{na2024guiding}: a spatio-temporal masked autoencoder trained to reconstruct randomly masked ECGs.
(2) KED~\citep{tian2024foundation}:  that aligns ECG signals with textual reports, enabling joint ECG–text representations.
(3) Moirai~\citep{woo2024unified}: a forecasting model for time series, supporting both univariate and multivariate prediction.
(4) Moment~\citep{goswami2024moment}: a transformer-based model designed for univariate time series tasks.

{\bf Self-supervised learning baselines}.
(1) SimCLR~\citep{chen2020simple}: a canonical contrastive learning framework, as described in \S\ref{sec:method}, Eq.~\eqref{eqn_ccl}.
(2) BYOL~\citep{grill2020bootstrap}: an augmentation-based approach that eliminates negative pairs by training another network to predict the target network’s representation of the same signal.
(3) MoCo~\citep{he2020momentum}:  maintains a dynamic memory bank to store representations from past batches, reducing the reliance on large in-batch negatives.

{\bf Supervised baseline.}
Concurrent to this work, Harvard–Emory researchers released the HEEDB dataset~\citep{ghanta2025harvard} of 12-lead ECG recordings with $10$M ECGs from $1.8$M patients annotated with $150$ diagnostic categories. While we could not get access to the dataset, they released a supervised ECG FM, ECGFounder~\citep{li2024electrocardiogram}, trained for multi-label classification over the $150$ categories. We run ECGFounder on our evaluation datasets, and since it is trained on a large labeled dataset, we consider it as an upper bound relative to semi-supervised methods, including our proposed~\acro{}.
Implementation details for all baselines are provided in Appendix~\ref{appsubsec:baseline_models} and~\ref{appsubsec:base_pretraining}.

\begin{table}[t]
\renewcommand{\arraystretch}{.92}
    \centering
    \caption{Results for finetuning on lead I (left) and lead II (right) ECGs. AUROCs are reported for $7$ clinical tasks:
    \briefExplain \bestResults}
    \label{tab:performance_single_lead}
    \setlength{\tabcolsep}{1pt}
    \resizebox{\columnwidth}{!}{%
    \small
    \begin{tabular}{lcc ccccc}
    \toprule
     & \multicolumn{7}{c}{\bf Lead I} \\\cmidrule{2-8}
    {Dataset}  & MIMIC-IV & \multicolumn{5}{c}{ PTB-XL} & Chapman \\
    {Task} & LVEF & Dx & SubDx & SupDx & Form & Rhyth & Arrhy \\
    \cmidrule(lr){2-2} \cmidrule(lr){3-7} \cmidrule(lr){8-8}

    Moirai~\citeyear{woo2024unified} (91M)& .4968 & .5003 & .4966 & .5011 & .5158 & .5031 & .4992 \\
        
    Moment~\citeyear{goswami2024moment} (125M)& .7763 & .7780 & .7627 & .7700 & .6322 & .8243 & .8150 \\
    
    ST-MEM~\citeyear{na2024guiding} (85M)& .7751 & .7763 & .7639 & .7800 & .5724 & .7549 & .8113 \\
    KED~\citeyear{tian2024foundation} (8M) & \underline{.8330} & .8390 & .8332 & .8302 & .6696 & .8887 & .8897 \\
    
    \midrule

    \acro-S (448K) & .8170 & .8268 & .8448 & \bf .8452 & .6738 & .9304 & .9033 \\
    \acro-M (30.7M) & .8083 & .8292 & \bf .8566 & \underline{.8430} & \underline{.7409} & \underline{.9361} & \underline{.9061} \\
    \acro-L (296M) & .7858 & \bf .8472 & .8397 & .8273 & .7162 & .9325 & .9010 \\

    \midrule

    ECGFounder~\citeyear{li2024electrocardiogram} (76.3M) & \bf .8512 & \underline{.8457} & \underline{.8500} & .8376 & \bf .7626 & \bf .9501 & \bf .9090 \\

    \bottomrule
    \end{tabular}
    \quad
    \begin{tabular}{cc ccccc}
    \toprule
    \multicolumn{7}{c}{\bf Lead II} \\\cmidrule{1-7}
    MIMIC-IV & \multicolumn{5}{c}{PTB-XL} & Chapman \\
    LVEF & Dx & SubDx & SupDx & Form & Rhyth & Arrhy \\
    \cmidrule(lr){1-1} \cmidrule(lr){2-6} 
    \cmidrule(lr){7-7} 
    .5000 & .5014 & .5001 & .4993 & .5003 & .4988 & .4987 \\
    .8024 & .7896 & .7788 & .8260 & .6164 & .8754 & .8905 \\
    .7499 & .6764 & .7284 & .7637 & .5319 & .7489 & .8228 \\
    .8073 & .8072 & .8304 & .8407 & .6492 & .8736 & .8941 \\
    \midrule
    .8166 & \underline{.8205} & \underline{.8486} & \underline{.8445} & .6913 & .9305 & \underline{.9047} \\
     .8079 & \bf .8307 & \bf .8555 & \bf .8446 & \underline{.7378} & \underline{.9326} & \bf .9089 \\
     \underline{.8194} & .8193 & .8438 & .8409 & \bf .7478 & \bf .9512 & .9011 \\

    \midrule
     \bf .8312 & .8024 & .8244 & .8437 & .6823 & .9229 & .8755 \\
    \bottomrule
    \end{tabular}
    }%
    \vspace{-10pt}
\end{table}

\section{Results}
\label{sec:results}

{\bf Finetuning on Lead I \& II (in-clinic datasets).} The AUROC results are detailed in~\tableautorefname~\ref{tab:performance_single_lead} (with further details in~\appendixautorefname~\ref{appsubsec:confidence_interval}). \acro{} outperforms the best baseline across both leads, with the best-performing~\acro{} variant for each task achieving an average AUROC improvement over the strongest baseline for that task of $3.1\%$ on lead I, and $4.8\%$ on lead II. On lead I, the~\acro-S,~-M, and~-L variants achieve improvements of $1.0\%$, $2.6\%$, and $1.3\%$, respectively, over the best performing baseline ($1.6\%$ on average), while on lead II, gains were even stronger, achieving $2.8\%$, $4.0\%$, and $4.2\%$ ($3.7\%$ average) performance gain. KED was the best-performing baseline in all but one task, and~\acro-M shows statistically significant superiority over KED (p = 0.010, by paired t-test).
~\acro{} outperforms KED by $\geq 1.0\%$ on lead I and $\geq 2.8\%$ on lead II across all $3$ variants, demonstrating its strong potential for pretraining. Additionally, the averaged confidence interval across all~\acro{} variants and tasks is $\pm 0.01$, showcasing~\acro's consistent performance. \acro{} performs particularly well on pattern identification tasks (form and rhythm tasks of PTB-XL, and the arrhythmia task of Chapman), with its best variant improving AUROC by $5.9\%$ (lead I) and $8.5\%$ (lead II) on average over the strongest baselines. These results suggest that~\acro{} excels at capturing both morphological features of individual beats and long-range rhythm dependencies across the signal.

\acro{} maintains robust performance across leads, with~\acro-S,~-M, and~-L differing by $0.3\%$, $0.0\%$, and $1.4\%$ on average in AUROC across $7$ tasks. In contrast, ST-MEM and KED which are specifically pretrained for ECG tasks, favored lead I, with AUROCs on lead II an average of $4.2\%$ and $1.5\%$ lower, respectively. When compared to the supervised model ECGFounder, \acro{} performs less well on lead I (AUROCs lower by $2.9\%$, $1.5\%$, and $2.7\%$ for \acro-S,~-M, and~-L, respectively) but better on lead II (AUROC increases by $1.3\%$, $2.5\%$, and $2.6\%$). This is primarily because ECGFounder's performance drops $3.8\%$ on lead II, compared to lead I (on which the single-lead version of ECGFounder was trained~\citep{li2024electrocardiogram}), while~\acro{} maintains performance across leads. Although the unsupervised~\acro{} did not perform as well as this SOTA supervised model on the lead for which that model was trained, it did achieve improved performance on another lead. Notably, our single-lead~\acro{} even outperforms some baselines when those baselines are pretrained and evaluated on 12-lead ECGs. Specifically,~\acro-S (evaluated on lead I \& II) outperforms ST-MEM on $5$ out of $7$ tasks (Details are reported in~\appendixautorefname~\ref{appsubsec:upperbound12lead} and~\tableautorefname~\ref{tab:performance_12lead}).

\begin{table}[b]
    \centering
    \vspace{-10pt}
    \begin{minipage}[t]{.48\linewidth}
      \caption{Finetuning using single-lead ECGs. AUROCs are for: SCD outcome prediction, beat and rhythm recognition, disposition after emergency department visit (ED dispo) and after hospital stay (DC dispo), and ED triage acuity.
      \bestResults
      }
      \label{tab:performance_wearable}
      \renewcommand{\arraystretch}{0.92}
      \centering
      \setlength{\tabcolsep}{1.9pt}
      \resizebox{\columnwidth}{!}{%
        \begin{tabular}{l c c c c c c}
        \toprule
        \multirow{2}{*}{\bf Model} & \bf MUSIC & \multicolumn{2}{c}{\bf Icentia11K} & \multicolumn{3}{c}{\bf MC-MED} \\
         & SCD & Beat & Rhythm & ED Dispo & DC Dispo & Acuity \\
        \cmidrule(lr){2-2} \cmidrule(lr){3-4} \cmidrule(lr){5-7}
        Moirai & .0106 & .4998 & .4927 & .5004 & .4996 & .4992 \\
        Moment & \bf .6223 & .9764 & .8540 & .5488 & .5042 & .5782 \\
        ST-MEM & .5389 & .9452 & .7115 & .5323 & .5645 & .4943 \\
        KED & .4701 & .9424 & .9228 & .5992 & .6200 & .5978 \\
        
        \midrule
        \acro-S & .5493 & \underline{.9801} & .9135 & .6065 & .6261 & .5650 \\
        \acro-M & .5304 & .9792 & .9328 & \underline{.6397} & \underline{.6607} & \underline{.6510} \\
        \acro-L & \underline{.5545} & .9800 & \underline{.9535} & .5805 & .5800 & .5940 \\
        \midrule
        ECGFounder & .5420 & \bf .9822 & \bf .9723 & \bf .6572 & \bf .6710 & \bf .6690 \\
        \bottomrule
        \end{tabular}
        }%
    \end{minipage}%
    \quad
    \begin{minipage}[t]{.48\linewidth}
    \renewcommand{\arraystretch}{0.92}
      \centering
        \caption{ECG regression tasks. MAEs are reported for LVEF prediction (in $\%$) on MIMIC-IV, and systolic and diastolic blood pressure prediction (in mmHg) on both the MC-MED and Aurora BP data sets. \bestResults}
        \label{tab:performance_regression}
        \setlength{\tabcolsep}{6pt}
        \resizebox{\columnwidth}{!}{%
        \begin{tabular}{lcccccc}
        \toprule
        \multirow{2}{*}{\bf Model} & \bf MIMIC-IV & \multicolumn{2}{c}{\bf Aurora BP} & \multicolumn{2}{c}{\bf MC-MED} \\
        & LVEF & SBP & DBP & SBP & DBP \\
        \cmidrule(lr){2-2} \cmidrule(lr){3-4} \cmidrule(lr){5-6}
        Moirai & 7.405 & 12.587 & 8.261 & 32.881 & 12.704 \\
        Moment & 7.277 & 12.707 & 8.150 & 18.623 & 12.576 \\
        ST-MEM & 7.149 & 12.707 & 8.157 & 18.623 & 12.552 \\
        KED & 8.313 & \underline{11.844} & \bf 8.032 & 18.598 & 12.647 \\
        
        \midrule

        \acro-S & \bf 6.569 & \bf 11.667 & 8.653 & 17.901 & \bf 12.314 \\
        \acro-M & \underline{6.805} & 12.110 & \underline{8.099} & \bf 17.880 & \underline{12.318} \\
        \acro-L & 7.313 & 12.819 & 8.865 & 18.763 & 12.441 \\
        
        \midrule
        ECGFounder & 7.900 & 13.577 & 8.332 & \bf 17.880 & 12.378 \\
        \bottomrule
    \end{tabular}
    }%
    \end{minipage} 
\end{table}

\begin{figure}[b]
    \centering
\includegraphics[width=\linewidth]{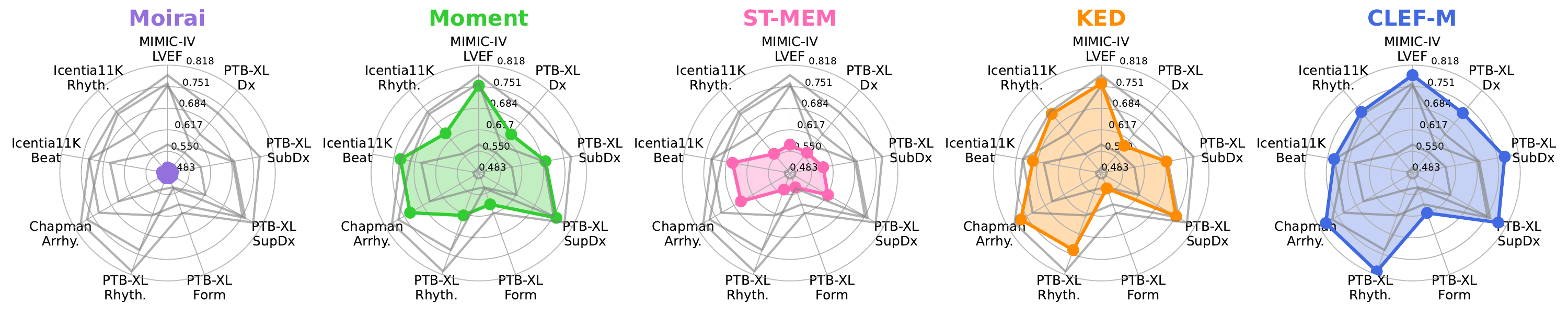}
    \caption{AUROC scores from linear probing on $9$ classification tasks, comparing Moirai, Moment, ST-MEM, KED, and our ~\acro. Each subplot focuses on one model, with others shown in gray for reference. Higher values indicate better performance (see further details in~\tableautorefname~\ref{tab:linear_single_lead_auroc}).}
    \label{fig:spider_plot_linear}
\end{figure}

{\bf Finetuning for Single-lead ECG (out-of-clinic datasets).} We use single-lead ECGs from MUSIC, Icentia11k, and AuroraBP collected by wearables, and MC-MED dataset of Emergency Department.

{\em Classification.} The AUROC results are detailed in \tableautorefname~\ref{tab:performance_wearable} (for classification tasks) and ~\tableautorefname~\ref{tab:performance_regression} (for regression tasks). In classification, \acro{} outperforms all semi-supervised baselines, with the best~\acro{} variant improving AUROC by $2.5\%$ on average over the strongest baseline of the task. The best-performing~\acro{} variant for each task achieves an average AUROC improvement over the strongest baseline for that task of $2.5\%$. Individually, ~\acro-S,~-M, and~-L achieve averaged AUROC improvements of $2.8\%$, $6.7\%$, and $2.5\%$ over the best baseline (KED). On Icentia11K Beat and Rhythm tasks,
the best-performing~\acro{} variant on each task achieves improved AUROC by $1.9\%$ over the best baselines. When looking at each~\acro{} variant,~\acro-S,~-M, and~-L achieve average AUROC improvements of $1.5\%$, $2.5\%$, and $3.7\%$ over the best baseline (KED), highlighting the effectiveness of~\acro{} in capturing quality ECG representations for wearable devices. However, Moment outperformed~\acro{} on MUSIC SCD prediction, the only case where a general time series model surpassed an ECG-specific model; likely because SCD risk depends on long-range and subtle temporal patterns in the ECG, making it more of a forecasting problem than a short-term classification.

{\em Regression.} We evaluate on the MIMIC-IV LVEF task (explored in prior work~\citep{li2024electrocardiogram}), and also systolic and diastolic blood pressure (SBP and DBP) prediction on the MC-MED and AuroraBP datasets. Target labels are first z-scored based on the statistics of the training set, and then at test time, predicted values are transformed back to the original scale of the labels for evaluation. See~\appendixautorefname~\ref{appsubsec:experiment_setup} for details of our experimental setup. The mean absolute error (MAE) is reported in~\tableautorefname~\ref{tab:performance_regression}. The best-performing~\acro{} for each task achieves an average MAE reduction in comparison to the strongest baseline for that task of $2.9\%$. The~\acro-S and~\acro-M~variants outperform all baselines, achieving average MAE reductions of $3.2\%$ compared to the best baseline (ST-MEM). In contrast,~\acro-L outperforms $3$ of the $4$ baselines (yet MAE is $2.3\%$ higher than ST-MEM on average). \acro{} also outperforms the supervised model ECGFounder, with~\acro-S,~-M, and~-L achieving lower MAEs than ECGFounder by $5.5\%$, $5.6\%$, and $0.2\%$ on average.

\begin{table}[t]
\centering
\caption{Comparing self-supervised pretraining methods with AUROCs of single-lead ECG classification. For datasets that originally have $12$ leads, lead I was used for analysis. Instances outperformed by \acro~are highlighted with cells having a grey background.
}
\renewcommand{\arraystretch}{0.9}
\small
\label{tab:pretrained_results}
\setlength{\tabcolsep}{4.5pt}
\resizebox{\textwidth}{!}{%
\begin{tabular}{lcccc ccccc ccc ccc}
\toprule
\multirow{2}{*}{\bf Model} & & \bf MIMIC-IV & \multicolumn{5}{c}{\bf PTB-XL} & \bf Chapman & \bf MUSIC & \multicolumn{2}{c}{\bf Icentia11K} & \multicolumn{3}{c}{\bf MC-MED} \\
& &  LVEF & Dx & SubDx & SupDx & Form & Rhyth & Arrhy & Outcome & Beat & Rhyth & ED Dispo & DC Dispo & Acuity \\
\cmidrule(lr){3-3} \cmidrule(lr){4-8} \cmidrule(lr){9-9} \cmidrule(lr){10-10} \cmidrule(lr){11-12}  \cmidrule(lr){13-15}

\multirow{3}{*}{MOCO~\citeyearpar{he2020momentum}} & S & .8194 & \cellcolor{gray!20}{.5378} & \cellcolor{gray!20}{.5803} & \cellcolor{gray!20}{.7260} & \cellcolor{gray!20}{.5300} & \cellcolor{gray!20}{.5128} & \cellcolor{gray!20}{.7743} & \cellcolor{gray!20}{.5477} & \cellcolor{gray!20}{.9133} & \cellcolor{gray!20}{.4737} & 0.6111 & \cellcolor{gray!20}{0.6075} & \cellcolor{gray!20}{0.5518} \\
 & M & \cellcolor{gray!20}{.8112} & \cellcolor{gray!20}{.7664} & \cellcolor{gray!20}{.7885} & \cellcolor{gray!20}{.7757} & \cellcolor{gray!20}{.5021} & \cellcolor{gray!20}{.5163}& \cellcolor{gray!20}{.8119} & \cellcolor{gray!20}{.4546} & \cellcolor{gray!20}{.9764} & \cellcolor{gray!20}{.7115}  & \cellcolor{gray!20}{0.5926} & \cellcolor{gray!20}{0.5963} & \cellcolor{gray!20}{0.5659} \\
 & L & \cellcolor{gray!20}{.6994} & \cellcolor{gray!20}{.5014} & \cellcolor{gray!20}{.4981} & \cellcolor{gray!20}{.5060} & \cellcolor{gray!20}{.5063} & \cellcolor{gray!20}{.4990} & \cellcolor{gray!20}{.5258} & \cellcolor{gray!20}{.4563} & \cellcolor{gray!20}{.8680} & \cellcolor{gray!20}{.5697} & \cellcolor{gray!20}{0.5331} & \cellcolor{gray!20}{0.4865} & \cellcolor{gray!20}{0.4781} \\
\midrule
\multirow{3}{*}{SimCLR~\citeyearpar{chen2020simple}} & S & \cellcolor{gray!20}{.8185} & \cellcolor{gray!20}{.7923} & \cellcolor{gray!20}{.7790} & \cellcolor{gray!20}{.8180} & \cellcolor{gray!20}{.5560} & \cellcolor{gray!20}{.8979} & \cellcolor{gray!20}{.8554} & \cellcolor{gray!20}{.5485} & \cellcolor{gray!20}{.9722} & .9180 & 0.6150 & 0.6376 & 0.6135 \\
 & M & .8243 & \cellcolor{gray!20}{.8247} & \cellcolor{gray!20}{.8352} & \cellcolor{gray!20}{.8210} & \cellcolor{gray!20}{.7338} & \cellcolor{gray!20}{.9248} & \cellcolor{gray!20}{.8888} & .5531 & \cellcolor{gray!20}{.9756} & \cellcolor{gray!20}{.8936} & \cellcolor{gray!20}{0.6380} & \cellcolor{gray!20}{0.6553} & \cellcolor{gray!20}{0.5743} \\
 & L & .8192 & \cellcolor{gray!20}{.8393} & \cellcolor{gray!20}{.8370} & \cellcolor{gray!20}{.8279} & \cellcolor{gray!20}{.7137} & \cellcolor{gray!20}{.9109} & \cellcolor{gray!20}{.8945} & \cellcolor{gray!20}{.4799} & \cellcolor{gray!20}{.9733} & \cellcolor{gray!20}{.9087} & 0.5984 & \cellcolor{gray!20}{0.6129} & 0.5988 \\
 \midrule
 \multirow{3}{*}{BYOL~\citeyearpar{grill2020bootstrap}} & S & \cellcolor{gray!20}{.8147} & \cellcolor{gray!20}{.8134} & \cellcolor{gray!20}{.8282} & \cellcolor{gray!20}{.8259} & \cellcolor{gray!20}{.6883} & \cellcolor{gray!20}{.9296} & \cellcolor{gray!20}{.8789} & \cellcolor{gray!20}{.5448} & \cellcolor{gray!20}{.9729} & \cellcolor{gray!20}{.9065} & 0.6166 & \cellcolor{gray!20}{0.6177} & 0.6188 \\
 & M & .8221 & .8421 & \cellcolor{gray!20}{.8221} & \cellcolor{gray!20}{.8251} & \cellcolor{gray!20}{.7215} & \cellcolor{gray!20}{.9351} & \cellcolor{gray!20}{.8917} & \cellcolor{gray!20}{.4713} & \cellcolor{gray!20}{.9766} & \cellcolor{gray!20}{.8919} & \cellcolor{gray!20}{0.6306} & \cellcolor{gray!20}{0.6447} & \cellcolor{gray!20}{0.5643} \\
 & L & .8221 & \cellcolor{gray!20}{.8409} & \cellcolor{gray!20}{.8378} & \cellcolor{gray!20}{.8239} & \cellcolor{gray!20}{.7144} & \cellcolor{gray!20}{.9257} & \cellcolor{gray!20}{.8978} & \cellcolor{gray!20}{.4719} & .9819 & \cellcolor{gray!20}{.9032} & \cellcolor{gray!20}{0.5515} & \cellcolor{gray!20}{0.5588} & \cellcolor{gray!20}{0.5147} \\
\bottomrule
\end{tabular}
}%
\vspace{-10pt}
\end{table}
\begin{figure}[b!]
    \centering
    \begin{minipage}[b]{0.32\textwidth}
        \centering
        \includegraphics[width=\textwidth]{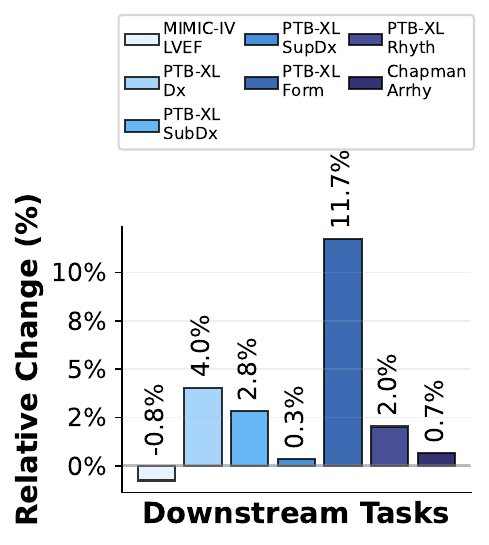}
        \caption{Changes in AUROC of KED after further training with \acro{} objectives across downstream ECG tasks.}
        \label{fig:pretrain_ked}
    \end{minipage}
    \hfill
    \begin{minipage}[b]{0.63\textwidth}
        \centering
        \includegraphics[width=\textwidth]{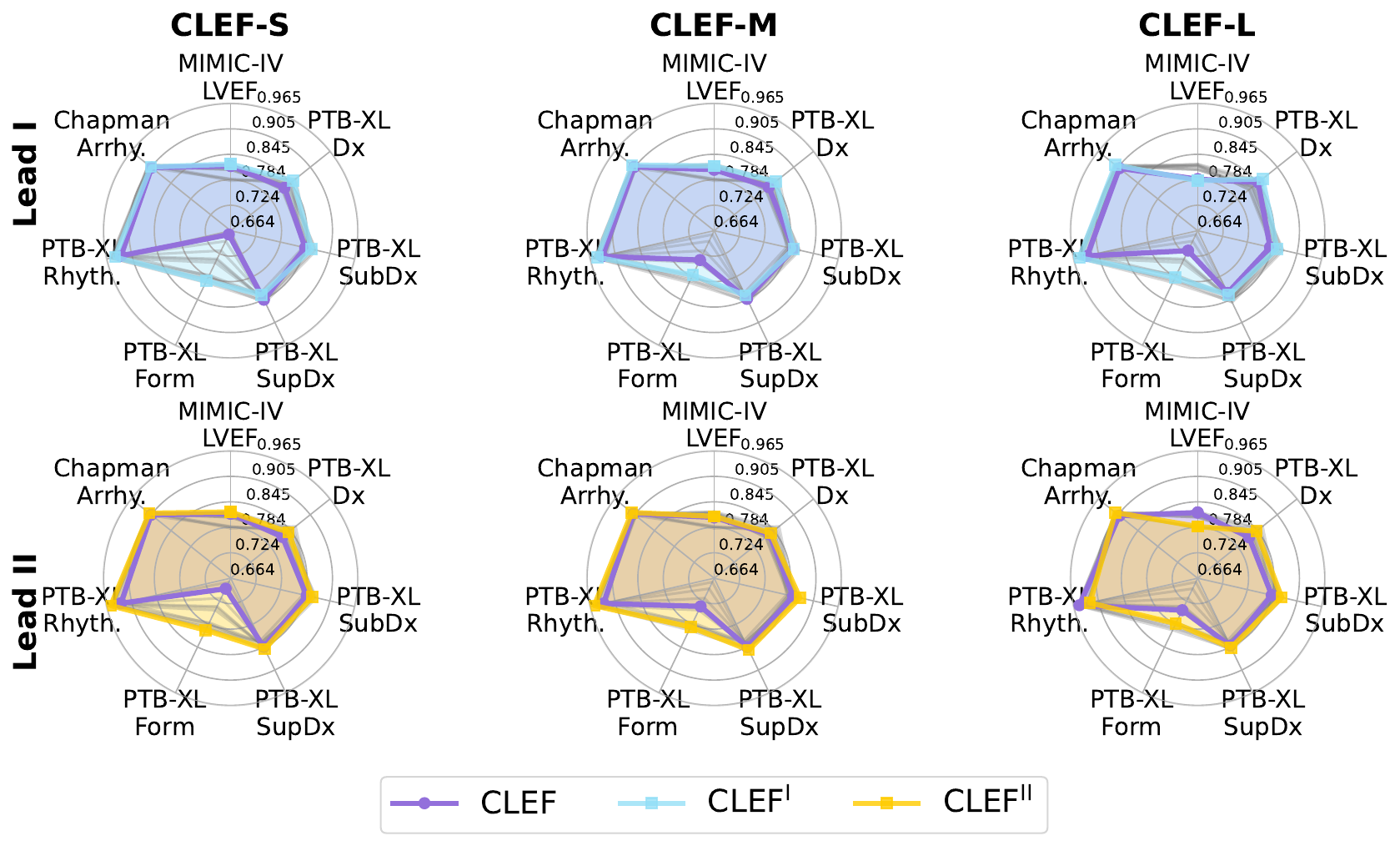}
        \caption{Spider plot comparing~\acro{} with~\acro{} model pretrained on a specific lead (\acro$^\texttt{I}$ and~\acro$^\texttt{II}$). AUROC is reported across $3$ model variants and $7$ downstream tasks.}
        \label{fig:spider_pretrain_1lead}
    \end{minipage}
\end{figure}
{\bf Linear probing.}
To evaluate the quality of the representations produced by~\acro{}, the parameters of FMs were kept frozen, and a linear classifier was trained on the output embeddings. The AUROCs are summarized in~\figureautorefname~\ref{fig:spider_plot_linear} for~\acro-M~and $4$ baselines. Full results including those for the other $2$ variants (\acro-S and~-L) are provided in~\tableautorefname~\ref{tab:linear_single_lead_auroc} (\appendixautorefname~\ref{appsubsec:linear_probe}). The best-performing \acro{} for each task outperforms the strongest baseline for that task of +$7.3\%$.~\acro-M and~\acro-L both outperform all baselines, with average AUROC improvements of $8.5\%$ and $9.9\%$ respectively over the best baseline (KED). However,~\acro-S does not outperform KED or Moment, which indicates that the larger~\acro{} models produce better quality representations. Models pretrained specifically for ECG tasks do not always yield better performance.  Moment (which~\acro-M outperforms by $10.0\%$) performs better than ST-MEM (which~\acro{} outperforms by $28.1\%$).

{\bf Comparison with Self-supervised Algorithms.} Our proposed clinically-guided contrastive learning method is compared against $3$ widely used self-supervised pretraining methods: SimCLR, BYOL, and MoCo. Each pretraining method is trained on all $3$ size variants of~\acro. The AUROC results for the comparator methods are presented in~\tableautorefname~\ref{tab:pretrained_results}. A total of $100$ out of $117$ instances were outperformed by~\acro{} (highlighted in gray). Overall,~\acro{} outperforms all $3$ self-supervised pretraining methods, with average AUROC improvements across all sizes of backbone models of $29.8\%$, $1.8\%$, and $2.3\%$ in comparison to MOCO, SimCLR, and BYOL, respectively.

{\bf Our Clinically-guided Contrastive Loss with Best Semi-supervised Baseline.} We initialize the best-performing baseline for ECG, \ie KED, with its original pretrained weights and apply our proposed clinically-guided pretraining approach. AUROC results are presented in~\figureautorefname~\ref{fig:pretrain_ked}. It can be observed that, apart from the LVEF classification task in MIMIC-IV, all tasks gained improvement, on average $3.0\%$ higher AUROC. 
The form task in PTB-XL gets the greatest improvement ($11.7\%$). More results on pretraining other ECG baseline models are provided in~\appendixautorefname~\ref{appsubsec:pretrain_over_baseline}.

{\bf Pretraining \acro{} on a Specific Lead.} To assess the potential upper bound of \acro{}, 
we pretrain models exclusively on the corresponding lead of the downstream tasks (lead I or II). Pretraining was performed for $10$ epochs, with the other downstream finetuning hyperparameters the same as the previous experiments. Results are summarized in \figureautorefname~\ref{fig:spider_pretrain_1lead}, and provided in full in \appendixautorefname~\ref{appsubsec:supp_pretraining_1lead}. On lead I, pretraining achieves average AUROC improvements over pretraining using all 12 leads of $3.4\%$, $1.4\%$, and $2.4\%$ for~\acro-S,~-M, and~-L, respectively. Moreover, this made the performance of~\acro{} comparable to that of the supervised ECGFounder model, with differences in average AUROC of only $0.2\%$, $-0.1\%$, and $-0.3\%$ for~\acro-S,~-M, and~-L, respectively, on lead I (the same lead ECGFounder is trained on too). Improvements are also observed when pretraining on lead II, of $3.4\%$, $2.0\%$, and $0.8\%$ for~\acro-S,~-M, and~-L, respectively.

\begin{table}[t]
\caption{Ablation results on handling missing metadata across~\acro~models of different sizes, where $\neg \mathbf{M}$ denotes the model trained without handling the missing metadata. We report both the results from the ablated models, and the performance changes relative to the corresponding \acro~models (shown in brackets). The $\uparrow$ and $\downarrow$ depict performance increase and drop, respectively. For better clarity, we highlighted cases where performance is inferior in \textcolor{blue}{blue}.}
    \label{tab:ablation_missing_metadata}
    \renewcommand{\arraystretch}{0.92}
    \centering
    \setlength{\tabcolsep}{1pt}
    \resizebox{\columnwidth}{!}{%
    \small
    \begin{tabular}{lcc ccccc ccc}
    \toprule
    \bf Task & \bf MIMIC-IV & \multicolumn{5}{c}{\bf PTB-XL} & \bf Chapman & \bf MUSIC & \multicolumn{2}{c}{\bf Icentia11K} \\
    &  LVEF & Dx & SubDx & SupDx & Form & Rhyth & Arrhy & Outcome & Beat & Rhyth \\
    \cmidrule(lr){2-2} \cmidrule(lr){3-7} \cmidrule(lr){8-8}  \cmidrule(lr){9-9} \cmidrule(lr){10-11}
    
    \acro-S $\neg \mathbf{M}$ 
    & .8245 ($\uparrow$ .01) 
    & .8152 (\textcolor{blue}{$\downarrow$ .01}) 
    & .8266 (\textcolor{blue}{$\downarrow$ .02}) 
    & .8210 (\textcolor{blue}{$\downarrow$ .03}) 
    & .6810 ($\uparrow$ .01) 
    & .9362 ($\uparrow$ .01) 
    & .8786 (\textcolor{blue}{$\downarrow$ .03}) 
    & .4763 (\textcolor{blue}{$\downarrow$ .13}) 
    & .9637 (\textcolor{blue}{$\downarrow$ .02}) 
    & .9052 (\textcolor{blue}{$\downarrow$ .01}) \\
    
    \acro-M $\neg \mathbf{M}$ 
    & .8191 ($\uparrow$ .01) 
    & .8268 (\textcolor{blue}{$\downarrow$ .00}) 
    & .8311 (\textcolor{blue}{$\downarrow$ .03}) 
    & .8245 (\textcolor{blue}{$\downarrow$ .02}) 
    & .7252 (\textcolor{blue}{$\downarrow$ .02}) 
    & .9230 (\textcolor{blue}{$\downarrow$ .01}) 
    & .8839 (\textcolor{blue}{$\downarrow$ .02}) 
    & .5513 ($\uparrow$ .04) 
    & .9749 (\textcolor{blue}{$\downarrow$ .00}) 
    & .8856 (\textcolor{blue}{$\downarrow$ .05}) \\
    
    \acro-L $\neg \mathbf{M}$ 
    & .7786 (\textcolor{blue}{$\downarrow$ .01}) 
    & .8299 (\textcolor{blue}{$\downarrow$ .02}) 
    & .8502 ($\uparrow$ .01) 
    & .8257 (\textcolor{blue}{$\downarrow$ .02}) 
    & .7024 (\textcolor{blue}{$\downarrow$ .02}) 
    & .9201 (\textcolor{blue}{$\downarrow$ .01}) 
    & .8939 (\textcolor{blue}{$\downarrow$ .01}) 
    & .4966 (\textcolor{blue}{$\downarrow$ .10}) 
    & .9757 (\textcolor{blue}{$\downarrow$ .00}) 
    & .8835 (\textcolor{blue}{$\downarrow$ .07}) \\
    
    \bottomrule
    \end{tabular}
    }%
\end{table}

{\bf Ablation on Handling Missing Metadata.} We assess the impact of our solution for handling missing metadata. As an ablation study, we report AUROC for all \acro{} variants across $10$ tasks including wearable datasets (MUSIC and Icentia11K) and single-lead ECG tasks (MIMIC-IV, PTB-XL, and Chapman) using lead I. \tableautorefname~\ref{tab:ablation_missing_metadata} shows that without $\mathbf{M}$ in Eq.~\eqref{eqn_missing_meta}, the performance of all models degrades on the majority of tasks. On average, AUROC decreases by $1.9\%$, with a larger decrease on wearable datasets ($4.0\%$ drop) compared to lead I datasets ($1.1\%$ drop).

{\bf Contribution of different loss components.} A study is conducted using CIFAR-100 to understand the contribution of different loss components. All models are initialized with ImageNet-1K pretrained ResNet-18 weights and further pretrained with: the standard contrastive loss $\mathcal{L}_{nce}$, the weighted contrastive loss $\mathcal{L}^w$, the dissimilarity alignment loss $\mathcal{L}^d$, $\mathcal{L}_{nce} + \mathcal{L}^{d}$, and our proposed $\mathcal{L}^w + \mathcal{L}^{d}$. A detailed experiment setup is provided in~\appendixautorefname~\ref{appsubsec:toy_cifar100}. Our proposed $\mathcal{L}^w + \mathcal{L}^d$ achieves clearer separation with lower variance. Further details in~\appendixautorefname~\ref{appsubsec:toy_cifar100},~\figureautorefname~\ref{fig:cifar_summ}, and~\tableautorefname~\ref{tab:cifar100}).

\section{Discussion and Conclusion}

We propose \acro{}: clinically-guided contrastive learning to train single-lead ECG foundation models. On average, \acro{} outperforms all baseline semi-supervised FMs across $18$ clinical classification and regression tasks on $7$ datasets. Furthermore, \acro~outperforms three self-supervised pretraining algorithms. When pretrained on the same lead as in downstream tasks, \acro~also performs comparably to ECGFounder, a state-of-the-art FM trained in a supervised manner on a labeled dataset. \acro~facilitates the creation of high-performance ECG FMs using only routinely-recorded metadata, without needing ECG-level annotations.

{\bf Limitations.}
We trained models on ECGs from a single hospital, which suit hospital-related tasks but are not representative of the general population. Performance on population-level tasks would likely improve with a more diverse cohort. The risk scores in this study are limited by only three out of seven SCORE2 input variables being available. Therefore, the scores cannot be considered a holistic assessment of cardiovascular risk, but instead are representative of real-world applications with incomplete metadata. Potentially richer embeddings could be obtained when using more complete metadata and, therefore, more precise risk scores. Finally, the single-lead ECGs used to assess performance in this study were measured using wet gel electrodes at the chest, and so are likely of higher quality than those typically measured by devices such as smartwatches and handheld ECGs.

{\bf Future work.}
One can investigate the utility of the proposed contrastive learning strategy for other modalities and health conditions. For instance, it may also be useful for developing photoplethysmography (PPG) FMs, since the PPG is also a cardiovascular signal and has been found to be associated with cardiovascular risk~\cite{wengPredicting2024}. It could also be applied to other health conditions, such as incorporating the risk of deterioration in chronic respiratory conditions, in contrastive learning for respiratory signals. The practical application of this strategy would be aided by understanding which metadata variables contribute most to the quality of the representations produced by \acro. A further ablation study, preferably conducted on a dataset with complete metadata, could identify the most important metadata variables and provide insight into how to apply this approach across diverse datasets with different metadata.
Finally, we can go beyond traditional classification and regression tasks. Because the \acro{}’s embedding space enables patient ECGs to be analyzed without prior metadata. It is possible to assess the proximity of current embeddings to prototypes of high-risk patients from the training set. This opens the door to clustering-based decision-making and other novel applications.

\bibliographystyle{abbrvnat}
\bibliography{main}

@article{friedman2024electrocardiogram,
  title={The electrocardiogram at 100 years: history and future},
  author={Friedman, Paul A},
  journal={Circulation},
  volume={149},
  number={6},
  pages={411--413},
  year={2024},
}

@article{yao2021artificial,
  title={Artificial intelligence--enabled electrocardiograms for identification of patients with low ejection fraction: a pragmatic, randomized clinical trial},
  author={Yao, Xiaoxi and Rushlow, David R and Inselman, Jonathan W and McCoy, Rozalina G and Thacher, Thomas D and Behnken, Emma M and Bernard, Matthew E and Rosas, Steven L and Akfaly, Abdulla and Misra, Artika and others},
  journal={Nature medicine},
  volume={27},
  number={5},
  pages={815--819},
  year={2021},
}

@article{lipRefining2010,
  title = {Refining {{Clinical Risk Stratification}} for {{Predicting Stroke}} and {{Thromboembolism}} in {{Atrial Fibrillation Using}} a {{Novel Risk Factor-Based Approach}}: {{The Euro Heart Survey}} on {{Atrial Fibrillation}}},
  author = {Lip, Gregory Y. H. and Nieuwlaat, Robby and Pisters, Ron and Lane, Deirdre A. and Crijns, Harry J. G. M.},
  year = {2010},
  journal = {CHEST},
  volume = {137},
  number = {2},
  pages = {263--272},
}

@article{xu2025lsm,
  title={LSM-2: Learning from Incomplete Wearable Sensor Data},
  author={Xu, Maxwell A and Narayanswamy, Girish and Ayush, Kumar and Spathis, Dimitris and Liao, Shun and Tailor, Shyam A and Metwally, Ahmed and Heydari, A Ali and Zhang, Yuwei and Garrison, Jake and others},
  journal={arXiv preprint arXiv:2506.05321},
  year={2025}
}

@article{miller2025wearable,
  title={A wearable-based aging clock associates with disease and behavior},
  author={Miller, Andrew C and Futoma, Joseph and Abbaspourazad, Salar and Heinze-Deml, Christina and Emrani, Saba and Shapiro, Ian and Sapiro, Guillermo},
  journal={Nature communications},
  volume={16},
  number={1},
  pages={9264},
  year={2025},
  publisher={Nature Publishing Group UK London}
}

@article{attia2022prospective,
  title={Prospective evaluation of smartwatch-enabled detection of left ventricular dysfunction},
  author={Attia, Zachi I and Harmon, David M and Dugan, Jennifer and Manka, Lukas and Lopez-Jimenez, Francisco and Lerman, Amir and Siontis, Konstantinos C and Noseworthy, Peter A and Yao, Xiaoxi and Klavetter, Eric W and others},
  journal={Nature medicine},
  volume={28},
  number={12},
  pages={2497--2503},
  year={2022},
}

@article{sun2022ecg,
  title={ECG for high-throughput screening of multiple diseases: {P}roof-of-concept using multi-diagnosis deep learning from population-based datasets},
  author={Sun, Weijie and Kalmady, Sunil Vasu and Salimi, Amir and Sepehrvand, Nariman and Ly, Eric and Hindle, Abram and Greiner, Russell and Kaul, Padma},
  journal={arXiv preprint arXiv:2210.06291},
  year={2022}
}

@article{kim2024deep,
  title={Deep learning for predicting rehospitalization in acute heart failure: {M}odel foundation and external validation},
  author={Kim, Mi-Na and Lee, Yong Seok and Park, Youngmin and Jung, Ayoung and So, Hanjee and Park, Joonwoong and Park, Jin-Joo and Choi, Dong-Joo and Kim, So-Ree and Park, Seong-Mi},
  journal={ESC Heart Failure},
  volume={11},
  number={6},
  pages={3702--3712},
  year={2024},
}

@inproceedings{kiyasseh2021clocs,
  title={Clocs: {C}ontrastive learning of cardiac signals across space, time, and patients},
  author={Kiyasseh, Dani and Zhu, Tingting and Clifton, David A},
  booktitle={International Conference on Machine Learning},
  pages={5606--5615},
  year={2021},
  organization={PMLR}
}

@inproceedings{oh2022lead,
  title={Lead-agnostic self-supervised learning for local and global representations of electrocardiogram},
  author={Oh, Jungwoo and Chung, Hyunseung and Kwon, Joon-myoung and Hong, Dong-gyun and Choi, Edward},
  booktitle={Conference on Health, Inference, and Learning},
  pages={338--353},
  year={2022},
  organization={PMLR}
}

@article{ramirez2024art,
  title={The art of selecting the {ECG} input in neural networks to classify heart diseases: a dual focus on maximizing information and reducing redundancy},
  author={Ramirez, Elisa and Ruiperez-Campillo, Samuel and Casado-Arroyo, Ruben and Merino, Jos{\'e} Luis and Vogt, Julia E and Castells, Francisco and Millet, Jos{\'e}},
  journal={Frontiers in physiology},
  volume={15},
  pages={1452829},
  year={2024},
}

@article{angelaki2025diagnostic,
  title={Diagnostic performance of single-lead electrocardiograms for arterial hypertension diagnosis: a machine learning approach},
  author={Angelaki, Eleni and Barmparis, Georgios D and Fragkiadakis, Konstantinos and Maragkoudakis, Spyros and Zacharis, Evangelos and Plevritaki, Anthi and Kampanieris, Emmanouil and Kalomoirakis, Petros and Kassotakis, Spyros and Kochiadakis, George and others},
  journal={Journal of Human Hypertension},
  volume={39},
  number={1},
  pages={58--65},
  year={2025},
}

@article{jimenez202212,
  title={From 12 to 1 {ECG} lead: multiple cardiac condition detection mixing a hybrid machine learning approach with a one-versus-rest classification strategy},
  author={Jim{\'e}nez-Serrano, Santiago and Rodrigo, Miguel and Calvo, Conrado J and Millet, Jos{\'e} and Castells, Francisco},
  journal={Physiological Measurement},
  volume={43},
  number={6},
  pages={064003},
  year={2022},
}

@inproceedings{li2023rethinking,
title={{Rethinking Negative Pairs in Code Search}},
author={Haochen Li and Xin Zhou and Anh Tuan Luu and Chunyan Miao},
booktitle={The 2023 Conference on Empirical Methods in Natural Language Processing},
year={2023},
}

@inproceedings{zhuang2024not,
  title={Not all negatives are equally negative: {S}oft contrastive learning for unsupervised sentence representations},
  author={Zhuang, Haojie and Emma Zhang, Wei and Yang, Jian and Chen, Weitong and Sheng, Quan Z},
  booktitle={Proceedings of the ACM International Conference on Information and Knowledge Management},
  pages={3591--3601},
  year={2024}
}

@article{ghanta2025harvard,
  author       = {Ghanta, Manohar and Sameni, Reza and Aguirre, Aaron and Li, Qiao and Zafar, Sahar and Clifford, Gari and Westover, M Brandon},
  title        = {{Harvard-Emory ECG Database}},
  year         = {2025},
  version      = {4.0},
  fdoi          = {10.60508/rv6h-7d10},
  furl          = {https://doi.org/10.60508/rv6h-7d10}
}

@inproceedings{xie2017aggregated,
  title={Aggregated residual transformations for deep neural networks},
  author={Xie, Saining and Girshick, Ross and Doll{\'a}r, Piotr and Tu, Zhuowen and He, Kaiming},
  booktitle={Proceedings of the IEEE conference on computer vision and pattern recognition},
  pages={1492--1500},
  year={2017}
}

@inproceedings{robinson2021contrastive,
title={{Contrastive Learning with Hard Negative Samples}},
author={Joshua David Robinson and Ching-Yao Chuang and Suvrit Sra and Stefanie Jegelka},
booktitle={International Conference on Learning Representations},
year={2021},
}

@article{yang2024does,
  title={Does negative sampling matter? a review with insights into its theory and applications},
  author={Yang, Zhen and Ding, Ming and Huang, Tinglin and Cen, Yukuo and Song, Junshuai and Xu, Bin and Dong, Yuxiao and Tang, Jie},
  journal={IEEE Transactions on Pattern Analysis and Machine Intelligence},
  volume={46},
  number={8},
  pages={5692--5711},
  year={2024},
}

@article{mehari2022self-supervised,
    title        = {{Self-supervised representation learning from 12-lead ECG data}},
    author       = {Mehari, Temesgen and Strodthoff, Nils},
    year         = 2022,
    journal      = {Computers in biology and medicine},
    volume       = 141,
    pages        = 105114
}

@inproceedings{na2024guiding,
    title        = {{Guiding Masked Representation Learning to Capture Spatio-Temporal Relationship of Electrocardiogram}},
    author       = {Yeongyeon Na and Minje Park and Yunwon Tae and Sunghoon Joo},
    year         = 2024,
    booktitle    = {International Conference on Learning Representations}
}

@article{li2024electrocardiogram,
    title        = {{An Electrocardiogram Foundation Model Built on over 10 Million Recordings with External Evaluation across Multiple Domains}},
    author       = {Li, Jun and Aguirre, Aaron and Moura, Junior and Liu, Che and Zhong, Lanhai and Sun, Chenxi and Clifford, Gari and Westover, Brandon and Hong, Shenda},
    year         = 2024,
    journal      = {arXiv preprint arXiv:2410.04133}
}

@article{khamisQRS2016,
  title = {{{QRS} Detection Algorithm for Telehealth Electrocardiogram Recordings}},
  author = {Khamis, Heba and Weiss, Robert and Xie, Yang and Chang, Chan-Wei and Lovell, Nigel H. and Redmond, Stephen J.},
  year = {2016},
  journal = {IEEE Transactions on Biomedical Engineering},
  volume = {63},
  number = {7},
  pages = {1377--1388},
  fdoi = {10.1109/TBME.2016.2549060},
}

@article{halvaeiIdentification2021,
  title = {{Identification of Transient Noise to Reduce False Detections in Screening for Atrial Fibrillation}},
  author = {Halvaei, Hesam and Svennberg, Emma and S{\"o}rnmo, Leif and Stridh, Martin},
  year = {2021},
  journal = {Frontiers in Physiology},
  volume = {12},
  fdoi = {10.3389/fphys.2021.672875}
}

@article{svennbergMass2015,
  title = {{Mass Screening for Untreated Atrial Fibrillation}},
  author = {Svennberg, Emma and Engdahl, Johan and {Al-Khalili}, Faris and Friberg, Leif and Frykman, Viveka and Rosenqvist, M{\aa}rten},
  year = {2015},
  journal = {Circulation},
  volume = {131},
  number = {25},
  pages = {2176--2184},
  fdoi = {10.1161/CIRCULATIONAHA.114.014343},
}

@article{hannunCardiologistLevel2019a,
  title = {{Cardiologist-Level Arrhythmia Detection and Classification in Ambulatory Electrocardiograms Using a Deep Neural Network}},
  author = {Hannun, Awni Y. and Rajpurkar, Pranav and Haghpanahi, Masoumeh and Tison, Geoffrey H. and Bourn, Codie and Turakhia, Mintu P. and Ng, Andrew Y.},
  year = {2019},
  journal = {Nature medicine},
  volume = {25},
  number = {1},
  pages = {65--69},
  fdoi = {10.1038/s41591-018-0268-3},
}

@article{wagner2020ptb-xl,
    title        = {{PTB-XL, a large publicly available electrocardiography dataset}},
    author       = {Wagner, Patrick and Strodthoff, Nils and Bousseljot, Ralf-Dieter and Kreiseler, Dieter and Lunze, Fatima I. and Samek, Wojciech and Schaeffter, Tobias},
    year         = 2020,
    journal      = {Scientific Data},
    volume       = 7,
    number       = 1,
    pages        = 154
}

@article{wengPredicting2024,
  title = {{Predicting Cardiovascular Disease Risk Using Photoplethysmography and Deep Learning}},
  author = {Weng, Wei-Hung and Baur, Sebastien and Daswani, Mayank and Chen, Christina and Harrell, Lauren and Kakarmath, Sujay and Jabara, Mariam and Behsaz, Babak and McLean, Cory Y. and Matias, Yossi and Corrado, Greg S. and Shetty, Shravya and Prabhakara, Shruthi and Liu, Yun and Danaei, Goodarz and Ardila, Diego},
  year = {2024},
  journal = {PLOS Global Public Health},
  volume = {4},
  number = {6},
  pages = {e0003204},
  fdoi = {10.1371/journal.pgph.0003204},
}

@article{tian2024foundation,
    title        = {{Foundation model of ECG diagnosis: Diagnostics and explanations of any form and rhythm on ECG}},
    author       = {Yuanyuan Tian and Zhiyuan Li and Yanrui Jin and Mengxiao Wang and Xiaoyang Wei and Liqun Zhao and Yunqing Liu and Jinlei Liu and Chengliang Liu},
    year         = 2024,
    journal      = {Cell Reports Medicine},
    volume       = 5,
    number       = 12,
    pages        = 101875
}

@inproceedings{liu2024zero-shot,
    title        = {{Zero-Shot {ECG} Classification with Multimodal Learning and Test-time Clinical Knowledge Enhancement}},
    author       = {Che Liu and Zhongwei Wan and Cheng Ouyang and Anand Shah and Wenjia Bai and Rossella Arcucci},
    year         = 2024,
    booktitle    = {International Conference on Machine Learning}
}

@inproceedings{jin2025reading,
    title        = {{Reading Your Heart: Learning {ECG} Words and Sentences via Pre-training {ECG} Language Model}},
    author       = {Jiarui Jin and Haoyu Wang and Hongyan Li and Jun Li and Jiahui Pan and Shenda Hong},
    year         = 2025,
    booktitle    = {International Conference on Learning Representations}
}

@article{mckeen2024ecg-fm,
    title        = {{Ecg-fm: An open electrocardiogram foundation model}},
    author       = {McKeen, Kaden and Oliva, Laura and Masood, Sameer and Toma, Augustin and Rubin, Barry and Wang, Bo},
    year         = 2024,
    journal      = {arXiv preprint arXiv:2408.05178}
}

@article{wang2024anyecg,
    title        = {{AnyECG: Foundational Models for Electrocardiogram Analysis}},
    author       = {Wang, Yue and Cao, Xu and Hu, Yaojun and Ying, Haochao and Rehg, James Matthew and Sun, Jimeng and Wu, Jian and Chen, Jintai},
    year         = 2024,
    journal      = {arXiv preprint arXiv:2411.17711}
}

@article{thoma2020geometrically,
    title        = {{Geometrically Mappable Image Features}},
    author       = {Thoma, Janine and Paudel, Danda Pani and Chhatkuli, Ajad and Gool, Luc Van},
    year         = 2020,
    journal      = {IEEE Robotics and Automation Letters},
    volume       = 5,
    number       = 2,
    pages        = {2062--2069}
}

@inproceedings{thoma2020soft,
    title        = {{Soft Contrastive Learning for Visual Localization}},
    author       = {Thoma, Janine and Paudel, Danda Pani and Gool, Luc V},
    year         = 2020,
    booktitle    = {Advances in Neural Information Processing Systems},
    publisher    = {Curran Associates, Inc.},
    volume       = 33,
    pages        = {11119--11130}
}

@inproceedings{chen2020simple,
    title        = {A Simple Framework for Contrastive Learning of Visual Representations},
    author       = {Chen, Ting and Kornblith, Simon and Norouzi, Mohammad and Hinton, Geoffrey},
    year         = 2020,
    booktitle    = {International Conference on Machine Learning},
    volume       = 119,
    pages        = {1597--1607}
}

@inproceedings{haslum2024metadata,
    title        = {Metadata-guided Consistency Learning for High Content Images},
    author       = {Haslum, Johan Fredin and Matsoukas, Christos and Leuchowius, Karl-Johan and M\"ullers, Erik and Smith, Kevin},
    year         = 2024,
    booktitle    = {Medical Imaging with Deep Learning},
    volume       = 227,
    pages        = {918--936}
}

@article{kim2025adaptive,
    title        = {Adaptive Metadata-Guided Supervised Contrastive Learning for Domain Adaptation on Respiratory Sound Classification},
    author       = {Kim, June-Woo and Toikkanen, Miika and Jalali, Amin and Kim, Minseok and Han, Hye-Ji and Kim, Hyunwoo and Shin, Wonwoo and Jung, Ho-Young and Kim, Kyunghoon},
    year         = 2025,
    journal      = {IEEE Journal of Biomedical and Health Informatics},
    pages        = {1--13}
}

@article{foxPrediction2006,
  title = {{Prediction of Risk of Death and Myocardial Infarction in the Six Months after Presentation with Acute Coronary Syndrome: Prospective Multinational Observational Study ({{GRACE}})}},
  author = {Fox, Keith A. A. and Dabbous, Omar H. and Goldberg, Robert J. and Pieper, Karen S. and Eagle, Kim A. and de Werf, Frans Van and Avezum, {\'A}lvaro and Goodman, Shaun G. and Flather, Marcus D. and Anderson, Frederick A. and Granger, Christopher B.},
  year = {2006},
  journal = {BMJ},
  volume = {333},
  number = {7578},
  pages = {1091},
  fdoi = {10.1136/bmj.38985.646481.55},
}

@article{bommasaniOpportunities2022,
  title = {On the {{Opportunities}} and {{Risks}} of {{Foundation Models}}},
  author = {Bommasani, Rishi and Hudson, Drew A. and Adeli, Ehsan and Altman, Russ and Arora, Simran and von Arx, Sydney and Bernstein, Michael S. and Bohg, Jeannette and Bosselut, Antoine and Brunskill, Emma and Brynjolfsson, Erik and Buch, Shyamal and Card, Dallas and Castellon, Rodrigo and Chatterji, Niladri and Chen, Annie and Creel, Kathleen and Davis, Jared Quincy and Demszky, Dora and Donahue, Chris and Doumbouya, Moussa and Durmus, Esin and Ermon, Stefano and Etchemendy, John and Ethayarajh, Kawin and {Fei-Fei}, Li and Finn, Chelsea and Gale, Trevor and Gillespie, Lauren and Goel, Karan and Goodman, Noah and Grossman, Shelby and Guha, Neel and Hashimoto, Tatsunori and Henderson, Peter and Hewitt, John and Ho, Daniel E. and Hong, Jenny and Hsu, Kyle and Huang, Jing and Icard, Thomas and Jain, Saahil and Jurafsky, Dan and Kalluri, Pratyusha and Karamcheti, Siddharth and Keeling, Geoff and Khani, Fereshte and Khattab, Omar and Koh, Pang Wei and Krass, Mark and Krishna, Ranjay and Kuditipudi, Rohith and Kumar, Ananya and Ladhak, Faisal and Lee, Mina and Lee, Tony and Leskovec, Jure and Levent, Isabelle and Li, Xiang Lisa and Li, Xuechen and Ma, Tengyu and Malik, Ali and Manning, Christopher D. and Mirchandani, Suvir and Mitchell, Eric and Munyikwa, Zanele and Nair, Suraj and Narayan, Avanika and Narayanan, Deepak and Newman, Ben and Nie, Allen and Niebles, Juan Carlos and Nilforoshan, Hamed and Nyarko, Julian and Ogut, Giray and Orr, Laurel and Papadimitriou, Isabel and Park, Joon Sung and Piech, Chris and Portelance, Eva and Potts, Christopher and Raghunathan, Aditi and Reich, Rob and Ren, Hongyu and Rong, Frieda and Roohani, Yusuf and Ruiz, Camilo and Ryan, Jack and R{\'e}, Christopher and Sadigh, Dorsa and Sagawa, Shiori and Santhanam, Keshav and Shih, Andy and Srinivasan, Krishnan and Tamkin, Alex and Taori, Rohan and Thomas, Armin W. and Tram{\`e}r, Florian and Wang, Rose E. and Wang, William and Wu, Bohan and Wu, Jiajun and Wu, Yuhuai and Xie, Sang Michael and Yasunaga, Michihiro and You, Jiaxuan and Zaharia, Matei and Zhang, Michael and Zhang, Tianyi and Zhang, Xikun and Zhang, Yuhui and Zheng, Lucia and Zhou, Kaitlyn and Liang, Percy},
  year = {2022},
  number = {arXiv:2108.07258},
  eprint = {2108.07258},
  primaryclass = {cs},
  fdoi = {10.48550/arXiv.2108.07258},
}

@inproceedings{soltaniehAnalysis2022a,
  title = {Analysis of {{Augmentations}} for {{Contrastive ECG Representation Learning}}},
  booktitle = {2022 {{International Joint Conference}} on {{Neural Networks}} ({{IJCNN}})},
  author = {Soltanieh, Sahar and Etemad, Ali and Hashemi, Javad},
  year = {2022},
  pages = {1--10},
  fissn = {2161-4407},
  fdoi = {10.1109/IJCNN55064.2022.9892600},
}

@article{score22021score2,
    title        = {SCORE2 risk prediction algorithms: new models to estimate 10-year risk of cardiovascular disease in Europe},
    author       = {SCORE2 working group and ESC Cardiovascular risk collaboration},
    year         = 2021,
    journal      = {European Heart Journal},
    volume       = 42,
    number       = 25,
    pages        = {2439--2454}
}

@article{score2-op2021score2-op,
    title        = {SCORE2-OP risk prediction algorithms: estimating incident cardiovascular event risk in older persons in four geographical risk regions},
    author       = {SCORE2-OP working group and ESC Cardiovascular risk collaboration},
    year         = 2021,
    journal      = {European Heart Journal},
    volume       = 42,
    number       = 25,
    pages        = {2455--2467}
}

@article{solosenko2021model,
    title        = {Model for simulating ecg and ppg signals with arrhythmia episodes},
    author       = {Solo{\v{s}}enko, Andrius and Petr{\.e}nas, Andrius and Paliakait{\.e}, Birut{\.e} and Marozas, Vaidotas and S{\"o}rnmo, Leif},
    year         = 2021,
    journal      = {PhysioNet},
    volume       = 101,
    pages        = 23
}

@article{lgoldberger2000physiobank,
    title        = {PhysioBank, PhysioToolkit, and PhysioNet},
    author       = {Ary L. Goldberger  and Luis A. N. Amaral  and Leon Glass  and Jeffrey M. Hausdorff  and Plamen Ch. Ivanov  and Roger G. Mark  and Joseph E. Mietus  and George B. Moody  and Chung-Kang Peng  and H. Eugene Stanley},
    year         = 2000,
    journal      = {Circulation},
    volume       = 101,
    number       = 23,
    pages        = {e215-e220}
}

@article{erturk2025sensor,
    title        = {Beyond Sensor Data: Foundation Models of Behavioral Data from Wearables Improve Health Predictions},
    author       = {Erturk, Eray and Kamran, Fahad and Abbaspourazad, Salar and Jewell, Sean and Sharma, Harsh and Li, Yujie and Williamson, Sinead and Foti, Nicholas J and Futoma, Joseph},
    year         = 2025,
    journal      = {arXiv preprint arXiv:2507.00191}
}

@article{gow2023mimic-iv-ecg,
    title        = {Mimic-iv-ecg: Diagnostic electrocardiogram matched subset},
    author       = {Gow, Brian and Pollard, Tom and Nathanson, Larry A and Johnson, Alistair and Moody, Benjamin and Fernandes, Chrystinne and Greenbaum, Nathaniel and Waks, Jonathan W and Eslami, Parastou and Carbonati, Tanner and others},
    year         = 2023,
    journal      = {Type: dataset},
    volume       = 6,
    pages        = {13--14}
}

@inproceedings{goswami2024moment,
    title        = {{MOMENT}: A Family of Open Time-series Foundation Models},
    author       = {Mononito Goswami and Konrad Szafer and Arjun Choudhry and Yifu Cai and Shuo Li and Artur Dubrawski},
    year         = 2024,
    booktitle    = {International Conference on Machine Learning}
}

@inproceedings{woo2024unified,
    title        = {Unified Training of Universal Time Series Forecasting Transformers},
    author       = {Gerald Woo and Chenghao Liu and Akshat Kumar and Caiming Xiong and Silvio Savarese and Doyen Sahoo},
    year         = 2024,
    booktitle    = {International Conference on Machine Learning}
}

@article{ribeiro2020automatic,
    title        = {{Automatic diagnosis of the 12-lead ECG using a deep neural network}},
    author       = {Ribeiro, Ant{\^o}nio H. and Ribeiro, Manoel Horta and Paix{\~a}o, Gabriela M. M. and Oliveira, Derick M. and Gomes, Paulo R. and Canazart, J{\'e}ssica A. and Ferreira, Milton P. S. and Andersson, Carl R. and Macfarlane, Peter W. and Meira Jr., Wagner and Sch{\"o}n, Thomas B. and Ribeiro, Antonio Luiz P.},
    year         = 2020,
    journal      = {Nature Communications},
    volume       = 11,
    number       = 1,
    pages        = 1760,
    issn         = {2041-1723}
}

@book{garcia200112,
    title        = {{12 Lead ECG: The Art of Interpretation}},
    author       = {Garcia, Tomas B and Holtz, Neil E},
    year         = 2001,
    publisher    = {Jones \& Bartlett Learning}
}

@article{geselowitz1989theory,
    title        = {On the theory of the electrocardiogram},
    author       = {Geselowitz, D.B.},
    year         = 1989,
    journal      = {Proceedings of the IEEE},
    volume       = 77,
    number       = 6,
    pages        = {857--876}
}

@book{lilly2012pathophysiology,
    title        = {Pathophysiology of heart disease: a collaborative project of medical students and faculty},
    author       = {Lilly, Leonard S},
    year         = 2012
}

@article{pnelwan2004reconstruction,
    title        = {Reconstruction of the 12-lead electrocardiogram from reduced lead sets},
    author       = {Stefan P Nelwan and Jan A Kors and Simon H Meij and Jan H {van Bemmel} and Maarten L Simoons},
    year         = 2004,
    journal      = {Journal of Electrocardiology},
    volume       = 37,
    number       = 1,
    pages        = {11--18}
}

@inproceedings{preejith2016wearable,
    title        = {Wearable ECG platform for continuous cardiac monitoring},
    author       = {Preejith, S P and Dhinesh, R and Joseph, Jayaraj and Sivaprakasam, Mohanasankar},
    year         = 2016,
    booktitle    = {Annual International Conference of the IEEE Engineering in Medicine and Biology Society (EMBC)},
    pages        = {623--626}
}

@article{majumder2018noncontact,
    title        = {Noncontact Wearable Wireless ECG Systems for Long-Term Monitoring},
    author       = {Majumder, Sumit and Chen, Leon and Marinov, Ognian and Chen, Chih-Hung and Mondal, Tapas and Deen, M. Jamal},
    year         = 2018,
    journal      = {IEEE Reviews in Biomedical Engineering},
    volume       = 11,
    pages        = {306--321}
}

@inproceedings{clark2018wearable,
    title        = {A wearable ECG monitoring system for real-time arrhythmia detection},
    author       = {Clark, Nicholar and Sandor, Edward and Walden, Calvin and Ahn, In Soo and Lu, Yufeng},
    year         = 2018,
    booktitle    = {IEEE International Midwest Symposium on Circuits and Systems (MWSCAS)},
    pages        = {787--790}
}

@article{ravanshad2014level-crossing,
    title        = {A Level-Crossing Based QRS-Detection Algorithm for Wearable ECG Sensors},
    author       = {Ravanshad, Nassim and Rezaee-Dehsorkh, Hamidreza and Lotfi, Reza and Lian, Yong},
    year         = 2014,
    journal      = {IEEE Journal of Biomedical and Health Informatics},
    volume       = 18,
    number       = 1,
    pages        = {183--192}
}

@inproceedings{huda2020low-cost,
    title        = {A Low-cost, Low-energy Wearable ECG System with Cloud-Based Arrhythmia Detection},
    author       = {Huda, Nurul and Khan, Sadia and Abid, Ragib and Shuvo, Samiul Based and Labib, Mir Maheen and Hasan, Taufiq},
    year         = 2020,
    booktitle    = {2020 IEEE Region 10 Symposium (TENSYMP)},
    pages        = {1840--1843}
}

@article{wang2022wearable,
    title        = {A wearable ECG monitor for deep learning based real-time cardiovascular disease detection},
    author       = {Wang, Peng and Lin, Zihuai and Yan, Xucun and Chen, Zijiao and Ding, Ming and Song, Yang and Meng, Lu},
    year         = 2022,
    journal      = {arXiv preprint arXiv:2201.10083}
}

@inproceedings{lin2019artificial,
    title        = {Artificial Intelligence of Things Wearable System for Cardiac Disease Detection},
    author       = {Lin, Yu-Jin and Chuang, Chen-Wei and Yen, Chun-Yueh and Huang, Sheng-Hsin and Huang, Peng-Wei and Chen, Ju-Yi and Lee, Shuenn-Yuh},
    year         = 2019,
    booktitle    = {IEEE International Conference on Artificial Intelligence Circuits and Systems (AICAS)},
    pages        = {67--70}
}

@inproceedings{melillo2015wearable,
    title        = {Wearable technology and ECG processing for fall risk assessment, prevention and detection},
    author       = {Melillo, Paolo and Castaldo, Rossana and Sannino, Giovanna and Orrico, Ada and de Pietro, Giuseppe and Pecchia, Leandro},
    year         = 2015,
    booktitle    = {Annual International Conference of the IEEE Engineering in Medicine and Biology Society (EMBC)},
    pages        = {7740--7743}
}

@article{sopic2018real-time,
    title        = {Real-Time Event-Driven Classification Technique for Early Detection and Prevention of Myocardial Infarction on Wearable Systems},
    author       = {Sopic, Dionisije and Aminifar, Amin and Aminifar, Amir and Atienza, David},
    year         = 2018,
    journal      = {IEEE Transactions on Biomedical Circuits and Systems},
    volume       = 12,
    number       = 5,
    pages        = {982--992}
}

@article{khunte2023detection,
    title        = {Detection of left ventricular systolic dysfunction from single-lead electrocardiography adapted for portable and wearable devices},
    author       = {Khunte, Akshay and Sangha, Veer and Oikonomou, Evangelos K. and Dhingra, Lovedeep S. and Aminorroaya, Arya and Mortazavi, Bobak J. and Coppi, Andreas and Brandt, Cynthia A. and Krumholz, Harlan M. and Khera, Rohan},
    year         = 2023,
    journal      = {npj Digital Medicine},
    volume       = 6,
    number       = 1,
    pages        = 124
}

@inproceedings{grill2020bootstrap,
    title        = {Bootstrap Your Own Latent - A New Approach to Self-Supervised Learning},
    author       = {Grill, Jean-Bastien and Strub, Florian and Altch\'{e}, Florent and Tallec, Corentin and Richemond, Pierre and Buchatskaya, Elena and Doersch, Carl and Avila Pires, Bernardo and Guo, Zhaohan and Gheshlaghi Azar, Mohammad and Piot, Bilal and kavukcuoglu, koray and Munos, Remi and Valko, Michal},
    year         = 2020,
    booktitle    = {Advances in Neural Information Processing Systems},
    volume       = 33,
    pages        = {21271--21284}
}

@inproceedings{he2020momentum,
    title        = {Momentum Contrast for Unsupervised Visual Representation Learning},
    author       = {He, Kaiming and Fan, Haoqi and Wu, Yuxin and Xie, Saining and Girshick, Ross},
    year         = 2020,
    booktitle    = {Proceedings of the IEEE/CVF Conference on Computer Vision and Pattern Recognition (CVPR)}
}

@article{qin2023mvkt,
    title        = {MVKT-ECG: Efficient single-lead ECG classification for multi-label arrhythmia by multi-view knowledge transferring},
    author       = {Yuzhen Qin and Li Sun and Hui Chen and Wenming Yang and Wei-Qiang Zhang and Jintao Fei and Guijin Wang},
    year         = 2023,
    journal      = {Computers in Biology and Medicine},
    volume       = 166,
    pages        = 107503
}

@article{conroy2003estimation,
    title        = {Estimation of ten-year risk of fatal cardiovascular disease in Europe: the {SCORE} project},
    author       = {Conroy, R M and Py{\"o}r{\"a}l{\"a}, K and Fitzgerald, A P and Sans, S and Menotti, A and De Backer, G and De Bacquer, D and Ducimeti{\`e}re, P and Jousilahti, P and Keil, U and Nj{\o}lstad, I and Oganov, R G and Thomsen, T and Tunstall-Pedoe, H and Tverdal, A and Wedel, H and Whincup, P and Wilhelmsen, L and Graham, I M and {SCORE project group}},
    year         = 2003,
    journal      = {Eur Heart J},
    volume       = 24,
    number       = 11,
    pages        = {987--1003}
}

@article{hughes2023deep,
    title        = {A deep learning-based electrocardiogram risk score for long term cardiovascular death and disease},
    author       = {Hughes, J. Weston and Tooley, James and Torres Soto, Jessica and Ostropolets, Anna and Poterucha, Tim and Christensen, Matthew Kai and Yuan, Neal and Ehlert, Ben and Kaur, Dhamanpreet and Kang, Guson and Rogers, Albert and Narayan, Sanjiv and Elias, Pierre and Ouyang, David and Ashley, Euan and Zou, James and Perez, Marco V.},
    year         = 2023,
    journal      = {npj Digital Medicine},
    volume       = 6,
    number       = 1,
    pages        = 169
}

@article{Zheng2020a,
    title        = {{A 12-lead electrocardiogram database for arrhythmia research covering more than 10,000 patients}},
    author       = {Zheng, Jianwei and Zhang, Jianming and Danioko, Sidy and Yao, Hai and Guo, Hangyuan and Rakovski, Cyril},
    year         = 2020,
    journal      = {Scientific Data},
    volume       = 7,
    number       = 1,
    pages        = 48
}

@article{martin2025music,
    title        = {{Music (sudden cardiac death in chronic heart failure)}},
    author       = {Martin-Yebra, Alba and Mart{\'\i}nez, Juan Pablo and Laguna, Pablo},
    year         = 2025
}

@article{tan2019icentia11k,
    title        = {{Icentia11k: An unsupervised representation learning dataset for arrhythmia subtype discovery}},
    author       = {Tan, Shawn and Androz, Guillaume and Chamseddine, Ahmad and Fecteau, Pierre and Courville, Aaron and Bengio, Yoshua and Cohen, Joseph Paul},
    year         = 2019,
    journal      = {arXiv preprint arXiv:1910.09570}
}

@inproceedings{hong2020holmes,
    title        = {{HOLMES: Health OnLine Model Ensemble Serving for Deep Learning Models in Intensive Care Units}},
    author       = {Hong, Shenda and Xu, Yanbo and Khare, Alind and Priambada, Satria and Maher, Kevin and Aljiffry, Alaa and Sun, Jimeng and Tumanov, Alexey},
    year         = 2020,
    booktitle    = {Proceedings of the ACM SIGKDD International Conference on Knowledge Discovery \& Data Mining},
    pages        = {1614--1624}
}

@article{mieloszyk2022comparison,
    title        = {{A Comparison of Wearable Tonometry, Photoplethysmography, and Electrocardiography for Cuffless Measurement of Blood Pressure in an Ambulatory Setting}},
    author       = {Mieloszyk, Rebecca and Twede, Hope and Lester, Jonathan and Wander, Jeremiah and Basu, Sumit and Cohn, Gabe and Smith, Greg and Morris, Dan and Gupta, Sidhant and Tan, Desney and Villar, Nicolas and Wolf, Moni and Malladi, Sailaja and Mickelson, Matt and Ryan, Lauren and Kim, Lindsey and Kepple, Jeffrey and Kirchner, Susanne and Wampler, Emma and Terada, Riena and Robinson, Joel and Paulsen, Ron and Saponas, T. Scott},
    year         = 2022,
    journal      = {IEEE Journal of Biomedical and Health Informatics},
    volume       = 26,
    number       = 7,
    pages        = {2864--2875}
}

@article{kansal2025multimodal,
    title        = {{Multimodal Clinical Monitoring in the Emergency Department (MC-MED)}},
    author       = {Kansal, Aman and Chen, Emma and Jin, Tom and Rajpurkar, Pranav and Kim, David},
    year         = 2025,
    journal      = {PhysioNet, v1},
    volume       = 10
}

@article{kligfield2007recommendations,
    title        = {{Recommendations for the Standardization and Interpretation of the Electrocardiogram}},
    author       = {Paul Kligfield  and Leonard S. Gettes  and James J. Bailey  and Rory Childers  and Barbara J. Deal  and E. William Hancock  and Gerard van Herpen  and Jan A. Kors  and Peter Macfarlane  and David M. Mirvis  and Olle Pahlm  and Pentti Rautaharju  and Galen S. Wagner},
    year         = 2007,
    journal      = {JACC},
    volume       = 49,
    number       = 10,
    pages        = {1109--1127},
}

@article{krizhevsky2009learning,
    title        = {{Learning multiple layers of features from tiny images.(2009)}},
    author       = {Krizhevsky, Alex and Hinton, Geoffrey and others},
    year         = 2009
}

@article{solovsenko2021model,
    title        = {{Model for simulating ecg and ppg signals with arrhythmia episodes}},
    author       = {Solo{\v{s}}enko, Andrius and Petr{\.e}nas, Andrius and Paliakait{\.e}, Birut{\.e} and Marozas, Vaidotas and S{\"o}rnmo, Leif},
    year         = 2021,
    journal      = {PhysioNet},
    volume       = 101,
    pages        = 23
}

@article{liu2018open,
    title        = {{An open access database for evaluating the algorithms of electrocardiogram rhythm and morphology abnormality detection}},
    author       = {Liu, Feifei and Liu, Chengyu and Zhao, Lina and Zhang, Xiangyu and Wu, Xiaoling and Xu, Xiaoyan and Liu, Yulin and Ma, Caiyun and Wei, Shoushui and He, Zhiqiang and others},
    year         = 2018,
    journal      = {Journal of Medical Imaging and Health Informatics},
    volume       = 8,
    number       = 7,
    pages        = {1368--1373}
}

@article{ribeiro2021code,
    title        = {{CODE-15\%: A large scale annotated dataset of 12-lead ECGs}},
    author       = {Ribeiro, Ant{\^o}nio H and Paixao, GM and Lima, Emilly M and Ribeiro, M Horta and Pinto Filho, Marcelo M and Gomes, Paulo R and Oliveira, Derick M and Meira Jr, Wagner and Schon, Th{\"o}mas B and Ribeiro, Antonio Luiz P},
    year         = 2021,
    journal      = {Zenodo, Jun},
    volume       = 9,
    pages        = {10--5281}
}

@article{alday2020classification,
    title        = {{Classification of 12-lead ecgs: the physionet/computing in cardiology challenge 2020}},
    author       = {Alday, Erick A Perez and Gu, Annie and Shah, Amit J and Robichaux, Chad and Wong, An-Kwok Ian and Liu, Chengyu and Liu, Feifei and Rad, Ali Bahrami and Elola, Andoni and Seyedi, Salman and others},
    year         = 2020,
    journal      = {Physiological measurement},
    publisher    = {IOP Publishing},
    volume       = 41,
    number       = 12,
    pages        = 124003
}

@inproceedings{reyna2021will,
    title        = {{Will two do? Varying dimensions in electrocardiography: the PhysioNet/Computing in Cardiology Challenge 2021}},
    author       = {Reyna, Matthew A and Sadr, Nadi and Alday, Erick A Perez and Gu, Annie and Shah, Amit J and Robichaux, Chad and Rad, Ali Bahrami and Elola, Andoni and Seyedi, Salman and Ansari, Sardar and others},
    year         = 2021,
    booktitle    = {2021 computing in cardiology (CinC)},
    volume       = 48,
    pages        = {1--4}
}

@inproceedings{wu2018unsupervised,
    title        = {{Unsupervised Feature Learning via Non-Parametric Instance Discrimination}},
    author       = {Wu, Zhirong and Xiong, Yuanjun and Yu, Stella X. and Lin, Dahua},
    year         = 2018,
    month        = {June},
    booktitle    = {Proceedings of the IEEE Conference on Computer Vision and Pattern Recognition (CVPR)}
}

\clearpage

\appendix

\setcounter{table}{0}
\renewcommand{\thetable}{S\arabic{table}}%
\setcounter{figure}{0}
\renewcommand{\thefigure}{S\arabic{figure}}%
\setcounter{equation}{0}
\renewcommand{\theequation}{S\arabic{equation}}%

\@toptitlebar
\section*{\Large\centering Supplementary Material
\\
\acro{}: CLINICALLY-GUIDED CONTRASTIVE LEARNING
FOR ELECTROCARDIOGRAM FOUNDATION MODELS}
\@bottomtitlebar

\section{Related Work}
\label{appsec:related_work}

\subsection{ECG and ML}

{\bf Utility of single-lead ECGs compared to $12$-lead.} Electrodes on the limbs and chest provide spatially diverse views of cardiac activity~\cite{kligfield2007recommendations}, producing $12$-lead ECGs to assess a range of heart conditions and abnormalities~\citep{garcia200112,ribeiro2020automatic}. The growing usage of wearables and mobile health devices with single-lead ECG functionality~\citep{hannunCardiologistLevel2019a,friedman2024electrocardiogram, attia2022prospective} has opened new opportunities for continuous health monitoring beyond traditional care settings~\citep{preejith2016wearable,majumder2018noncontact,clark2018wearable}, with applications including early detection of cardiac events~\citep{lin2019artificial,huda2020low-cost,wang2022wearable}, long-term pattern tracking~\citep{ravanshad2014level-crossing,majumder2018noncontact}, and proactive health interventions~\citep{melillo2015wearable,sopic2018real-time, bommasaniOpportunities2022}.

Prior work~\cite{jimenez202212} studied the Computing in Cardiology 2021 ECG dataset (covering $88{,}253$ annotated $12$-lead training recordings). They compared ML performance over $26$ target cardiac conditions when processing only lead I versus all $12$ leads. They found that $12$ leads give the best overall performance, but using only lead I yields a modest average degradation rather than a collapse in accuracy: the average of G-metric (geometric mean of sensitivity and specificity) falls from $0.80$ to $0.74$. Specifically, among the $26$ conditions, performance is similar (absolute G-metric change lower than $0.03$) for $12$ conditions; there is a moderate loss(G-metric drop from $0.04$ to $0.07$) for $7$ conditions; and a significant drop (G-metric drop more than $0.07$) for $7$ conditions. On average, the drop in the G-metric is $0.06$. Overall, this shows that single-lead (lead I) ECGs can reliably detect many rhythm-based abnormalities (making them attractive for wearables/screening), but they perform substantially worse for axis deviations, certain conduction blocks, and morphology-dependent findings that require precordial leads. Another observational study~\cite{angelaki2025diagnostic} on $1{,}254$ subjects showed that a single-lead ECG (lead I), when combined with other demographic features, can diagnose arterial hypertension with high accuracy (AUC $0.831$, sensitivity $72\%$, specificity $82\%$), demonstrating that even single-lead ECGs contain sufficient diagnostic information when analyzed effectively. Additionally, a study by~\cite{ramirez2024art} demonstrates that the standard 12-lead ECG contains redundant information when classifying cardiovascular diseases using CNNs. By selecting subsets of leads or applying transformations to the ECG signals, the authors evaluated how these adjustments influenced a CNN's diagnostic performance. Their main finding is that carefully optimizing input configurations, hence reducing redundancy while preserving essential information, can improve deep learning model performance and enable efficient diagnostics even with fewer leads.

{\bf Diagnostic potential and limitations of machine learning for ECG analysis.}
A proof-of-concept study by~\cite{sun2022ecg} demonstrates that a ResNet model trained on large-scale, population-based ECG datasets can accurately predict a wide range of diseases, including many non-cardiovascular conditions. Using over $1.5$ million ECGs linked to $11$K unique WHO ICD codes from 240K patients across $26$ hospitals in Canada, the authors identified $700$ disease categories with sufficient data for modeling. Models achieved strong discriminative performance (AUROC > $80\%$) for $80$ disease categories, with $18$ categories exceeding AUROC $>90\%$ (including non-cardivascular conditions such as silicosis, type 1 diabetes mellitus, liver diseases, behavioral disorder due to drug use, and some maternal diseases during pregnancy). Despite excellent AUROC values, precision was limited for many conditions due to their low prevalence, suggesting greater utility for rule-out screening rather than definitive diagnosis. The findings highlight the untapped diagnostic potential of ECGs for diverse diseases, while also noting that predictions may reflect correlated comorbidities or patient characteristics rather than direct disease-specific ECG changes, and that further targeted, clinically adjudicated studies are needed before deployment. Another study~\cite{kim2024deep} using data from $919$ patients found that time-dependent follow-up features contributed more strongly to predicting heart failure rehospitalization than admission or discharge variables, highlighting the dynamic nature of heart failure risk and underscoring the importance of ongoing monitoring and medication adherence during the post-discharge period for more accurate risk stratification and targeted intervention.

\begin{table}[t]
    \centering
    \renewcommand{\arraystretch}{.92}
    \caption{Available metadata for each database. M-iv-ecg is short for MIMIC-IV-ECG, G12EC is short for Georgia 12-lead ECG Challenge, and Phy2021 is short for PhysioNet 2021. }
    \label{apptab:ecg_dataset}
    \setlength{\tabcolsep}{2pt}
    \resizebox{\columnwidth}{!}{%
    \small
    \begin{tabular}{lcccccccc}
    \toprule
    Dataset & 
    \href{https://physionet.org/content/mimic-iv-ecg/}{M-iv-ecg}~\citeyearpar{gow2023mimic-iv-ecg} & \href{https://figshare.com/articles/figure/CPSC2018/22197715}{CPSC2018}~\citeyearpar{liu2018open} & 
    \href{https://physionet.org/content/ecg-arrhythmia/}{Chapman}~\citeyearpar{Zheng2020a} & \href{https://physionet.org/content/ptb-xl/1.0.3/}{PTB-XL}~\citeyearpar{wagner2020ptb-xl} & \href{https://zenodo.org/records/4916206}{CODE-15}~\citeyearpar{ribeiro2021code} & \href{https://physionet.org/content/challenge-2020/}{G12EC}~\citeyearpar{alday2020classification} & \href{https://bdsp.io/content/heedb/}{HEEDB}~\citeyearpar{ghanta2025harvard} & \href{https://physionet.org/content/challenge-2021/}{Phy2021}~\citeyearpar{reyna2021will} \\
    \midrule
    Lead number    & 12 & 12 & 12 & 12 & 12 & 12 & 12 & 12 \\
    Record number  & 800{,}035 & - & 45{,}152 & 21{,}799 & 345{,}779 & - & 10{,}471{,}531 & - \\
    Patient number  &  161{,}352 & 50{,}165 & 45{,}152 & 18{,}869 & 233{,}770 & - &  1{,}818{,}247 & - \\
    Sample rate (Hz) & 500 & 500 & 500 & 500 & 400 & 500 & 500 & varied \\
    Duration & 10~s & 10~s & 10~s & 10~s & 10~s & varied & 10~s & varied \\
    
    \midrule

    Sex & \cmark & \cmark & \cmark & \cmark & \cmark & \cmark & \cmark & \cmark \\ 
    Sex at Birth & & & & & & & \cmark \\
    Gender identity & & & & & & & & \cmark \\
    Age     & \cmark & \cmark & \cmark & \cmark & \cmark & \cmark & \cmark & \cmark \\
    Weight & \cmark & & & \cmark \\
    Height & \cmark & & & \cmark \\
    BMI     & \cmark & & & \\
    Race    & \cmark & & & & & & \cmark \\
    Ethnicity & & & & & & & \cmark \\
    Marital status & & & & & & & \cmark \\
    Religion & & & & & & & \cmark \\
    Language & & & & & & & \cmark \\
    Veteran & & & & & & & \cmark \\
    Education & & & & & & & \cmark \\
    Date of Birth & & & & & & & \cmark \\
    Date of Death  & \cmark & & & & & & \cmark \\
    Last visit date & & & & & & & \cmark \\
    SNOMED & & & \cmark \\
    sinus rhythm & & & & & \cmark \\
    AF & & & & & \cmark \\
    bundle branch block & & & & & \cmark \\
    \midrule
    Used Studies & \begin{tabular}[c]{@{}l@{}} 
\citeauthor{tian2024foundation}\\
\citeauthor{liu2024zero-shot}\\
\citeauthor{jin2025reading}\\
\citeauthor{mckeen2024ecg-fm}
\end{tabular} 
& \begin{tabular}[c]{@{}l@{}} 
\citeauthor{mehari2022self-supervised}\\
\citeauthor{wang2024anyecg}
\end{tabular} 
& \citeauthor{na2024guiding} 
& \begin{tabular}[c]{@{}l@{}} 
\citeauthor{mehari2022self-supervised}\\
\citeauthor{wang2024anyecg}
\end{tabular} 
& \citeauthor{na2024guiding} 
& \citeauthor{wang2024anyecg} 
& \citeauthor{li2024electrocardiogram} 
& \citeauthor{mckeen2024ecg-fm} \\
    \bottomrule
    \end{tabular}
    }%
\end{table}

\subsection{Contrastive Approaches}

{\bf Adapting self-supervised learning for ECGs.}
Classic self-supervised methods based on instance discrimination and latent forecasting are adapted by~\cite{mehari2022self-supervised} to the ECG domain. They utilize time-series-specific data augmentations (including physiological noise models), modifying the CPC architecture to suit ECG’s temporal resolution, employing a joint 12-lead encoder, and systematically evaluating multiple frameworks on large public datasets for their impact on downstream performance, label efficiency, and robustness. They find out that contrastive predictive coding~(CPC) with these adaptations achieves near-supervised linear evaluation performance and significantly improves downstream accuracy, data efficiency, and robustness to physiological noise.

{\bf Multi-objective contrastive learning.}
A combination of two contrastive losses for pretraining is used by~\cite{oh2022lead}, where a Wav2Vec architecture captures local lead-level temporal features and a contrastive multi-segment coding architecture from~\cite{kiyasseh2021clocs} captures global patient-level context. The total loss is the sum of both, enabling the model to learn fine-grained intra-signal information and broader inter-patient relationships. Also, during pretraining, the authors introduce Random Lead Masking, where each ECG lead is randomly masked with a fixed probability to simulate reduced-lead scenarios. This augmentation makes the single pretrained model robust to arbitrary lead configurations during fine-tuning.

\begin{table}[ht]
\renewcommand{\arraystretch}{.93}
\centering
\caption{ECG diagnostic labels in the PTB-XL dataset. The dataset contains $42$ diagnostic labels in a $3$-level hierarchy: superclass, subclass, and specific diagnosis. The \textbf{Label} column denotes the specific diagnostic label, while the \textbf{Subclass} and \textbf{Superclass} column indicates its corresponding subclass and superclass categories.}
\label{tab:ptbxl_labels}
\footnotesize
\begin{tabular}{@{}p{0.12\textwidth}p{0.45\textwidth}p{0.18\textwidth}p{0.1\textwidth}@{}}
\toprule
\textbf{Label} & \textbf{Description} & \textbf{Subclass} & \textbf{Superclass} \\
\midrule
\multicolumn{4}{l}{\textbf{\textit{Conduction Disturbances (CD)}}} \\
\midrule
LAFB   & Left anterior fascicular block & LAFB/LPFB & CD \\
LPFB   & Left posterior fascicular block & LAFB/LPFB & CD \\
IRBBB  & Incomplete right bundle branch block & IRBBB & CD \\
ILBBB  & Incomplete left bundle branch block & ILBBB & CD \\
CRBBB  & Complete right bundle branch block & CRBBB & CD \\
CLBBB  & Complete left bundle branch block & CLBBB & CD \\
AVB    & First degree AV block & AVB & CD \\
3AVB   & Third degree AV block & AVB & CD \\
2AVB   & Second degree AV block & AVB & CD \\
IVCD   & Non-specific intraventricular conduction disturbance & IVCD & CD \\
WPW    & Wolff-Parkinson-White syndrome & WPW & CD \\
\midrule
\multicolumn{4}{l}{\textbf{\textit{Hypertrophy (HYP)}}} \\
\midrule
LVH    & Left ventricular hypertrophy & LVH & HYP \\
LAO/LAE & Left atrial overload/enlargement & LAO/LAE & HYP \\
RVH    & Right ventricular hypertrophy & RVH & HYP \\
RAO/RAE & Right atrial overload/enlargement & RAO/RAE & HYP \\
SEHYP  & Septal hypertrophy & SEHYP & HYP \\
\midrule
\multicolumn{4}{l}{\textbf{\textit{Myocardial Infarction (MI)}}} \\
\midrule
IMI    & Inferior myocardial infarction & IMI & MI \\
ILMI   & Inferolateral myocardial infarction & IMI & MI \\
IPLMI  & Inferoposterolateral myocardial infarction & IMI & MI \\
IPMI   & Inferoposterior myocardial infarction & IMI & MI \\
INJIN  & Subendocardial injury in inferior leads & IMI & MI \\
INJIL  & Subendocardial injury in inferolateral leads & IMI & MI \\
ASMI   & Anteroseptal myocardial infarction & AMI & MI \\
AMI    & Anterior myocardial infarction & AMI & MI \\
ALMI   & Anterolateral myocardial infarction & AMI & MI \\
INJAS  & Subendocardial injury in anteroseptal leads & AMI & MI \\
INJAL  & Subendocardial injury in anterolateral leads & AMI & MI \\
INJLA  & Subendocardial injury in lateral leads & AMI & MI \\
LMI    & Lateral myocardial infarction & LMI & MI \\
PMI    & Posterior myocardial infarction & PMI & MI \\
\midrule
\multicolumn{4}{l}{\textbf{\textit{ST/T Changes (STTC)}}} \\
\midrule
NDT    & Non-diagnostic T abnormalities & STTC & STTC \\
DIG    & Digitalis effect & STTC & STTC \\
LNGQT  & Long QT interval & STTC & STTC \\
ANEUR  & ST-T changes compatible with ventricular aneurysm & STTC & STTC \\
EL     & Electrolytic disturbance or drug effect & STTC & STTC \\
NST    & Non-specific ST changes & NST & STTC \\
ISC    & Non-specific ischemic changes & ISC & STTC \\
ISCIN  & Ischemic changes in inferior leads & ISCI & STTC \\
ISCIL  & Ischemic changes in inferolateral leads & ISCI & STTC \\
ISCAL  & Ischemic changes in anterolateral leads & ISCA & STTC \\
ISCAS  & Ischemic changes in anteroseptal leads & ISCA & STTC \\
ISCLA  & Ischemic changes in lateral leads & ISCA & STTC \\
ISCAN  & Ischemic changes in anterior leads & ISCA & STTC \\
\midrule
\multicolumn{4}{l}{\textbf{\textit{Normal (NORM)}}} \\
\midrule
NORM   & Normal ECG & NORM & NORM \\

\bottomrule
\end{tabular}
\end{table}

\begin{table}[t]
\centering
\renewcommand{\arraystretch}{.92}
\caption{Form labels and label descriptions in the PTB-XL dataset. Labels represent morphological and rhythm abnormalities detected in ECG recordings.}
\label{tab:ptbxl_form}
\footnotesize
\resizebox{\columnwidth}{!}{%
\begin{tabular}{@{}p{0.08\textwidth}p{0.4\textwidth}p{0.08\textwidth}p{0.4\textwidth}@{}}
\toprule
\textbf{Label} & \textbf{Description} & \textbf{Label} & \textbf{Description} \\
\midrule
ABQRS  & Abnormal QRS & LOWT   & Low amplitude T-waves \\
DIG    & Digitalis effect & LPR    & Prolonged PR interval \\
HVOLT  & High QRS voltage & LVOLT  & Low QRS voltages \\
INVT   & Inverted T-waves & NDT    & Non-diagnostic T abnormalities \\
LNGQT  & Long QT interval & NST    & Non-specific ST changes \\
NT     & Non-specific T-wave changes & PAC    & Atrial premature complex \\
PRC(S) & Premature complex(es) & PVC    & Ventricular premature complex \\
QWAVE  & Q waves present & STD    & Non-specific ST depression \\
STE    & Non-specific ST elevation & TAB    & T-wave abnormality \\
VCLVH  & Voltage criteria for LVH &  &  \\
\bottomrule
\end{tabular}
}%
\end{table}

\begin{table}[t]
\centering
\renewcommand{\arraystretch}{.92}
\caption{Rhythm labels and label descriptions in the PTB-XL dataset. Labels represent various cardiac rhythm patterns and arrhythmias detected in ECG recordings.}
\label{tab:ptbxl_rhythm}
\footnotesize
\resizebox{\columnwidth}{!}{%
\begin{tabular}{@{}p{0.08\textwidth}p{0.4\textwidth}p{0.08\textwidth}p{0.4\textwidth}@{}}
\toprule
\textbf{Label} & \textbf{Description} & \textbf{Label} & \textbf{Description} \\
\midrule
AFIB   & Atrial fibrillation & PACE   & Normal functioning artificial pacemaker \\
AFLT   & Atrial flutter & PSVT   & Paroxysmal supraventricular tachycardia \\
BIGU   & Bigeminal pattern (unknown origin) & SARRH  & Sinus arrhythmia \\
SBRAD  & Sinus bradycardia & SR     & Sinus rhythm \\
STACH  & Sinus tachycardia & SVARR  & Supraventricular arrhythmia \\
SVTAC  & Supraventricular tachycardia & TRIGU  & Trigeminal pattern (unknown origin) \\
\bottomrule
\end{tabular}
}%
\end{table}

\subsection{ECG Dataset Overview}

The ECG datasets used in prior work exhibit substantial variability in scale, metadata richness, and clinical scope, directly impacting foundation model development capabilities. A detailed summarization of the current available ECG dataset is tabulated in~\tableautorefname~\ref{apptab:ecg_dataset}. Among the datasets, MIMIC-IV-ECG emerges as a particularly well-suited one for health risk assessment applications due to its combination of large-scale clinical data ($161{,}352$ patients) and a relatively inclusive metadata, including gender, age, and SBP. In contrast, alternative datasets present significant limitations: smaller-scale collections like Chapman and PTB-XL lack the statistical power for robust foundation model training, while metadata-sparse datasets preclude comprehensive health risk modeling. Although HEEDB offers a superior scale (1.8 million patients), its restricted accessibility limits reproducibility and benchmarking capabilities.

\section{Details of the datasets used for pretraining and downstream evaluation}
\label{appsec:dataset_details}

{\bf MIMIC-IV-ECG} dataset is a large-scale clinical ECG database containing $800{,}035$ 12-lead diagnostic recordings collected from $161{,}352$ unique patients.\footnote{MIMIC-IV-ECG dataset available at: \href{https://physionet.org/content/mimic-iv-ecg/0.1.0/}{\texttt{physionet.org/content/mimic-iv-ecg}}.} Each ECG is $10$ seconds in length and sampled at $500$Hz. Corresponding metadata can be matched to each patient using the unique patient code.\footnote{Metadata for MIMIC-IV available at: \href{https://physionet.org/content/mimiciv/3.1/}{\texttt{physionet.org/content/mimiciv}}.} A more detailed procedure of aligning metadata for each ECG sample is provided in~\appendixautorefname~\ref{appsubsec:meta_align_mimiciv}. We use this dataset in $2$ phases: (1) for pretraining, we use the first unique ECG recording from each patient and train over all distinct patients, and (2) for downstream evaluation, we follow prior work~\citep{li2024electrocardiogram} who obtained the labels for the LVEF estimation task from the discharge section of MIMIC-IV-Notes. Specifically, LVEF values are provided as continuous labels for the regression task, while an LVEF of $50\%$ or higher is considered normal and below $50\%$ abnormal, thereby defining a binary classification task. We sequentially split the dataset into training ($60{,}329$ samples, $80\%$), validation ($7{,}541$ samples, $10\%$), and test sets ($7{,}542$ samples, $10\%$).

{\bf MUSIC} dataset (MUerte Subita en Insuficiencia Cardiaca) focuses on assessing cardiac mortality and sudden cardiac death (SCD) in ambulatory patients with chronic heart failure (CHF).\footnote{MUSIC dataset available at: \href{https://physionet.org/content/music-sudden-cardiac-death/1.0.1/}{\texttt{physionet.org/content/music-sudden-cardiac-death}}.} It contains $992$ patients with CHF consecutively enrolled from the specialized HF clinics of eight University Spanish Hospitals between April 2003 and December 2004. All patients are measured with a 3-lead resting electrocardiogram (ECG), or a 3-lead Holter ECG. In this study, we focus on using the first lead of the collected Holter ECG signal. In the original dataset, outcome labels (non-cardiovascular death, sudden cardiac death, or pump-failure death) are provided at the patient level. For each patient, we extract the first 10 seconds of ECG recordings with non-zero signals. For the downstream experiments, we take $188$ samples ($20\%$) as a held-out test set, while the remaining $655$ samples are for training and $93$ for validation.

{\bf PTB-XL} dataset is a large-scale ECG dataset that has been widely used in prior research (cite) for evaluating model capacity in signal pattern and disease identification.\footnote{PTB-XL dataset available at: \href{https://physionet.org/content/ptb-xl/1.0.3/}{\texttt{physionet.org/content/ptb-xl}.}} We adopt the preprocessing and label alignment procedures described in the original dataset publication.\footnote{Data preprocessing code for PTB-XL available at: \href{https://github.com/helme/ecg_ptbxl_benchmarking}{\texttt{github.com/helme/ecg\_ptbxl\_benchmarking}}.} Details of the tasks of PTB-XL are tabulated in~\tableautorefname s~\ref{tab:ptbxl_labels},~\ref{tab:ptbxl_form}, and~\ref{tab:ptbxl_rhythm} To ensure fair comparison, we follow prior work~\citep{wagner2020ptb-xl} and adopt a data partitioning of $[80/10/10]\%$, yielding $16{,}832$, $2{,}100$, and $2{,}098$ samples for training, validation, and testing phases, respectively.

{\bf Icentia11K} dataset is a wearable dataset that contains ECG signals collected from single-lead chest-mounted wearable devices.\footnote{Icentia11K dataset available at: \href{https://physionet.org/content/icentia11k-continuous-ecg/1.0/}{\texttt{physionet.org/content/icentia11k-continuous-ecg}}.} The sample rate for the dataset is $250$ Hz. Beat and rhythm labels are extracted from the dataset’s annotation files, focusing on three beat types (Normal, Supraventricular, Ventricular) and three rhythm types (Normal Sinus Rhythm, Atrial Fibrillation, Atrial Flutter), see~\tableautorefname s~\ref{tab:beat_symbols} and~\ref{tab:rhythm_symbols} for a more detailed description. Note that the dataset also includes an undefined beat class, which we omit here as it does not correspond to a physiologically interpretable beat type. For both tasks, segments are generated by slicing $10$-second windows starting from the annotated event and assigning numerical labels for classification. To ensure balanced representation across patients, we employ a patient-stratified sampling strategy: for each patient, we search their recordings for each target beat type and randomly select one representative instance, continuing until all types are found or available segments are exhausted. This results in relatively balanced distributions across beat classes (N: 10,866; S: 9,844; V: 9,287), whereas rhythm classes remain imbalanced due to the limited prevalence of atrial fibrillation and atrial flutter (N: 10,239; AFib: 743; AFL: 516). For experimental evaluation, the dataset is partitioned into training, validation, and test sets $(90\%/10\%/10\%)$, comprising $20,994/4,532/4,511$ samples for beat classification and $8,042/1,723/1,733$ samples for rhythm classification.

\begin{table}[t]
\centering
\renewcommand{\arraystretch}{.92}
\small
\caption{Beat Symbol Definitions with Icentia11K data set}
\label{tab:beat_symbols}
\begin{tabular}{ll}
\toprule
\textbf{Symbol} & \textbf{Beat Description} \\
\midrule
N & Normal \\ 
S & ESSV (PAC): Premature or ectopic supraventricular beat, premature atrial contraction \\ 
V & ESV (PVC): Premature ventricular contraction, premature ventricular contraction \\ 
\bottomrule
\end{tabular}
\end{table}

\begin{table}[t]
\centering
\renewcommand{\arraystretch}{.92}
\small
\caption{Rhythm symbol definitions with Icentia11K data set.}
\label{tab:rhythm_symbols}
\begin{tabular}{lll}
\toprule
\textbf{Symbol} & \textbf{Rhythm Type} & \bf Rhythm description\\
\midrule
(N \ldots ) & NSR (Normal sinus rhythm) & - \\
(AFIB \ldots ) & AFib (Atrial fibrillation) & Irregular rhythm with absent P waves and irregular RR intervals \\
(AFL \ldots ) & AFlutter (Atrial flutter) & Regular atrial arrhythmia with sawtooth flutter waves \\
\bottomrule
\end{tabular}

\end{table}

{\bf Chapman} dataset is a collection of 12-lead ECG recordings from $45{,}152$ patients labeled by clinical experts to support research in arrhythmia and cardiovascular disease detection.\footnote{Chapman dataset available at: \href{https://physionet.org/content/ecg-arrhythmia/1.0.0/}{\texttt{physionet.org/content/ecg-arrhythmia}}.} The dataset contains $10$-second recordings sampled at 500 Hz, featuring $11$ common cardiac rhythms and $67$ additional cardiovascular conditions, all validated through a rigorous multi-physician review process with senior physician arbitration for diagnostic disagreements. We follow prior work~\citep{jin2025reading} and use a refined version of the dataset, which contains $23{,}026$ ECG recordings with $38$ distinct labels. The samples are then split into training ($16,546$ samples, $70\%$), validation ($1,860$ samples, $10\%$), and test sets ($4,620$ samples, $20\%$).

{\bf MC-MED} dataset is a multimodal collection of emergency department visits from $118{,}385$ adult patients at Stanford Health Care between 2020 and 2022.\footnote{MC-MED dataset available at: \href{https://physionet.org/content/mc-med/1.0.0/}{\texttt{physionet.org/content/mc-med}}.} The dataset combines continuous physiological monitoring (lead II ECG, photoplethysmography, respiration waveforms), clinical data (demographics, medical histories, laboratory results, medications, radiology reports), and temporal visit outcomes. It covers emergency department patients during and after the COVID-19 pandemic with granular physiological measurements.

The prediction targets encompass both classification and regression tasks across the clinical care continuum. For classification, there are $3$ tasks: (1) emergency department (ED) disposition, which determines patient placement after ED assessment. This includes $4$ classes: Discharge (outpatient), Inpatient (hospital admission), Observation (extended monitoring without full admission), and ICU (critical care); (2) Discharge (DC) disposition, classification of patient outcomes at hospital discharge. We categorized the outcome into $5$ categories: home care, care facility, hospital transfer, psychiatric care, and death; and (3) Triage acuity assessment, which is predicting clinical urgency as determined by healthcare professionals during initial patient evaluation, including $5$ tasks: Resuscitation, emergent, urgent, Semi-Urgent, and Non-Urgent. The regression tasks involve continuous prediction of systolic blood pressure (SBP) and diastolic blood pressure (DBP) values. We follow the chronological splitting provided in the original dataset based on patient admission dates, and allocate the first $78\%$ ($37{,}438$ samples) for training, the following $11\%$ ($5{,}540$ samples) for validation, and the most recent $11\%$ ($5{,}572$ samples) for testing. Note that the exact number may differ from the original dataset, as we only include patients with lead II ECG recordings.

{\bf Aurora BP} dataset is a collection of simultaneous multi-modal physiological recordings collected from $1{,}221$ diverse participants,\footnote{A sample data of Aurora BP is available at: \href{https://github.com/microsoft/aurorabp-sample-data}{github.com/microsoft/aurorabp-sample-data}. Access can be provided via application.} serving the purpose for cuffless blood pressure research. The dataset contains synchronized tonometry, photoplethysmography (PPG), electrocardiography (ECG), accelerometry, and reference blood pressure measurements collected during both laboratory and 24-hour ambulatory monitoring phases, with participants spanning a wide range of ages and hypertensive status to ensure real-world applicability. The prediction targets are systolic blood pressure (SBP) and diastolic blood pressure (DBP), with the task formulated as the estimation of their actual values through regression. We use a $70\%/15\%/15\%$ train/validation/test split on patient ID, resulting in $786$/$169$/$169$ patients, which corresponds to $9{,}237$/$1{,}913$/$1{,}854$ samples for training, validation, and testing, respectively.

\subsection{Metadata alignment with MIMIC-IV-ECG dataset}
\label{appsubsec:meta_align_mimiciv}

The MIMIC-IV-ECG database provides raw electrocardiogram (ECG) waveforms together with patient identifiers and relative timestamps, but does not directly include demographic or clinical metadata. To match each ECG with patient information, we aligned it with the corresponding records in MIMIC-IV,\footnote{MIMIC-IV database available at: \href{https://physionet.org/content/mimiciv/3.1/}{physionet.org/content/mimiciv/3.1/}} which contains both static demographics (e.g., sex, age at admission) and time-stamped clinical observations (e.g., blood pressure). Each ECG was first linked to the corresponding patient using the unique patient identifier, which is shared across MIMIC-IV and MIMIC-IV-ECG. Demographic variables that remain constant over time were directly assigned to all ECGs of the same patient. For time-varying clinical measurements, we selected the most recent value recorded at or before the ECG timestamp, ensuring that the metadata reflected the patient’s state at the time of acquisition. If no prior measurement was available, the variable was marked as missing, and no forward-filling across admissions or imputation beyond this step was performed. This procedure ensures that every ECG recording is annotated with the most temporally relevant metadata.

\section{Supplementary experiment setup and baselines}
\label{appsec:supp_experiment_setup}

\subsection{Experiment setup for downstream evaluation}
\label{appsubsec:experiment_setup}

This section supplements the experiment setup in~\sectionautorefname~\ref{sec:experiment_evaluation} of the main paper. All experiments were conducted using a Linux server (Ubuntu 22.04.5) with 4 NVIDIA L40S GPUs, 2 Intel Xeon Gold 6542Y CPUs. 

\begin{figure}[!t]
    \centering
    \includegraphics[width=0.7\linewidth]{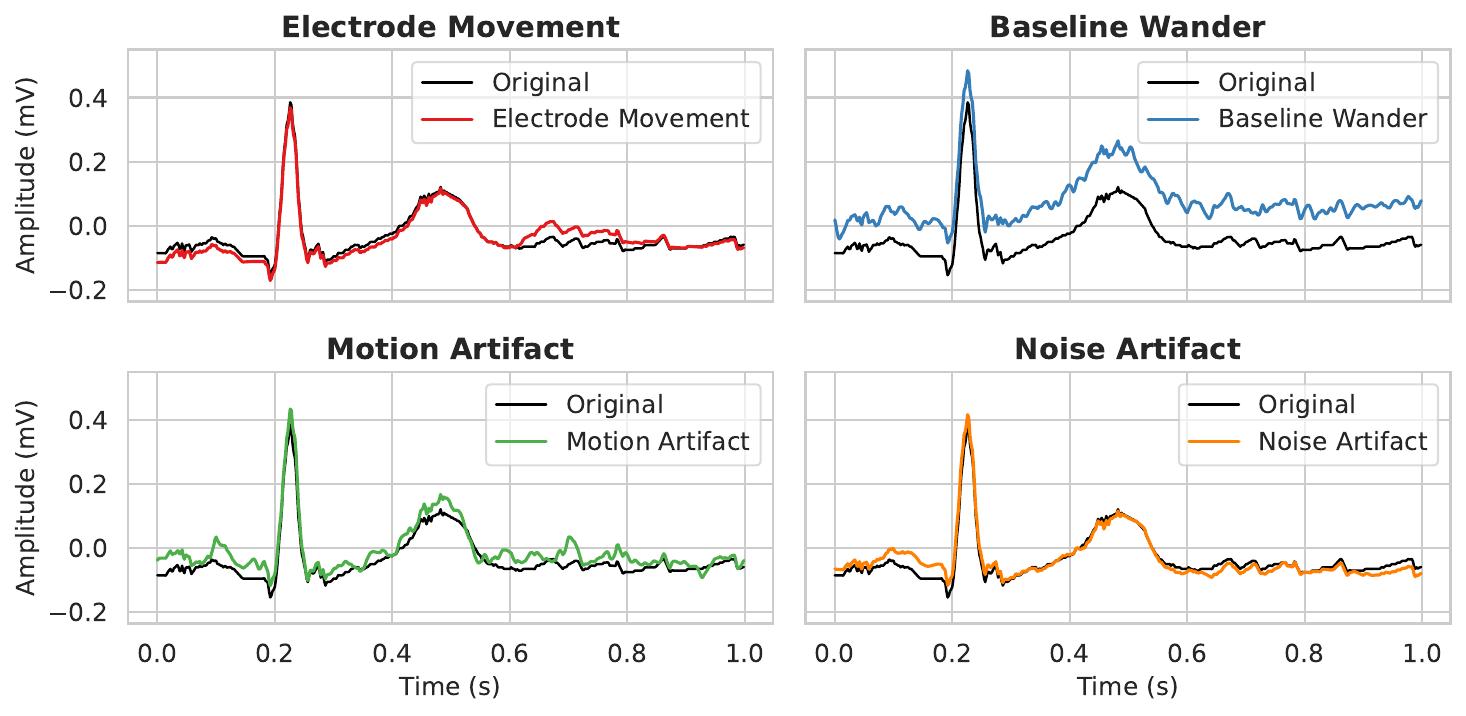}
    \caption{Physiological noise for single-lead ECG contrastive learning. We include four common sources of signal degradation in wearable ECG devices: electrode movement artifacts, baseline wander, motion-induced distortions, and additive noise. Original signals are shown in black, with augmented versions in color.}
    \label{fig:noise_augmentation}
\end{figure}

\textbf{Pretraining:} All experiments are conducted under fixed hyperparameter settings. All models are pretrained for $100$ epochs with a batch size of $64$ using the AdamW optimizer, except for the experiments for pretraining on a specific lead due to a reduced data diversity (\sectionautorefname~\ref{sec:results}, pretraining on a specific lead), where we pretrain for $10$ epochs. A learning rate of $1 \times 10^{-4}$ and a weight decay of $5 \times 10^{-5}$ are employed, with a Cosine Annealing Learning Rate scheduler to adjust the learning rate. Training stops if the validation loss does not decrease for $20$ consecutive epochs. Seed is fixed to $42$ for all pretraining. For parameters specific to contrastive pretraining, we set the clinical-guidance coefficient to $\alpha = 0.2$ and temperature to $\tau = 0.07$, consistent with prior literature~\citep{wu2018unsupervised, he2020momentum}. The effect of $\alpha$ and $\tau$ is further examined in~\appendixautorefname~\ref{appsubsec:parameter},~\tableautorefname~\ref{apptab:alpha}, and~\tableautorefname~\ref{apptab:tau}.

\textbf{Downstream tasks:} For downstream evaluation, training is also conducted for $100$ epochs with a batch size of $64$. We use binary cross-entropy loss for classification and L1 loss for regression tasks. Model parameters are optimized using Adam with a learning rate of $1 \times 10^{-3}$ and a weight decay of $1 \times 10^{-5}$. The learning rate is adjusted using a Cosine Annealing with Warm Restarts scheduler. Training stops if the validation loss does not decrease for $5$ consecutive epochs. Seed is fixed to $42$ for all downstream evaluation, except for the seed robustness evaluation in~\appendixautorefname~\ref{appsubsec:confidence_interval}.

\textbf{Downstream preprocessing:} For all downstream tasks, the ECG signals are resampled to 500 Hz. For datasets containing recordings longer than 10 seconds, we keep the first 10 seconds as our input. We preprocessed all ECG signals using a 5th-order Butterworth bandpass filter with cutoff frequencies of 0.67 Hz and 40 Hz to remove low-frequency baseline wander (e.g. due to breathing) whilst maintaining heart rate frequencies, and to remove high-frequency noise (e.g. power-line interference) whilst maintaining fundamental frequencies of P, QRS, and T-waves. The input ECG samples are then standardized using z-scoring, an operation that transforms the data to zero mean and unit standard deviation.

\textbf{Regression specific setup:} In all regression tasks, the target labels are also z-scored using statistics computed from the training set. Predictions are de-normalized back to the original scale before comparison with the ground-truth labels. The errors are evaluated on the original scale.

\subsection{Data augmentation for contrastive pretraining}
\label{appsubsec:data_augmentation}

This section supplements the \textbf{stochastic data augmentation} in~\sectionautorefname~\ref{sec:experiment_evaluation}.

\paragraph{Simulate physiological noise}

We inject physiological noise collected from the MIT-BIH noise stress test database~\citep{solovsenko2021model}, as a form of data augmentation that encourages the model to learn representations invariant to electrode movement and motion-induced artifacts, etc.\footnote{The noise is available at: \href{https://physionet.org/content/ecg-ppg-simulator-arrhythmia/1.3.1/}{\texttt{physionet.org/content/ecg-ppg-simulator-arrhythmia}}.} 
The noise injection is achieved by selecting the noise signal from the corresponding lead, given by $\mathbf{\hat{e}}^l_s = \mathbf{e}^l_s + \phi\!\times\!\mathbf{n}^l$, where $\phi$ controls the noise intensity, $\mathbf{n}^l$ is the noise vector of the $l$-th lead.
We provide visualization in~\figureautorefname~\ref{fig:noise_augmentation} of how different noises influence and augment the original signal. Note that we set $\phi=0.02$ following prior work~\citep{lgoldberger2000physiobank,solosenko2021model}, a parameter also used in the main experiment.

\paragraph{Random mask}

Randomly masking segments of the ECG signal encourages the model to learn representations that are invariant to missing information. In our experiments, we set the masking probability to p = $0.2$, with each selected segment having $10\%$ of its signal randomly masked.

\subsection{Backbone ResNet Models}
\label{appsubsec:backbone_resnet}

\begin{table}[t]
\centering
\small
\caption{ResNeXt1D Model Architecture Configurations for~\acro{} variants. \textbf{Left:} Shared hyperparameters of all three variants. \textbf{Right:} Detailed stage-wise model configurations.}
\label{apptab:resnet1d_configs}
\resizebox{0.5\columnwidth}{!}{%
\renewcommand{\arraystretch}{1.635}
\begin{tabular}{lccc}
\toprule
\textbf{Parameter} & \textbf{\acro-S} & \textbf{\acro-M} & \textbf{\acro-L} \\
\midrule
\textbf{Model Size} & 448K & 30.7M & 296M \\
\textbf{Hidden Dimension $d$} & 32 & 64 & 128 \\
\textbf{Ratio $\iota$} & 0.5 & 1.0 & 1.5 \\
\textbf{Groups Width $g$} & 8 & 16 & 32 \\
\textbf{Number of Stages $\varpi$} & 6 & 7 & 9 \\

\bottomrule
\end{tabular}
}
\quad
\setlength{\tabcolsep}{13.5pt}
\resizebox{0.45\columnwidth}{!}{%
\begin{tabular}{cccc}
\toprule
\multirow{2}{*}{\bf Stage} & \bf \acro-S & \bf \acro-M & \bf \acro-L \\
& $h$ / $\gamma$ & $h$ / $\gamma$ & $h$ / $\gamma$ \\
\midrule
1 & 32 / 1 & 64 / 2 & 128 / 2 \\
2 & 64 / 1 & 160 / 2 & 256 / 3 \\
3 & 64 / 2 & 160 / 2 & 256 / 3 \\
4 & 128 / 2 & 400 / 3 & 512 / 4 \\
5 & 128 / 2 & 400 / 3 & 512 / 4 \\
6 & 256 / 2 & 1024 / 4 & 1024 / 5 \\
7 & -- & 1024 / 4 & 1024 / 5 \\
8 & -- & -- & 2048 / 6 \\
9 & -- & -- & 2048 / 6 \\
\bottomrule
\end{tabular}
}%
\end{table}

\acro{} is built upon a multi-layer ResNeXt1D architecture, where each stage contains multiple residual blocks. We implement three variants, \ie small, medium, and large, for our proposed~\acro{} model with different computational complexities. All variants follow the same architectural principles but differ in depth, width, and parameter count. 

{\bf Convolution layer.} The network begins with an initial convolutional layer with $1$ channel, a kernel size of $16$, and a stride of $2$. This layer takes the input ECG signal $\mathbf{e}_s^l \in \mathbb{R}^{t}$ from the $s$-th sample and $l$-th lead, where $t$ denotes the signal length. The input $\mathbf{e}_s^l$ is first transformed by the layer and passed through a swish activation, yielding the hidden state $\mathbf{H}_s^l \in \mathbb{D}^{d \times \ell}$, where $\ell=\frac{t}{2}$ is the sequence length, and the hidden dimension $d$ is set to $32$, $64$, or $128$ for small, medium, and large variants, respectively. For brevity, we omit the sample \& lead indices $s$ and $l$ in the intermediate representations.

{\bf Residual stages.}
The network then processes $\mathbf{H}_s^l$ through a sequence of $\varpi$ residual stages. Each stage consists of $\gamma$ residual blocks. Each residual block processes features through three convolution layers sequentially:

(1) A $1 \!\times \!1$ convolution layer with stride $1$ transforms $\mathbf{H}_s^l$ to an intermediate dimension $\mathbf{H}^{(1)}\!\in\!\mathbb{R}^{d^{(1)}\!\times\!\ell}$, where $d^{(1)} = d\!\times\!\iota$, with $\iota\!\in\!\{0.5, 1.0, 1.5\}$ is a predefined ratio parameter for small, medium, and large models, respectively. 

(2) A $16\!\times\!16$ convolution layer with stride $1$ is applied to $\mathbf{H}^{(1)}$, producing the layer output $\mathbf{H}^{(2)}\!\in\!\mathbb{R}^{d^{(1)}\!\times\!\ell}$. Specifically, the input is first divided into $d^{(1)}/g$ groups along channel dimension, where $g\!\in\!\{8, 16, 32\}$ specifies the group width for the small, medium, and large variants, respectively. Each group is convolved independently with its own set of filters, and the resulting features are concatenated along the channel dimension to form the output $\mathbf{H}^{(2)}$. 

(3) Finally, a $1\!\times\!1$ convolution layer maps $\mathbf{H}^{(2)}$ to the output tensor $\mathbf{H}^{(3)}\!\in\!\mathbb{R}^{h\!\times\!\ell}$, where $h$ denotes the final output channel dimension of the residual block. 

A swish activation is used between layers. Following the final convolutional output of the last block of each stage, the output matrix $\mathbf{H}^{(3)}$ is first averaged along sequence dimension, forming vector $\mathbf{h}_s\!\in\!\mathbb{R}^{h}$. This vector $\mathbf{h}_s$ is passed through a swish-activated $2$-layer perception (hidden dimension is $\frac{h}{2}$, output dimension is $h$), followed by a sigmoid activation, forming channel-wise gating scores $\mathbf{h}_r\!\in\!\mathbb{R}^h$. Each channel vector of $\mathbf{H}^{(3)}$ is multiplied by the corresponding element of $\mathbf{h}_r$, and the reweighted channel vectors collectively form the residual tensor $\mathbf{H}_r \in \mathbb{R}^{h \times \ell}$. This residual term is then added with $\mathbf{H}^{(3)}$ to form the output of the state, given by $\mathbf{H}_o = \mathbf{H}^{(3)}+\mathbf{H}_r$. The final output of the last stage is averaged over the sequence dimension to obtain the feature embedding of the input signal, denoted as $\mathbf{z}\!\in\!\mathbb{R}^h$. \tableautorefname~\ref{apptab:resnet1d_configs} details the component parameters of each variant:

\subsection{Baseline Models}
\label{appsubsec:baseline_models}

{\bf ST-MEM}~\citep{na2024guiding} is a self-supervised learning framework specifically designed to capture spatio-temporal dependencies in 12-lead ECG signals. It employs a spatio-temporal masked auto-encoder that reconstructs randomly masked signal segments, enabling the model to learn rich ECG representations.

{\bf KED}~\citep{tian2024foundation} is an ECG foundation model pretrained with raw ECG signal and textual input (ECG reports). They used contrastive learning to align the representation of two modalities, enhancing diagnostic performance.

{\bf ECGFounder}~\citep{li2024electrocardiogram} is a foundation model for ECG trained in a supervised manner on large-scale labeled datasets for multi-label classification. To address the challenge of missing labels that are commonly present in long-tailed clinical datasets, the authors propose a modified loss function that enables more robust and balanced representation learning despite label sparsity.

{\bf Moirai}~\citep{woo2024unified} is a foundation model for time series designed for general-purpose forecasting. It supports both univariate and multivariate forecasting. For multivariable input, to get an aligned sequence, they reshape the input sequence into a single, aligned sequence that integrates both temporal and variable dimensions, enabling unified processing through the transformer architecture.

{\bf Moment}~\citep{goswami2024moment} is a transformer-based model built on the T5 architecture, designed for univariate time series tasks. It supports both sequence-to-sequence forecasting and time series classification, offering a unified and scalable foundational framework for temporal modeling.

\subsection{Baseline pretraining methods}
\label{appsubsec:base_pretraining}

{\bf SimCLR}~\citep{chen2020simple} is a contrastive learning framework whose objective is to bring the representations of augmented views of the same sample closer together, while pushing apart representations of different samples. It relies on a large set of diverse negative samples, which are typically achieved through large batch sizes, to provide effective contrastive learning signals.

{\bf BYOL}~\citep{grill2020bootstrap} is a self-supervised learning method that does not require negative samples. It proposed to have an online network and a target network. The online network is trained to predict the target network’s representation of the same augmented input. The target network is updated via an exponential moving average of the online parameters.

{\bf MoCo}~\citep{he2020momentum} is also a modified contrastive learning framework. Instead of relying on negatives sampled directly from the dataset, they propose to have a dynamic memory bank that stores encoded representations from previous batches. This memory bank is updated using a momentum encoder, which ensures stable and consistent representations over time. The stored representations serve as negative samples for contrastive learning, providing a large and diverse set of negatives without requiring large batch sizes.

\subsection{Details in obtaining SCORE2 risk scores}
\label{appsubsec:obtain_riskscore}

The SCORE2 risk score estimates the $10$-year risk of fatal and non-fatal cardiovascular disease.\footnote{SCORE2 risk score is obtained following the implementation available at: \href{https://github.com/dvicencio/RiskScorescvd}{\texttt{github.com/dvicencio/RiskScorescvd}}.} Demographic information (\eg sex and age) and clinical measurements (\eg blood pressure) are used to predict cardiovascular risk~\citep{score22021score2,score2-op2021score2-op}. Specifically, for sample $s$, the score uses $7$ features, including: age ($\texttt{a}^1_s$, in years), gender ($\texttt{a}^2_s\!\in\!\{\text{male}, \text{female}\}$), smoking status ($\texttt{a}^3_s\!\in\!\{0, 1\}$), systolic blood pressure (SBP, $\texttt{a}^4_s$ in mmHg), diabetes status ($\texttt{a}^5_s\!\in\!\{0, 1\}$), total cholesterol ($\texttt{a}^6_s$ in mmol/L), and high-density lipoprotein (HDL) cholesterol ($\texttt{a}^7_s$ in mmol/L).

\paragraph{Detailed risk score computation.} The $7$ covariates are first scaled by subtracting their corresponding mean values and then scaling by their pre-defined units (\eg, $5$ years for age, $20$ mmHg for SBP), except for the $3$ binary features, gender, smoking status, and diabetes status. Given sample $s$, the transformed variables are given as:
\begin{equation}
u_s^1 := \frac{\texttt{a}^1_s - 60}{5}, \quad
u_s^2 := \texttt{a}^3_s, \quad
u_s^3 := \frac{\texttt{a}^4_s - 120}{20}, \quad
u_s^4 := \texttt{a}^5_s, \quad
u_s^5 := \texttt{a}^6_s - 6, \quad
u_s^6 := \frac{\texttt{a}^7_s - 1.3}{0.5} \, \, .
\end{equation}
These normalized covariates, collected in vector $\mathbf{u} = [u_s^1, u_s^2, \cdots, u_s^6]\!\in\!\mathbb{R}^6$, are then transformed with parameters $\mathbf{b}_1 = [\beta_1, \beta_2, \beta_3, \beta_4, \beta_5, \beta_6]\!\in\!\mathbb{R}^6$, and age-dependent transformation $\mathbf{b}_2 = [0, \beta_7, \beta_8, \beta_{9}, \beta_{10}, \beta_{11}]\!\in\!\mathbb{R}^6$, given by:
\begin{equation}
    \chi_s = \mathbf{b}_1 \mathbf{u}_s^\top + u_1 \mathbf{b}_2 \mathbf{u}_s^\top \, \, ,
\end{equation}
where $\chi$ marks an intermediate score.
Different parameter sets $\mathbf{b}_1$ and $\mathbf{b}_2$ are used for males and females, and also for different age groups. Details are tabulated in~\tableautorefname~\ref{tab:score2_coefficients}. 
\begin{table}[!t]
\centering
\small
\renewcommand{\arraystretch}{0.92}
\caption{SCORE2 model coefficients ($\beta_i$) by gender and age group ($<$70 vs. $\geq$70 years).}
\label{tab:score2_coefficients}
\begin{tabular}{ccccc}
\toprule
\bf Coefficient & \bf Male $<$ 70 & \bf Female $<$ 70 & \bf Male $\geq$ 70 & \bf Female $\geq$ 70 \\
\midrule
$\beta_1$ & 0.3742 & 0.4648 & 0.0634 & 0.0789 \\
$\beta_2$ & 0.6012 & 0.7744 & 0.3524 & 0.4921 \\
$\beta_3$ & 0.2777 & 0.3131 & 0.0094 & 0.0102 \\
$\beta_4$ & 0.6457 & 0.8096 & 0.4245 & 0.6010 \\
$\beta_5$ & 0.1458 & 0.1002 & 0.0850 & 0.0605 \\
$\beta_6$ & -0.2698 & -0.2606 & -0.3564 & -0.3040 \\
$\beta_7$ & -0.0755 & -0.1088 & -0.0247 & -0.0255 \\
$\beta_8$ & -0.0255 & -0.0277 & -0.0005 & -0.0004 \\
$\beta_9$ & -0.0281 & -0.0226 & 0.0073 & -0.0009 \\
$\beta_{10}$ & 0.0426 & 0.0613 & 0.0091 & 0.0154 \\
$\beta_{11}$ & -0.0983 & -0.1272 & -0.0174 & -0.0107 \\
$S_0(t)$ & 0.9605 & 0.9776 & 0.7576 & 0.8082 \\
$c$ & 0 & 0 & 0.0929 & 0.2290 \\
\bottomrule
\end{tabular}
\end{table}

The health risk score $r$ of a patient sample is then obtained using an exponential transformation of $\chi$:
\begin{equation}
r_s = 1 - S_0(t)^{\exp(\chi_s-c)} \, \, ,
\end{equation}
where $S_0(t)$ is the baseline survival function at time $t$ (10 years) and $c$ is an offset. 
Different $S_0(t)$ and $c$ are also used for different gender and age groups. Details are tabulated in~\tableautorefname~\ref{tab:score2_coefficients}. Note that the original SCORE2 calculation includes a step for regional calibration. As regional information is unavailable in our dataset, this step was omitted. We directly used the uncalibrated 10-year risk score, with no further region-specific adjustment applied.

The SCORE2 risk score initially uses $7$ variables, but because MIMIC-IV-ECG only records the age, gender, and systolic BP, we infer missing values with simple imputation strategies. Smoking and diabetes status were assumed to be absent if not recorded (\ie, set to $0$). The missing total cholesterol and HDL cholesterol values were estimated using population-based reference values stratified by sex, with added Gaussian noise to reflect natural biological variation. Specifically, 
the total cholesterol was imputed as $5.2 \pm 0.5\ \text{mmol/L}$ and HDL cholesterol as $1.3 \pm 0.2\ \text{mmol/L}$, based on the mean values of age group 40-45~\citep{score22021score2}.
Additionally, there are $15$ records with missing age, which we impute as $40$. The number of missing values is accounted for in the \textbf{handling missing metadata} step described in~\sectionautorefname~\ref{subsec:risk_score}.

\begin{table}[t]
    \centering
    \small
    \caption{Results for finetuning on lead I (upper table) and lead II (lower table) ECGs with confidence interval. AUROCs are reported for seven clinical tasks:
    \briefExplain}
    \label{tab:confidence_interval}
    \setlength{\tabcolsep}{5pt}
    \setlength{\aboverulesep}{0pt}
    \setlength{\belowrulesep}{0pt}
    \resizebox{\columnwidth}{!}{%
    
    \setlength{\extrarowheight}{2pt}
    \begin{tabular}{lcc ccccc}
    \toprule
    {Dataset}  & MIMIC-IV & \multicolumn{5}{c}{ PTB-XL} & Chapman \\
    {Task} & LVEF & Dx & SubDx & SupDx & Form & Rhyth & Arrhy \\
    \cmidrule(lr){2-2} \cmidrule(lr){3-7} \cmidrule(lr){8-8}

    \bf Lead I \\

    \acro-S (448K) & 0.8276$\pm0.007$ & 0.8473$\pm0.011$ & 0.8515$\pm0.009$ & 0.8311$\pm0.007$ & 0.7632$\pm0.047$ & 0.9515$\pm0.011$ & 0.9001$\pm0.003$ \\
    \acro-M (30.7M) & 0.8230$\pm0.010$ & 0.8499$\pm0.011$ & 0.8581$\pm0.003$ & 0.8335$\pm0.005$ & 0.7708$\pm0.019$ & 0.9415$\pm0.006$ & 0.9074$\pm0.003$ \\
    \acro-L (296M) & 0.7875$\pm0.015$ & 0.8533$\pm0.005$ & 0.8543$\pm0.008$ & 0.8320$\pm0.004$ & 0.7732$\pm0.032$ & 0.9442$\pm0.007$ & 0.9041$\pm0.004$ \\

    \midrule

    \bf Lead II \\

    \acro-S (448K) & 0.8186$\pm0.004$ & 0.8284$\pm0.008$ & 0.8515$\pm0.004$ & 0.8475$\pm0.003$ & 0.7727$\pm0.042$ & 0.9510$\pm0.011$ & 0.9026$\pm0.003$ \\
    \acro-M (30.7M) & 0.8115$\pm0.005$ & 0.8361$\pm0.004$ & 0.8615$\pm0.005$ & 0.8482$\pm0.004$ & 0.7849$\pm0.025$ & 0.9531$\pm0.011$ & 0.9098$\pm0.003$ \\
    \acro-L (296M) & 0.7850$\pm0.032$ & 0.8333$\pm0.009$ & 0.8604$\pm0.011$ & 0.8489$\pm0.007$ & 0.7836$\pm0.020$ & 0.9538$\pm0.005$ & 0.9094$\pm0.005$ \\
    
    \bottomrule
    \end{tabular}
 }%
\end{table}

\begin{table}[b]
    \caption{Results for finetuning on 12-lead ECGs. AUROCs are reported, with numbers in brackets denoting the change relative to the corresponding single-lead baselines using lead I (in left hand side of \tableautorefname~\ref{tab:performance_single_lead}), where ($\uparrow$) and ($\downarrow$) denote a performance gain and degradation, respectively. A grey background denotes that the method was outperformed by \acro. 
     ~\briefExplain}
     \renewcommand{\arraystretch}{0.92}
    \label{tab:performance_12lead}
    \centering
    \setlength{\tabcolsep}{6pt}
    \resizebox{\columnwidth}{!}{%
    \small
    \begin{tabular}{lc c ccccc c}
    \toprule
    \multirow{2}{*}{\bf Model} & \bf MIMIC-IV & \multicolumn{5}{c}{\bf PTB-XL} & \bf Chapman \\
    & LVEF & Dx & SubDx & SupDx & Form & Rhyth. & Arrhy. \\
    \cmidrule(lr){2-2} \cmidrule(lr){3-7} \cmidrule(lr){8-8}
    ST-MEM & \cellcolor{gray!20}{.8095 ($\uparrow$.08)} & .8340 ($\uparrow$.23) & \cellcolor{gray!20}{.8261 ($\uparrow$.13)} & .8459 ($\uparrow$.11) & \cellcolor{gray!20}{.6716 ($\uparrow$.26)} & \cellcolor{gray!20}{.8419 ($\uparrow$.12)} & \cellcolor{gray!20}{.8462 ($\uparrow$.03)} \\
    KED & .8704 ($\uparrow$.08) & .9293 ($\uparrow$.15) & .9079 ($\uparrow$.09) & .9250 ($\uparrow$.10) & .8549 ($\uparrow$.32) & .9683 ($\uparrow$.11) & .9419 ($\uparrow$.05) \\
    Moment & \cellcolor{gray!20}{.7884 ($\downarrow$.02)} & .8995 ($\uparrow$.14) & .8727 ($\uparrow$.12) & .8926 ($\uparrow$.08) & .7874 ($\uparrow$.28) & .9528 ($\uparrow$.09) & .9298 ($\uparrow$.04) \\
    ECGFounder & .8748 ($\uparrow$.05) & .9107 ($\uparrow$.14) & .9106 ($\uparrow$.10) & .9085 ($\uparrow$.08) & .8046 ($\uparrow$.18) & \cellcolor{gray!20}{.9321 ($\uparrow$.01)} & .9362 ($\uparrow$.07) \\
    
    \bottomrule
    \end{tabular}
    }%
    
\end{table}

\section{Supplementary Evaluations}
\label{appsec:supp_results}

\subsection{Detailed confidence interval results}
\label{appsubsec:confidence_interval}

This section supplements the results reported in~\sectionautorefname~\ref{sec:results}.
To quantify uncertainty for the proposed~\acro{} model, we conducted $6$ rounds of experiments with the prediction head initialized with different seeds `\texttt{10, 42, 111, 123, 1111, 1234}', and computed $95\%$ confidence intervals for all reported metrics. \tableautorefname~\ref{tab:confidence_interval} enumerates the results, where for the majority of the tasks, most confidence intervals remain tight $(±0.003-0.011)$, apart from Form classification in the PTB-XL dataset, which shows larger variability across seeds. It can also be observed that the selected seed `\texttt{42}' yields relatively conservative results compared to other initializations.

\subsection{Comparison with finetuning on 12-Lead ECGs}
\label{appsubsec:upperbound12lead}

\begin{table}[t]
    \centering
    \renewcommand{\arraystretch}{.92}
    \caption{Results for linear probing on lead I ECG representations expressed as AUROCs. \briefExplain~\bestResults}
    \label{tab:linear_single_lead_auroc}
    \setlength{\tabcolsep}{6pt}
    \resizebox{\columnwidth}{!}{%
    \small
    \begin{tabular}{lc ccccc ccc}
    \toprule
    \multirow{2}{*}{\bf Model} & \bf MIMIC-IV & \multicolumn{5}{c}{\bf PTB-XL} & \bf Chapman & \multicolumn{2}{c}{\bf Icentia11K} \\
    & LVEF & Dx & SubDx & SupDx & Form & Rhyth & Arrhy & Beat & Rhyth \\
    \cmidrule(lr){2-2} \cmidrule(lr){3-7} \cmidrule(lr){8-8} \cmidrule(lr){9-10}
    Moirai & .5000 & .4989 & .5000 & .5000 & .5000 & .5000 & .4988 & .5002 & .4927 \\
    Moment & .7537 & .6392 & .6932 & .7611 & .5863 & .6227 & .7292 & .7298 & .6434 \\
    ST-MEM & .5709 & .5652 & .5872 & .6189 & .5300 & .5374 & .6587 & .6640 & .5605 \\
    KED & .7607 & .5935 & .6883 & .7514 & .5333 & .7372 & .7712 & .6984 & .7221 \\
    \midrule
     \acro-S & .7728 & .6005 & .6509 & .7503 & .5266 & .5872 & .6913 & .6581 & .6761 \\
     \acro-M & .7864 & .7256 & \underline{.7736} & .7903 & \underline{.6146} & \underline{.8077} & .7937 & .7302 & .7302 \\
     \acro-L & \underline{.7939} & \underline{.7388} & .7713 & \underline{.7971} & .6110 & .7966 & \underline{.8010} & \underline{.7843} & \underline{.7479} \\
     \midrule
     ECGFounder & \bf .8228 & \bf .7945 & \bf .8423 & \bf .8443 & \bf .7412 & \bf .9422 & \bf .8428 & \bf .9692 & \bf .9721 \\
    \bottomrule
    \end{tabular}
    }%
\end{table}

To better quantify the performance upper bounds for downstream tasks, we compare single-lead results for~\acro{} against baseline methods that use full 12-lead ECG data. Table \ref{tab:performance_12lead} summarizes the finetuning results for baseline methods on 12-lead ECGs. Several 12-lead baselines fail to surpass the performance of the single-lead~\acro{} model (shown as gray-highlighted cells). Performance changes relative to the corresponding single-lead baselines are shown in parentheses. All models benefited from multi-lead inputs for all tasks, except for Moment, where there is a small reduction in performance on the LVEF classification task when using 12-lead data. This confirms that additional leads provide valuable complementary information.

\subsection{Linear Probing}
\label{appsubsec:linear_probe}

This section supplements the linear probing experiments in~\sectionautorefname~\ref{sec:results} in the main paper, where we analyzed the performance of~\acro-M linear probing against the baselines. \tableautorefname~\ref{tab:linear_single_lead_auroc} presents the detailed performance for~\acro-M and further results for~\acro-S and~\acro-L, comparing with the baseline models. Among the $3$ variants,~\acro-L achieves the highest linear probing performance with a $7.4\%$ improvement over best baselines on average across $9$ downstream classification tasks. The second best being~\acro-M ($6.0\%$ improvement), while \acro{}-S exhibits a $7.3\%$ performance deficit. This descending trend indicates that larger models develop more expressive and generalizable representations during pretraining, which can be effectively leveraged through simple linear classification heads.

Notably, comparing results in~\tableautorefname~\ref{tab:linear_single_lead_auroc} with~\tableautorefname~\ref{tab:performance_single_lead}, \acro{}-M generally achieves the best balance of capacity and data efficiency during fine-tuning, where it outperforms \acro{}-L on many tasks, particularly those with limited labeled data. This potentially reflects a capacity–data mismatch under shared fine-tuning hyperparameters. The 296M-parameter \acro{}-L requires stronger regularization and/or more labeled data for stable end-to-end fine-tuning. When it is given with relatively small downstream datasets (especially on wearable and ED tasks), it is more prone to overfitting or reaching optimization plateaus when using the same fine-tuning recipe as \acro{}-M. Nevertheless, the stronger linear-probe performance of \acro{}-L indicates that its representations are of higher quality.

Additionally, the supervised model ECGFounder outperformed all three variants of \acro{}, with~\acro-M AUROCs an average of $12.7\%$ lower than those of ECGFounder. This is expected, as ECGFounder was pretrained on a much larger labeled dataset, whereas \acro{} relies purely on self-supervised learning.

\subsection{Effectiveness of risk-guided pretraining on baseline models}
\label{appsubsec:pretrain_over_baseline}

\begin{table}[b]
\renewcommand{\arraystretch}{.92}
    \caption{Results when retraining the backbone models with clinically-guided pretraining and finetuning, using lead I ECG. Numbers in brackets denote the change relative to the corresponding single-lead baselines using lead I, with (\textcolor{teal}{$\uparrow$}) denoting a performance gain and ($\downarrow$) a degradation. \briefExplain~For better clarity, we highlighted cases where performance is better in \textcolor{teal}{teal}.}
    \label{tab:pretrain_on_base}
    \centering
    \small
    \setlength{\tabcolsep}{6pt}
    \resizebox{\textwidth}{!}{%
    \begin{tabular}{lc c ccccc ccc}
    \toprule
    \multirow{2}{*}{\bf Model} & \bf MIMIC-IV & \multicolumn{5}{c}{\bf PTB-XL} & \bf Chapman \\
    & LVEF & Dx & SubDx & SupDx & Form & Rhyth & Arrhy \\ 
    \cmidrule(lr){2-2} \cmidrule(lr){3-7} \cmidrule(lr){8-8} 
        ST-MEM     & .7788 (\textcolor{teal}{$\uparrow$ .00}) & .7690 ($\downarrow$ .01) & .7568 ($\downarrow$ .01) & .7878 (\textcolor{teal}{$\uparrow$ .01}) & .5468 ($\downarrow$ .04) & .8078 (\textcolor{teal}{$\uparrow$ .07}) & .8046 ($\downarrow$ .01)  \\
        KED        & .8267 ($\downarrow$ .01) & .8729 (\textcolor{teal}{$\uparrow$ .04}) & .8569 (\textcolor{teal}{$\uparrow$ .03}) & .8330 (\textcolor{teal}{$\uparrow$ .00}) & .7481 (\textcolor{teal}{$\uparrow$ .12}) & .9067 (\textcolor{teal}{$\uparrow$ .02}) & .8956 (\textcolor{teal}{$\uparrow$ .01})  \\
        ECGFounder & .8263 ($\downarrow$ .03) & .8515 (\textcolor{teal}{$\uparrow$ .01}) & .8649 (\textcolor{teal}{$\uparrow$ .02}) & .8322 ($\downarrow$ .01) & .7957 (\textcolor{teal}{$\uparrow$ .04}) & .9516 (\textcolor{teal}{$\uparrow$ .00}) & .9038 ($\downarrow$ .01)  \\

         \bottomrule
    \end{tabular}
    }%
\end{table}

\begin{table}[t]
\renewcommand{\arraystretch}{.92}
    \centering
    \caption{Results on finetuning~\acro{} variants pretrained on the corresponding lead. The parentheses indicate the lead on which the model was pretrained. AUROCs are shown for Lead I (left) and Lead II (right). A \textcolor{teal}{teal} text denotes improved performance compared to~\tableautorefname~\ref{tab:performance_single_lead}.
    \briefExplain}
    \label{tab:performance_pretrain_single_lead}
    \setlength{\tabcolsep}{1pt}
    \resizebox{\columnwidth}{!}{%
    \small
    \begin{tabular}{lcc ccccc}
    \toprule
    \multirow{2}{*}{\bf Task} & \bf MIMIC-IV & \multicolumn{5}{c}{\bf PTB-XL} & \bf Chapman \\
    & LVEF & Dx & SubDx & SupDx & Form & Rhyth & Arrhy \\
    \cmidrule(lr){2-2} \cmidrule(lr){3-7} \cmidrule(lr){8-8}

    \acro$^\texttt{I}$-S & 
    \textcolor{teal}{.8215} & 
    \textcolor{teal}{.8536} & 
    \textcolor{teal}{.8604} & 
    .8324 & 
    \textcolor{teal}{.7958} & 
    \textcolor{teal}{.9449} & 
    \textcolor{teal}{.9057} \\
    
    \acro$^\texttt{I}$-M & 
    \textcolor{teal}{.8170} & 
    \textcolor{teal}{.8505} & 
    \textcolor{teal}{.8575} & 
    .8336 & 
    \textcolor{teal}{.7805} & 
    \textcolor{teal}{.9476} & 
    \textcolor{teal}{.9128} \\
    
    \acro$^\texttt{I}$-L & 
    .7821 & 
    \textcolor{teal}{.8606} & 
    \textcolor{teal}{.8564} & 
    \textcolor{teal}{.8327} & 
    \textcolor{teal}{.7870} & 
    \textcolor{teal}{.9513} & 
    \textcolor{teal}{.9133} \\
    
    \bottomrule
    \end{tabular}
    \quad
    \begin{tabular}{lcc ccccc}
    \toprule
    \multirow{2}{*}{\bf Task} & \bf MIMIC-IV & \multicolumn{5}{c}{\bf PTB-XL} & \bf Chapman \\
    & LVEF & Dx & SubDx & SupDx & Form & Rhyth & Arrhy \\
    \cmidrule(lr){2-2} \cmidrule(lr){3-7} \cmidrule(lr){8-8} 

    \acro$^\texttt{II}$-S & 
    \textcolor{teal}{.8207} & 
    \textcolor{teal}{.8386} & 
    \textcolor{teal}{.8620} & 
    \textcolor{teal}{.8497} & 
    \textcolor{teal}{.8001} & 
    \textcolor{teal}{.9552} & 
    \textcolor{teal}{.9107} \\
    
    \acro$^\texttt{II}$-M & 
    \textcolor{teal}{.8102} & 
    \textcolor{teal}{.8349} & 
    \textcolor{teal}{.8719} & 
    \textcolor{teal}{.8523} & 
    \textcolor{teal}{.7921} & 
    \textcolor{teal}{.9552} & 
    \textcolor{teal}{.9135} \\
    
    \acro$^\texttt{II}$-L & 
    .7862 & 
    \textcolor{teal}{.8421} & 
    \textcolor{teal}{.8680} & 
    \textcolor{teal}{.8464} & 
    \textcolor{teal}{.7829} & 
    .9257 & 
    \textcolor{teal}{.9134} \\

    \bottomrule
    \end{tabular}
    }%
\end{table}

\begin{table}[!b]
\centering
\small
\caption{``Risk score'' of CIFAR-100 superclasses and subclasses. Subclasses within the same superclass have similar risk scores, while more distant superclasses have larger dissimilar scores.}
\label{tab:cifar100-risk-scores}
\setlength{\tabcolsep}{10pt}
\begin{tabular}{llc}
\toprule
\textbf{Superclass} & \textbf{Subclass} & \textbf{Risk Score} \\
\midrule
\multirow{4}{*}{Natural Environment (0.05--0.20)} 
  & Large natural outdoor scenes & 0.05 \\
  & Trees & 0.10 \\
  & Flowers & 0.15 \\
  & Fruit and vegetables & 0.20 \\
\midrule
\multirow{2}{*}{Aquatic Life (0.25--0.30)} 
  & Fish & 0.25 \\
  & Aquatic mammals & 0.30 \\
\midrule
\multirow{2}{*}{Invertebrates (0.35--0.40)} 
  & Insects & 0.35 \\
  & Non-insect invertebrates & 0.40 \\
\midrule
\multirow{6}{*}{Land Mammals (0.45--0.70)} 
  & Small mammals & 0.45 \\
  & Medium mammals & 0.50 \\
  & Reptiles & 0.55 \\
  & Large omnivores and herbivores & 0.60 \\
  & Large carnivores & 0.65 \\
  & People & 0.70 \\
\midrule
\multirow{6}{*}{Human-Made Objects (0.75--1.00)} 
  & Food containers & 0.75 \\
  & Household furniture & 0.80 \\
  & Household electrical devices & 0.85 \\
  & Large man-made outdoor things & 0.90 \\
  & Vehicles 1 & 0.95 \\
  & Vehicles 2 & 1.00 \\
\bottomrule
\end{tabular}
\end{table}

This section supplements the experiment in~\sectionautorefname~\ref{sec:results} on applying our clinically-guided contrastive loss to the best semi-supervised baseline, KED. We further extend our analysis by applying the proposed pretraining approach to other baseline models, thereby testing its generalizability across architectures. Each baseline model has been initialized with its original pretrained weights and pretrained with our proposed clinically-guided pretraining approach. The AUROC results are presented in~\tableautorefname~\ref{tab:pretrain_on_base}, with values in the brackets indicating the changes relative to the finetuning results of the original model weights, as reported in~\tableautorefname~\ref{tab:performance_single_lead}. On average, pretraining with our proposed clinically-guided method yields an overall performance gain of $0.7\%$. Among the three baselines, KED shows the largest improvement, with its AUROC score increasing by $3.0\%$. This substantial improvement likely stems from KED's reliance on purely self-supervised learning without explicit clinical guidance, making it particularly receptive to our clinically-guided pretraining strategy that provides essential domain-specific knowledge.

\subsection{Supplements in pretraining with a specific lead}
\label{appsubsec:supp_pretraining_1lead}

This section supplements~\sectionautorefname~\ref{sec:results} on \textbf{pretraining on a specific lead}, which explores the upper bound performance by pretraining exclusively on lead I or lead II data, corresponding to the leads used in our downstream evaluation tasks. Although less generalizable, this method reduces potential domain shift compared with our primary pretraining approach using all $12$ leads.

\tableautorefname~\ref{tab:performance_pretrain_single_lead} present the results. The ones colored in red with an uparrow ($\uparrow$) denote a performance improvement compared to the performance of the corresponding-sized~\acro{} model pretrained on all $12$ leads, while those with a downarrow ($\downarrow$) indicate a performance degradation. Overall, lead-specific pretraining yields performance gains across all tasks, with an average improvement of $2.0\%$. Specifically, ~\acro$^\texttt{I}$-S, ~\acro$^\texttt{I}$-M, and ~\acro$^\texttt{I}$-L achieve improvements of $3.4\%$, $1.6\%$, and $1.0\%$, respectively, compared to their corresponding models pretrained on all leads across all tasks and both leads. Interestingly, the \textbf{Small} variant gives the largest performance gain. This may be because the reduced capacity of the smaller model helps prevent overfitting on the relatively limited data, allowing it to capture the essential patterns more effectively than the larger counterparts.

\subsection{Effectiveness of different loss components}
\label{appsubsec:toy_cifar100}

An ablation study is conducted using CIFAR-100~\citep{krizhevsky2009learning} (the representative computer vision benchmark) to understand the contribution of different loss components. This dataset consists of $100$ subclasses, each grouped into a higher-level superclass. While instances within the same superclass may belong to different subclasses, they are generally more similar to each other than to instances from different superclasses. To simulate the setup in informing the contrastive learning process with risk scores, we assign each CIFAR-100 superclass a risk score within the range $[0,1]$. The assigned scores are decided in a way such that subclasses belonging to the same superclass receive relatively close risk scores, while subclasses from more distinct superclasses are assigned increasingly dissimilar scores. The detailed scores given to each subclass are provided in~\tableautorefname~\ref{tab:cifar100-risk-scores}.

\begin{table}[t!]
    \centering
    \small
    \renewcommand{\arraystretch}{0.92}
    \caption{Cosine similarity statistics between class pairs. A spade symbol (\samerel) indicates that the two classes belong to the same superclass in the CIFAR-100 dataset, whereas a diamond symbol (\diffrel) represents pairs from different superclasses. Values are reported as mean $\pm$ standard deviation of cosine similarity computed over the sampled pairs.
    We also report the classification accuracy on both sub and superclass classification, which has $100$ and $20$ classes in total, with mean and standard deviation reported on $6$ different random seed initializations. Best classification results are \textbf{bolded}.
    }
    \label{tab:cifar100}
    \setlength{\tabcolsep}{12pt}
    \resizebox{\linewidth}{!}{%
    \begin{tabular}{lc c c c c c}
    \toprule
    \textbf{Class Pair} & \textbf{Groups} & $\mathcal{L}_{nce}$ & $\mathcal{L}^{w}$ & $\mathcal{L}^{d}$ & $\mathcal{L}_{nce} + \mathcal{L}^{d}$ & $\mathcal{L}^{w} + \mathcal{L}^{d}$ \\
    \cmidrule(lr){3-3} \cmidrule(lr){4-4} \cmidrule(lr){5-5} \cmidrule(lr){6-6} \cmidrule(lr){7-7}
    Palm -- Pine Tree & \samerel & .058 $\pm$ .125 & .081 $\pm$ .139 & .699 $\pm$ .214 & .067 $\pm$ .132 & .556 $\pm$ .153 \\
    Palm Tree -- Baby & \diffrel & .018 $\pm$ .060 & .025 $\pm$ .064 & .548 $\pm$ .131 & .016 $\pm$ .052 & .432 $\pm$ .075 \\
    Pine Tree -- Baby & \diffrel & .012 $\pm$ .052 & .016 $\pm$ .053 & .564 $\pm$ .136 & .010 $\pm$ .045 & .438 $\pm$ .075 \\
    
    \midrule
    Shark -- Trout & \samerel & .022 $\pm$ .081 & .024 $\pm$ .078 & .738 $\pm$ .104 & .019 $\pm$ .077 & .542 $\pm$ .074 \\
    Shark -- Bicycle & \diffrel & .014 $\pm$ .060 & .026 $\pm$ .067 & .527 $\pm$ .146 & .012 $\pm$ .051 & .385 $\pm$ .094 \\
    Trout -- Bicycle & \diffrel & .023 $\pm$ .073 & .038 $\pm$ .081 & .501 $\pm$ .137 & .019 $\pm$ .062 & .380 $\pm$ .100 \\
    \midrule
    Apple -- Pear & \samerel & .053 $\pm$ .125 & .062 $\pm$ .148 & .761 $\pm$ .110 & .083 $\pm$ .162 & .559 $\pm$ .116 \\
    Apple -- Bottle & \diffrel & .015 $\pm$ .055 & .017 $\pm$ .059 & .544 $\pm$ .107 & .030 $\pm$ .060 & .411 $\pm$ .081 \\
    Pear -- Bottle & \diffrel & .036 $\pm$ .099 & .038 $\pm$ .094 & .589 $\pm$ .121 & .035 $\pm$ .083 & .452 $\pm$ .089 \\
    
    \midrule
    Bridge -- House & \samerel & .065 $\pm$ .129 & .087 $\pm$ .136 & .762 $\pm$ .135 & .057 $\pm$ .119 & .561 $\pm$ .115 \\
    Bridge -- Sea & \diffrel & .044 $\pm$ .102 & .056 $\pm$ .113 & .311 $\pm$ .222 & .046 $\pm$ .107 & .284 $\pm$ .165 \\
    House -- Sea & \diffrel & .018 $\pm$ .064 & .021 $\pm$ .067 & .307 $\pm$ .217 & .016 $\pm$ .062 & .262 $\pm$ .154 \\
    
    \midrule
    
    Cloud -- Forest & \samerel & .013 $\pm$ .052 & .013 $\pm$ .048 & .654 $\pm$ .215 & .011 $\pm$ .050 & .482 $\pm$ .133 \\
    Cloud -- Bus & \diffrel & .015 $\pm$ .060 & .018 $\pm$ .055 & .220 $\pm$ .196 & .014 $\pm$ .053 & .201 $\pm$ .153 \\
    Forest -- Bus & \diffrel & .033 $\pm$ .086 & .041 $\pm$ .091 & .357 $\pm$ .264 & .025 $\pm$ .074 & .275 $\pm$ .181 \\

    \midrule
    \multicolumn{7}{@{}l@{}}{\textit{ResNet18} pretrained} \\
    \multicolumn{2}{l}{\quad Subclass (100)} & .621 $\pm$ .003 & .624 $\pm$ .005 & .635 $\pm$ .002 & .611 $\pm$ .004 & \textbf{.646} $\pm$ .006 \\
    \multicolumn{2}{l}{\quad Superclass (20)} & .638 $\pm$ .004 & .639 $\pm$ .004 & .646 $\pm$ .003 & .627 $\pm$ .009 & \textbf{.658} $\pm$ .007 \\
    \bottomrule
    \end{tabular}%
    }
    \vspace{-10pt}
\end{table}

\begin{figure}[!b]
    \centering
    \includegraphics[width=0.44\linewidth]{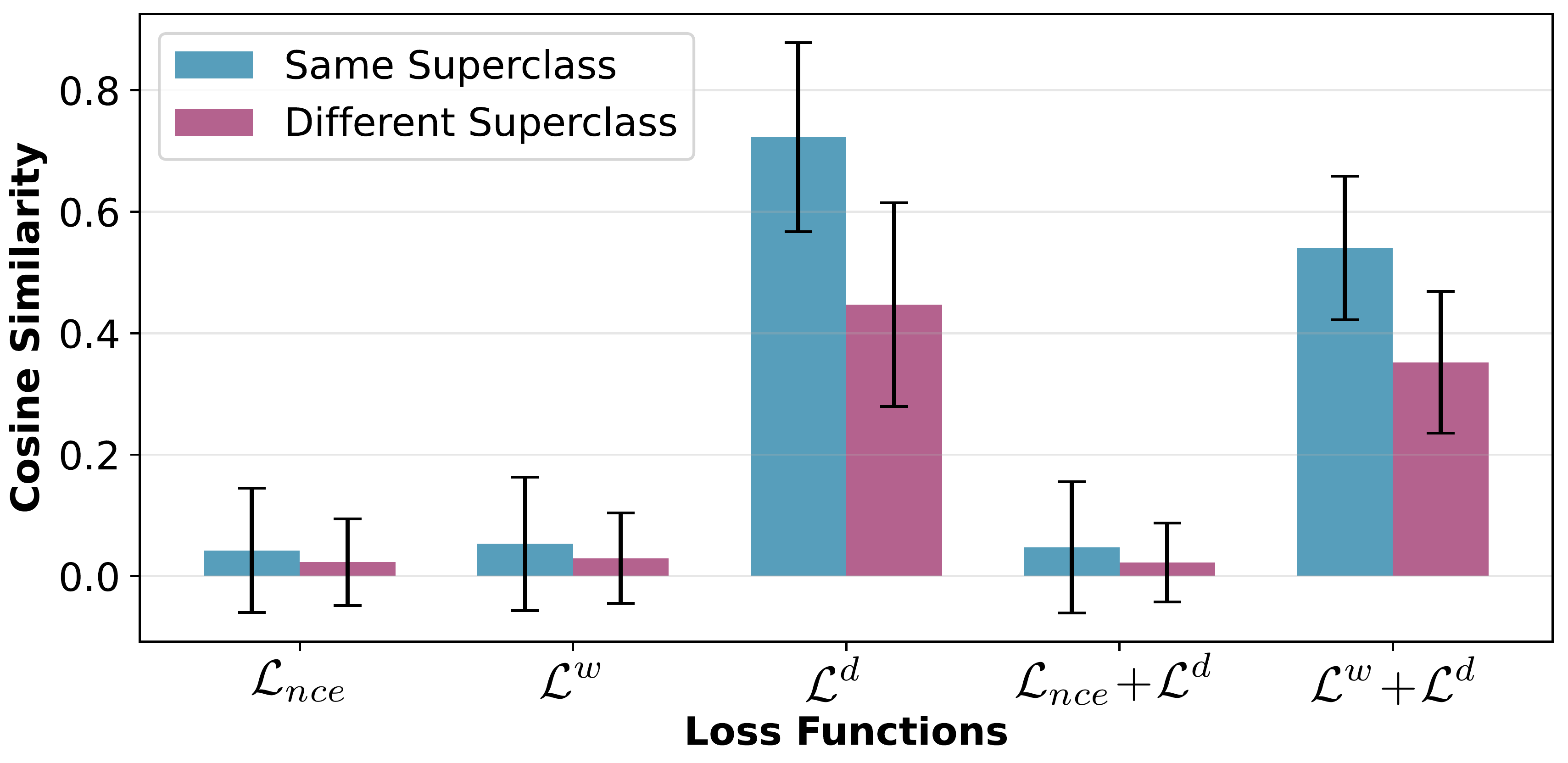}
    \quad \quad
    \includegraphics[width=0.44\linewidth]{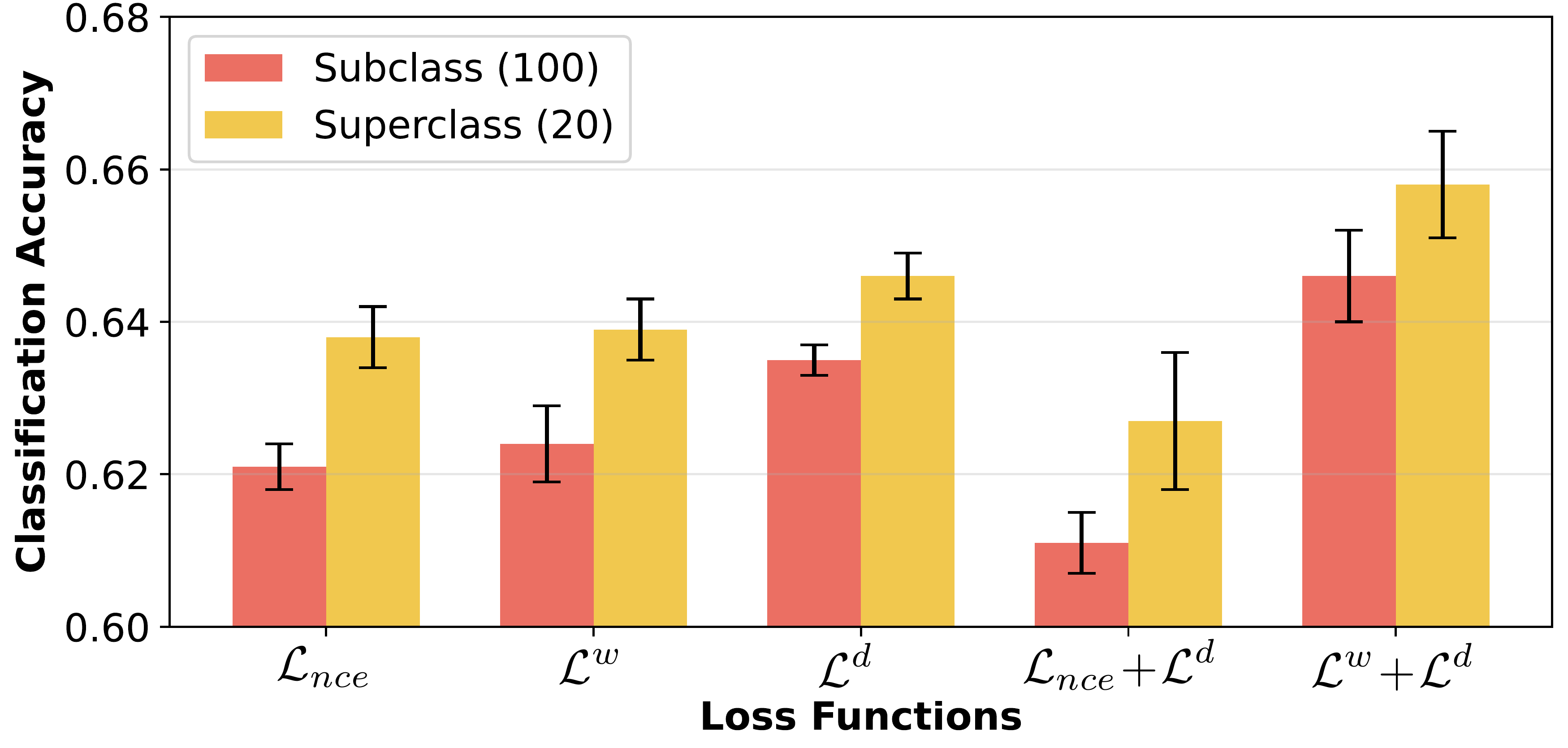}
     \caption{
     Ablation study on CIFAR-100 dataset. \textbf{Left:} Cosine similarity between samples from the same \vs different superclasses. Results averaged across $5$ randomly selected triplet groups, with $2$ of them from the same superclass.
     \textbf{Right:} Classification accuracy on CIFAR-100 subclass (100 classes) and superclass (20 classes) tasks.
     }
    \label{fig:cifar_summ}
\end{figure}

For the experiments, we pretrained the ResNet18 model (initialized with ImageNet-1K pretrained weights) using different loss functions, which are the standard contrastive loss $\mathcal{L}_{nce}$, the weighted contrastive loss $\mathcal{L}^w$, the dissimilarity alignment loss $\mathcal{L}^d$, $\mathcal{L}_{nce} + \mathcal{L}^{d}$, and our proposed $\mathcal{L}^w + \mathcal{L}^{d}$. The models were trained for $50$ epochs with a batch size of $256$, optimized using the AdamW optimizer with an initial learning rate of $1\times10^{-3}$. A Cosine Annealing learning rate schedule was used, and we stop the training when the training loss does not decrease over $10$ epochs.

{\bf Inter-class similarity analysis.}
We first analyzed the similarity relationships between groups by randomly selecting $5$ sets of triplet classes. Each triplet consisted of $2$ subclasses from the same superclass and one subclass from a different superclass. For each triplet, we obtain the similarity scores between all pairs of samples (resulting in $10{,}000$ scores per subclass pair, given $100$ samples per subclass), and report the mean and standard deviation across groups. Results are summarized on the left of~\figureautorefname~\ref{fig:cifar_summ} and detailed in the upper part of~\tableautorefname~\ref{tab:cifar100}. It can be observed that training the model with only $\mathcal{L}_{nce}$ yields minimal distinction between similarity scores of samples within the same superclass (intra-class) compared to those from different superclasses (inter-class). When training with $\mathcal{L}^{w}$, the inter-class differentiation becomes marginally more pronounced, although it remains insufficiently distinct. Meanwhile, training with $\mathcal{L}^{d}$ produces substantially more distinguishable inter-class similarity differences. However, this loss function enforces hard mapping of inter-class distances to align with risk scores without constraining intra-class clustering, consequently increasing the intra-class similarity variance. This indicates that samples of the same class are scattered rather than forming compact clusters, which is a less desirable property for downstream classification. Meanwhile, the combination of $\mathcal{L}_{nce}$ with $\mathcal{L}^{d}$ offered limited gains compared to $\mathcal{L}_{nce}$. In contrast, our proposed combination $\mathcal{L}^{w}+\mathcal{L}^{d}$ achieves notable differentiation in similarity scores whilst maintaining consistency, which can be inferred from the reduced standard deviation.

{\bf Classification accuracy on CIFAR-100.}
We also provide classification accuracy over the 100 subclasses and also 20 superclasses, experiment initialized with $6$ random seeds, which are `\texttt{10, 42, 111, 123, 1111, 1234}', and we present the average and standard deviation of the classification accuracy. Results are summarized on the right of~\figureautorefname~\ref{fig:cifar_summ} and enumerated in the lower part of~\tableautorefname~\ref{tab:cifar100}. For subclass classification (100 classes), our proposed $\mathcal{L}^{w} + \mathcal{L}^{d}$ combination achieves the highest accuracy of $64.6\% \pm 0.6\%$, outperforming all baseline approaches by at least ($1.7\%$). Notably, whilst $\mathcal{L}^{d}$ alone yields the second best performance ($63.5\% \pm 0.2\%$), the combination of $\mathcal{L}_{nce} + \mathcal{L}^{d}$ exhibits degraded classification accuracy by ($3.8\%$) ($61.1\% \pm 0.4\%$), indicating that simply combining the InfoNCE loss with the difference alignment loss would contrarily harm the model performance, which is consistent with our similarity score observations. For superclass classification (20 classes), the performance remains consistent, with our proposed method achieving $65.8\% \pm 0.7\%$, bringing a substantial improvement by ($1.9\%$) over the best-performance baseline ($64.6\% \pm 0.3\%$) and by $3.0\%$ compared to the na\"ive contrastive baseline ($63.8\% \pm 0.4\%$).

\subsection{Analysis over different loss weighting}

\begin{table}[!t]
\centering
\caption{Effect of weighting over loss. All combinations are normalized by dividing by the sum of their coefficients to maintain comparable loss scales. Experiments were conducted over all classification tasks. For the 12-lead ECG datasets, we use lead I as input. Table cells are colored where darker blue indicates better performance and lighter blue indicates worse performance. Best results are \textbf{bolded}.}
\label{apptab:weighting_loss}
\setlength{\tabcolsep}{2pt}
\resizebox{\textwidth}{!}{%
\renewcommand{\arraystretch}{.8}
\begin{tabular}{cccccccccccccc}
\toprule
\multirow{2}{*}{\bf Loss} 
& \bf MIMIC-IV 
& \multicolumn{5}{c}{\bf PTB-XL} 
& \bf Chapman 
& \bf MUSIC 
& \multicolumn{2}{c}{\bf Icentia11K} 
& \multicolumn{3}{c}{\bf MC-MED} \\
 & LVEF & Dx & SubDx & SupDx & Form & Rhyth & Arrhy & Outcome & Beat & Rhyth & ED Dispo & DC Dispo & Acuity \\
\cmidrule(lr){2-2} \cmidrule(lr){3-7} \cmidrule(lr){8-8} 
\cmidrule(lr){9-9} \cmidrule(lr){10-11} \cmidrule(lr){12-14}
$\mathcal{L}^d$
& \cellcolor{blue!10}0.8051 & \cellcolor{blue!5}0.7990 & \cellcolor{blue!5}0.8135 & \cellcolor{blue!5}0.8363 & \cellcolor{blue!5}0.6451 & \cellcolor{blue!5}0.9206 
& \cellcolor{blue!5}0.8808 & \cellcolor{blue!10}0.4767 & \cellcolor{blue!15}0.9803 & \cellcolor{blue!10}0.9251 
& \cellcolor{blue!5}0.6016 & \cellcolor{blue!10}0.6475 & \cellcolor{blue!15}0.6222 \\
$\mathcal{L}^w+5\mathcal{L}^d$
& \cellcolor{blue!15}0.8096 & \cellcolor{blue!15}0.8195 & \cellcolor{blue!20}0.8536 & \cellcolor{blue!15}0.8467 & \cellcolor{blue!15}0.6917 & \cellcolor{blue!20}0.9581 
& \cellcolor{blue!15}0.9028 & \cellcolor{blue!5}0.4747 & \cellcolor{blue!5}0.9780 & \cellcolor{blue!5}0.9224 
& \cellcolor{blue!10}0.6070 & \cellcolor{blue!5}0.6341 & \cellcolor{blue!5}0.6179 \\
$\mathcal{L}^w+2\mathcal{L}^d$ 
& \cellcolor{blue!5}0.7965 & \cellcolor{blue!20}0.8253 & \cellcolor{blue!15}0.8518 & \cellcolor{blue!10}0.8396 & \cellcolor{blue!20}0.7365 & \cellcolor{blue!15}0.9325 
& \cellcolor{blue!20}0.9088 & \cellcolor{blue!25}\textbf{0.5513} & \cellcolor{blue!5}0.9755 & \cellcolor{blue!15}0.9289 
& \cellcolor{blue!15}0.6148 & \cellcolor{blue!15}0.6416 & \cellcolor{blue!5}0.5861 \\
$\mathcal{L}^w+\mathcal{L}^d$
& \cellcolor{blue!20}0.8083 & \cellcolor{blue!25}\textbf{0.8292} & \cellcolor{blue!25}\textbf{0.8566} & \cellcolor{blue!20}0.8430 & \cellcolor{blue!15}0.7409 & \cellcolor{blue!15}0.9361 
& \cellcolor{blue!15}0.9061 & \cellcolor{blue!15}0.5304 & \cellcolor{blue!20}0.9792 & \cellcolor{blue!20}0.9328 
& \cellcolor{blue!25}\textbf{0.6397} & \cellcolor{blue!25}\textbf{0.6607} & \cellcolor{blue!25}\textbf{0.6510} \\
$\mathcal{L}^w+.5\mathcal{L}^d$
& \cellcolor{blue!25}\textbf{0.8204} & \cellcolor{blue!15}0.8271 & \cellcolor{blue!15}0.8503 & \cellcolor{blue!15}0.8473 & \cellcolor{blue!25}\textbf{0.7471} & \cellcolor{blue!15}0.9360 
& \cellcolor{blue!25}\textbf{0.9130} & \cellcolor{blue!20}0.5496 & \cellcolor{blue!25}\textbf{0.9858} & \cellcolor{blue!25}\textbf{0.9350} 
& \cellcolor{blue!5}0.6026 & \cellcolor{blue!20}0.6492 & \cellcolor{blue!20}0.6221 \\
$\mathcal{L}^w+.2\mathcal{L}^d$
& \cellcolor{blue!15}0.8118 & \cellcolor{blue!10}0.8158 & \cellcolor{blue!20}0.8520 & \cellcolor{blue!25}\textbf{0.8484} & \cellcolor{blue!15}0.7325 & \cellcolor{blue!25}\textbf{0.9593} 
& \cellcolor{blue!15}0.9032 & \cellcolor{blue!5}0.4747 & \cellcolor{blue!20}0.9794 & \cellcolor{blue!5}0.9211 
& \cellcolor{blue!10}0.6082 & \cellcolor{blue!10}0.6369 & \cellcolor{blue!15}0.6216 \\
$\mathcal{L}^w$
& \cellcolor{blue!5}0.8018 & \cellcolor{blue!5}0.7990 & \cellcolor{blue!10}0.8188 & \cellcolor{blue!10}0.8387 & \cellcolor{blue!10}0.6494 & \cellcolor{blue!10}0.9215 
& \cellcolor{blue!10}0.8838 & \cellcolor{blue!5}0.4701 & \cellcolor{blue!10}0.9786 & \cellcolor{blue!5}0.9241 
& \cellcolor{blue!5}0.6017 & \cellcolor{blue!5}0.6446 & \cellcolor{blue!10}0.6153 \\
\bottomrule
\end{tabular}}
\end{table}

In Appendix~\ref{appsubsec:toy_cifar100}, we used CIFAR-100 to investigate how different combinations of losses influence the cosine similarity between representation pairs and classification accuracy. In this section, we extend this analysis to ECG tasks to examine the effect of having different weighting between the two proposed losses, $\mathcal{L}^w$ and $\mathcal{L}^d$. Specifically,  we evaluated four configurations: (1) $\mathcal{L}^w$ only, (2) $\mathcal{L}^d$ only, (3) our default balanced weight $\mathcal{L}^w + \mathcal{L}^d$, and (4) unbalanced mixes with coefficients $\lambda = \{5, 2, 0.5, 0.2\}$ applied to $\mathcal{L}^w + \lambda\mathcal{L}^d$. All combinations have been normalized by dividing the sum of their corresponding coefficients to maintain comparable loss scales. Results are enumerated in~\tableautorefname~\ref{apptab:weighting_loss}. It can be observed that~\acro{} is not particularly sensitive to the loss weight ratio (with an averaged standard deviation of $0.016$ across all tasks), and the two components provide complementary guidance. Using balanced weights for $\mathcal{L}^w$ and $\mathcal{L}^d$ consistently provides strong performance, confirming our initial motivational analysis using CIFAR100.

\subsection{Analysis over pretraining parameters}
\label{appsubsec:parameter}

\begin{table*}[!b]
\centering
\caption{Sensitivity analysis regarding $\alpha$. Experiments were conducted over all classification tasks. For the 12-lead ECG datasets, we use lead I as input. Table cells are colored where darker blue indicates better performance and lighter blue indicates worse performance. Best results are \textbf{bolded}.}
\label{apptab:alpha}
\setlength{\tabcolsep}{2pt}
\resizebox{\textwidth}{!}{%
\renewcommand{\arraystretch}{.8}
\begin{tabular}{cccccccccccccc}
\toprule
\multirow{2}{*}{\bf $\alpha$} 
& \bf MIMIC-IV 
& \multicolumn{5}{c}{\bf PTB-XL} 
& \bf Chapman 
& \bf MUSIC 
& \multicolumn{2}{c}{\bf Icentia11K} 
& \multicolumn{3}{c}{\bf MC-MED} \\
 & LVEF & Dx & SubDx & SupDx & Form & Rhyth & Arrhy & Outcome & Beat & Rhyth & ED Dispo & DC Dispo & Acuity \\
\cmidrule(lr){2-2} \cmidrule(lr){3-7} \cmidrule(lr){8-8}
\cmidrule(lr){9-9} \cmidrule(lr){10-11}
\cmidrule(lr){12-14}
0.5 &
\cellcolor{blue!15}0.8050 & \cellcolor{blue!5}0.8084 & \cellcolor{blue!5}0.8452 & \cellcolor{blue!5}0.8357 & \cellcolor{blue!5}0.7111 & \cellcolor{blue!5}0.9212 &
\cellcolor{blue!5}0.8904 & \cellcolor{blue!5}0.4737 & \cellcolor{blue!10}0.9662 & \cellcolor{blue!15}0.9381 &
\cellcolor{blue!10}0.6053 & \cellcolor{blue!5}0.6112 & \cellcolor{blue!5}0.5694 \\
0.4 &
\cellcolor{blue!5}0.7983 & \cellcolor{blue!10}0.8117 & \cellcolor{blue!10}0.8550 & \cellcolor{blue!10}0.8419 & \cellcolor{blue!10}0.7291 & \cellcolor{blue!10}0.9213 &
\cellcolor{blue!10}0.8950 & \cellcolor{blue!10}0.4767 & \cellcolor{blue!5}0.9550 & \cellcolor{blue!5}0.9251 &
\cellcolor{blue!5}0.6019 & \cellcolor{blue!10}0.6475 & \cellcolor{blue!10}0.6407 \\
0.3 &
\cellcolor{blue!10}0.8030 & \cellcolor{blue!25}\textbf{0.8295} & \cellcolor{blue!25}\textbf{0.8699} & \cellcolor{blue!25}\textbf{0.8515} & \cellcolor{blue!20}0.7524 & \cellcolor{blue!15}0.9338 &
\cellcolor{blue!15}0.9023 & \cellcolor{blue!25}\textbf{0.5513} & \cellcolor{blue!25}\textbf{0.9848} & \cellcolor{blue!25}\textbf{0.9475} &
\cellcolor{blue!20}0.6260 & \cellcolor{blue!15}0.6491 & \cellcolor{blue!25}\textbf{0.6548} \\
\bf 0.2 &
\cellcolor{blue!20}0.8083 & \cellcolor{blue!20}0.8292 & \cellcolor{blue!15}0.8566 & \cellcolor{blue!15}0.8430 & \cellcolor{blue!15}0.7409 & \cellcolor{blue!25}\textbf{0.9361} &
\cellcolor{blue!20}0.9061 & \cellcolor{blue!15}0.5304 & \cellcolor{blue!20}0.9792 & \cellcolor{blue!10}0.9328 &
\cellcolor{blue!25}\textbf{0.6397} & \cellcolor{blue!25}\textbf{0.6607} & \cellcolor{blue!20}0.6510 \\
0.1 &
\cellcolor{blue!25}\textbf{0.8176} & \cellcolor{blue!15}0.8203 & \cellcolor{blue!20}0.8596 & \cellcolor{blue!20}0.8484 & \cellcolor{blue!25}\textbf{0.7625} & \cellcolor{blue!20}0.9360 &
\cellcolor{blue!25}\textbf{0.9119} & \cellcolor{blue!20}0.5476 & \cellcolor{blue!15}0.9733 & \cellcolor{blue!20}0.9397 &
\cellcolor{blue!15}0.6186 & \cellcolor{blue!20}0.6593 & \cellcolor{blue!15}0.6427 \\
\bottomrule
\end{tabular}
}%
\end{table*}

\begin{table*}[!t]
\centering
\caption{Sensitivity analysis regarding $\tau$. Experiments were conducted over all classification tasks. For the 12-lead ECG datasets, we use lead I as input. Table cells are colored where darker blue indicates better performance and lighter blue indicates worse performance. Best results are \textbf{bolded}.}
\label{apptab:tau}
\setlength{\tabcolsep}{2pt}
\resizebox{\textwidth}{!}{%
\renewcommand{\arraystretch}{.8}
\begin{tabular}{cccccccccccccc}
\toprule
\multirow{2}{*}{\bf $\tau$} 
& \bf MIMIC-IV 
& \multicolumn{5}{c}{\bf PTB-XL} 
& \bf Chapman 
& \bf MUSIC 
& \multicolumn{2}{c}{\bf Icentia11K} 
& \multicolumn{3}{c}{\bf MC-MED} \\
 & LVEF & Dx & SubDx & SupDx & Form & Rhyth & Arrhy & Outcome & Beat & Rhyth & ED Dispo & DC Dispo & Acuity \\
\cmidrule(lr){2-2} \cmidrule(lr){3-7} \cmidrule(lr){8-8} 
\cmidrule(lr){9-9} \cmidrule(lr){10-11} \cmidrule(lr){12-14}
\bf .07 & 
\cellcolor{blue!25}\textbf{.8083} & \cellcolor{blue!5}.8292 & \cellcolor{blue!15}.8566 & \cellcolor{blue!5}.8430 & \cellcolor{blue!5}.7409 & \cellcolor{blue!5}.9361 & \cellcolor{blue!25}\textbf{.9061} &
\cellcolor{blue!25}\textbf{.5304} & \cellcolor{blue!25}\textbf{.9792} & \cellcolor{blue!25}\textbf{.9328} & \cellcolor{blue!25}\textbf{.6397} & \cellcolor{blue!25}\textbf{.6607} & \cellcolor{blue!25}\textbf{.6510} \\
 .1 &
\cellcolor{blue!15}.7809 & \cellcolor{blue!15}.8405 & \cellcolor{blue!25}\textbf{.8568} & \cellcolor{blue!25}\textbf{.8519} & \cellcolor{blue!25}\textbf{.7988} & \cellcolor{blue!25}\textbf{.9604} & \cellcolor{blue!15}.9024 &
\cellcolor{blue!15}.4737 & \cellcolor{blue!15}.9440 & \cellcolor{blue!15}.9265 & \cellcolor{blue!5}.6033 & \cellcolor{blue!15}.6432 & \cellcolor{blue!15}.6463 \\
 .5 &
\cellcolor{blue!5}.7760 & \cellcolor{blue!25}\textbf{.8416} & \cellcolor{blue!5}.8440 & \cellcolor{blue!15}.8429 & \cellcolor{blue!15}.7742 & \cellcolor{blue!15}.9220 & \cellcolor{blue!5}.9022 &
\cellcolor{blue!5}.4708 & \cellcolor{blue!5}.9345 & \cellcolor{blue!5}.9255 & \cellcolor{blue!15}.6038 & \cellcolor{blue!5}.6421 & \cellcolor{blue!5}.6308 \\
\bottomrule
\end{tabular}}
\end{table*}

\textbf{Effect of parameter $\alpha$}:
We provide experiments on the effect of $\alpha$, using the medium variant of the \acro{} model backbone, and pretrained the model with $\alpha=\{.1, .2, .3, .4, .5\}$. Results are enumerated in~\tableautorefname~\ref{apptab:alpha}. The model remains robust when $\alpha$ is within the range $0.1-0.3$, with $0.007$ standard deviation averaged across all tasks. Increasing $\alpha$ beyond $0.3$ results in a performance degradation, indicating that over-compressing the weighting factor can lead to suboptimal representation learning. Interestingly, if we compare the performance with Table 4, it can be observed that the performance when $\alpha=0.5$ is similar to SimCLR, implying the weighting factor is over-compressed and becomes less effective. Overall, our results confirm that while $\alpha$ can affect the shape of the embedding space, it does not directly drive the performance gains. The main improvements in the performance are due to using clinically-guided relative risk scores rather than the exact margin offset defined by $\alpha$.

\textbf{Effect of parameter $\tau$}
In our experiment, $\tau$ is set to $0.07$ following prior work in contrastive learning~\citep{wu2018unsupervised, he2020momentum}. To further validate the sensitivity of the model to $\tau$ we conducted further experiments using the small and medium variants of the \acro{} model backbone, with $\tau$ set to $\{0.01, 0.1, 0.5\}$. Results are enumerated in~\tableautorefname~\ref{apptab:tau}. Performance is best at small $\tau$ ($\tau=\{0.07, 0.1\}$) and degrades by little as $\tau$ increases. This indicates that lower $\tau$ enhances the discrimination between positives and negatives and is beneficial for learning clinical-informed ECG representations.

\begin{algorithm}[!b]
\caption{Clinically-Guided Contrastive Pretraining and Evaluation}
\label{alg:clef}
\textbf{Pretraining Phase:}\\
\KwIn{ Unlabeled ECG dataset $\mathbbm{D}=\{(\mathbf{e}_s=(\mathbf{e}_s^1, \dots, \mathbf{e}_s^{12}), \texttt{a}_s)\}_{s=1}^{N}$,
      risk score function $\mathcal{R}(\cdot)$,}       
\KwOut{ Foundation ECG model $\mathcal{F}(\cdot)$}
\For{each minibatch $\{\mathbf{e}^l_s\}_{s=1}^B$}{
    \For{each $\mathbf{e}^l_s$}{
        Sample stochastic augmentations $\mathbf{x}_i = \mathcal{T}(\mathbf{e}^l_s), \; \mathbf{x}_j = \mathcal{T}(\mathbf{e}^l_s)$ \;
        Compute embeddings $\mathbf{z}_i = \mathcal{F}(\mathbf{x}_i), \; \mathbf{z}_j = \mathcal{F}(\mathbf{x}_j)$ \;
        Obtain risk score $r_s = \mathcal{R}(\texttt{a}_s)$ from clinical metadata (Eq.~\eqref{eqn_risk_score}) \;
    }
    \For{each pair $(\mathbf{x}_i, \mathbf{x}_k)$ in batch}{
        Compute risk dissimilarity $\delta_{ik} = (r^i - r^k)^2$ \;
        Normalize to $\mathbf{D}_{ik}\!\in\![\alpha, 1]$ (Eq.~\eqref{eqn_rs_diss}) \;
        Compute metadata missingness weight $\mathbf{M}_{ik}$ (Eq.~\eqref{eqn_missing_meta}) \;
        Set final weight $\mathbf{W}_{ik} = \mathbf{D}_{ik} \cdot \mathbf{M}_{ik}$ \;
    }
    Compute weighted contrastive loss $\mathcal{L}^{w}$ (Eq.~\eqref{eqn_weighted_cl}) \;
    Compute dissimilarity alignment loss $\mathcal{L}^{d}$ (Eq.~\eqref{eqn_dissim_align}) \;
    Update encoder $\mathcal{F}(\cdot)$ by minimizing $\mathcal{L} = \mathcal{L}^{w} + \mathcal{L}^{d}$ \;
}
\vspace{1em}  
\textbf{Evaluation Phase:}\\
\KwIn{
downstream labeled dataset $\mathbbm{C}=\{(\mathbf{z}_s=\mathcal{F}(\mathbf{e}^l_s), {y}_s)\}_{s=1}^{M}$
}
\KwOut{ downstream linear model $\mathcal{G}(\cdot)$}
\For{each labeled ECG $(\mathbf{e}_s, y_s)$ in $\mathbbm{C}_{\text{train}}$}{
    Extract embedding $\mathbf{z}_s = \mathcal{F}(\mathbf{e}_s)$ \;
    Train/update a linear classifier/regressor $\hat{y}_s = \mathcal{G}(\mathbf{z}_s)$ \;
}
\For{each labeled ECG $(\mathbf{e}_s, y_s)$ in $\mathbbm{C}_{\text{val}}$}{
    Extract embedding $\mathbf{z}_s = \mathcal{F}(\mathbf{e}_s)$ \;
    Predict discrete/continuous values $\hat{y}_s = \mathcal{G}(\mathbf{z}_s)$ \;
}
Select $\mathcal{G}$ with best validation loss \;
\For{each labeled ECG $(\mathbf{e}_s, y_s)$ in $\mathbbm{C}_{\text{test}}$}{
    Extract embedding $\mathbf{z}_s = \mathcal{F}(\mathbf{e}_s)$ \;
    Predict discrete/continuous values $\hat{y}_s = \mathcal{G}(\mathbf{z}_s)$ \;
}
Evaluate $\{\hat{y}_s\}$ against $\{y_s\}$ using AUROC (classification) or MAE (regression) \;

\end{algorithm}

\subsection{Computational Complexity Analysis}

The proposed clinically-guided pretraining framework introduces minimal computational overhead compared to the baseline SimCLR. 
Specifically, risk scores are computed once per patient from clinical metadata during the preprocessing stage, prior to model training. Following the SCORE2 algorithm detailed in~\appendixautorefname~\ref{appsubsec:obtain_riskscore}, each patient's risk score is derived from $7$ clinical covariates through a series of operations (Equations S1--S3). For a single patient, this requires $O(1)$ operations, as the number of covariates and parameters is fixed. 

Additionally, during training, the clinical guidance adds two operations per mini-batch: (i) obtaining weight matrix $\mathbf{W}_{i,k}$ from pairwise risk score dissimilarities $\mathbf{D}_{i,k}$ and missing metadata matrix $\mathbf{M}$, and (ii) computing the dissimilarity alignment loss $L^d$ (Eq.~\ref{eqn_dissim_align}). Both operations require $O(B^2)$ computations for batch size $B$. In contrast, the standard SimCLR contrastive loss requires $O(B^2 d)$ operations for $d$-dimensional embeddings. Since for \acro{}, $d \in \{256, 1024, 2048\}$, the additional $O(B^2)$ cost results in no significant increase in complexity.

\section{Pretraining Algorithms}

Algorithm~\ref{alg:clef} summarizes our clinically-guided contrastive pretraining and downstream evaluation procedure. In the pretraining phase, we sample minibatches of ECG embeddings and generate stochastic augmentations, which are encoded by the foundation model $\mathcal{F}(\cdot)$. Risk scores from clinical metadata are used to compute pairwise dissimilarities, which are combined with multipliers weighted by the number of missing metadata to form the final pairwise weights. The foundation model $\mathcal{F}$ is updated by minimizing a combination of weighted contrastive loss $\mathcal{L}^{w}$ and dissimilarity alignment loss $\mathcal{L}^{d}$. In the evaluation phase, embeddings of downstream labeled ECGs are extracted and used to train a linear classifier or regressor $\mathcal{G}(\cdot)$. Model selection is performed by choosing the model that achieves the lowest validation loss. The final model is evaluated on the held-out test set using AUROC (classification) or MAE (regression).

\section{The Choice of Risk Score}
\label{app:choice_risk_score}

In this work, we used the SCORE2 risk score to guide contrastive learning. SCORE2 was selected because it is a standard cardiovascular risk score relevant to ECG data, which has been externally validated on data from over one million individuals from $15$ countries, and its required input variables overlap with the metadata in the MIMIC-IV-ECG pretraining dataset. However, we do not claim that SCORE2 is the optimal risk score for \acro{} as we did not test other risk scores. Furthermore, the optimal risk score might vary across applications. For instance, SCORE2 was developed using data from European subjects, for adults aged 40-69, and for long-term cardiovascular risk prediction. Alternative risk scores may be more appropriate for non-European populations, for different age groups, and for non-cardiovascular risk assessment. In addition, the optimal risk score will likely vary depending on the available metadata in a pretraining dataset.

\end{document}